\documentclass[10pt,twocolumn,letterpaper]{article}

\usepackage{iccv}
\usepackage{times}
\usepackage{epsfig}
\usepackage{graphicx}
\usepackage{amsmath}
\usepackage{amssymb}
\usepackage{adjustbox}
\usepackage{multirow}
\usepackage{algpseudocode}
\usepackage{algorithm}
\usepackage{pgf} 
\usepackage{color}
\usepackage{colortbl}
\usepackage{etoolbox}
\usepackage{pgf} 
\usepackage{pgfplots}
\usepackage{amssymb}
\usepackage{pifont,xcolor}
\usepackage[format=plain,labelformat=simple,labelsep=period,font=small,skip=4pt,compatibility=false]{caption}
\usepackage[font=footnotesize,skip=2pt]{subcaption}
\usepackage{csquotes}
\usepackage{booktabs}
\usepackage{array}
\usepackage{nicefrac}
\usepackage{tabularx}
\usepackage{siunitx}

\usepackage[accsupp]{axessibility}  

\RequirePackage{cite}     

\newcommand{\cmark}{\ding{51}}%
\newcommand{\xmark}{\ding{55}}%
\definecolor{low}{HTML}{0071BC}
\definecolor{high}{HTML}{42D022}
\newcommand*{\opacity}{60}
\newcommand*{\minval}{23.3}
\newcommand*{\maxval}{78.9}

\newcommand*{\minvalpmr}{11.0}
\newcommand*{\maxvalpmr}{52.4}

\newcommand*{\minvalpm}{30}
\newcommand*{\maxvalpm}{65.7}

\newcommand*{\minvalc}{16.2}
\newcommand*{\maxvalc}{48}

\newcommand*{\minvalcmr}{6.4}
\newcommand*{\maxvalcmr}{27.4}

\newcommand*{\minvalcm}{15}
\newcommand*{\maxvalcm}{35.7}

\newcommand*{\minvald}{14.9}
\newcommand*{\maxvald}{37.2}

\newcommand*{\minvaldmr}{3.3}
\newcommand*{\maxvaldmr}{24.0}

\newcommand*{\minvaldm}{12.5}
\newcommand*{\maxvaldm}{24.2}

\newcommand*{\minvalg}{35}
\newcommand*{\maxvalg}{78.3}

\newcommand*{\minvalpmrg}{10.1}
\newcommand*{\maxvalpmrg}{50.5}

\newcommand*{\minvalpmg}{33.0}
\newcommand*{\maxvalpmg}{64.4}

\newcommand*{\minvalcg}{21}
\newcommand*{\maxvalcg}{41.5}

\newcommand*{\minvalcmrg}{7.9}
\newcommand*{\maxvalcmrg}{22.5}

\newcommand*{\minvalcmg}{19.8}
\newcommand*{\maxvalcmg}{29.9}

\newcommand*{\minvaldg}{14.8}
\newcommand*{\maxvaldg}{33.1}

\newcommand*{\minvaldmrg}{5.6}
\newcommand*{\maxvaldmrg}{20.1}

\newcommand*{\minvaldmg}{14.8}
\newcommand*{\maxvaldmg}{23.5}

\newcommand*{\minvala}{33.5}
\newcommand*{\maxvala}{36.6}

\newcommand*{\minvalamr}{7.0}
\newcommand*{\maxvalamr}{17.4}

\newcommand*{\minvalam}{19.8}
\newcommand*{\maxvalam}{26.9}

\newcommand{\gradient}[1]{
    \ifdimcomp{#1pt}{>}{\maxval pt}{#1}{
        \ifdimcomp{#1pt}{<}{\minval pt}{#1}{
            \pgfmathparse{int(round(100*(#1/(\maxval-\minval))-(\minval*(100/(\maxval-\minval)))))}
            \xdef\tempa{\pgfmathresult}
            \cellcolor{high!\tempa!low!\opacity} #1
    }}
}

\newcommand{\gradientpmr}[1]{
    \ifdimcomp{#1pt}{>}{\maxvalpmr pt}{#1}{
        \ifdimcomp{#1pt}{<}{\minvalpmr pt}{#1}{
            \pgfmathparse{int(round(100*(#1/(\maxvalpmr-\minvalpmr))-(\minvalpmr*(100/(\maxvalpmr-\minvalpmr)))))}
            \xdef\tempa{\pgfmathresult}
            \cellcolor{high!\tempa!low!\opacity} #1
    }}
}

\newcommand{\gradientpm}[1]{
    \ifdimcomp{#1pt}{>}{\maxvalpm pt}{#1}{
        \ifdimcomp{#1pt}{<}{\minvalpm pt}{#1}{
            \pgfmathparse{int(round(100*(#1/(\maxvalpm-\minvalpm))-(\minvalpm*(100/(\maxvalpm-\minvalpm)))))}
            \xdef\tempa{\pgfmathresult}
            \cellcolor{high!\tempa!low!\opacity} #1
    }}
}

\newcommand{\gradientc}[1]{
    \ifdimcomp{#1pt}{>}{\maxvalc pt}{#1}{
        \ifdimcomp{#1pt}{<}{\minvalc pt}{#1}{
            \pgfmathparse{int(round(100*(#1/(\maxvalc-\minvalc))-(\minvalc*(100/(\maxvalc-\minvalc)))))}
            \xdef\tempa{\pgfmathresult}
            \cellcolor{high!\tempa!low!\opacity} #1
    }}
}

\newcommand{\gradientcmr}[1]{
    \ifdimcomp{#1pt}{>}{\maxvalcmr pt}{#1}{
        \ifdimcomp{#1pt}{<}{\minvalcmr pt}{#1}{
            \pgfmathparse{int(round(100*(#1/(\maxvalcmr-\minvalcmr))-(\minvalcmr*(100/(\maxvalcmr-\minvalcmr)))))}
            \xdef\tempa{\pgfmathresult}
            \cellcolor{high!\tempa!low!\opacity} #1
    }}
}

\newcommand{\gradientcm}[1]{
    \ifdimcomp{#1pt}{>}{\maxvalcm pt}{#1}{
        \ifdimcomp{#1pt}{<}{\minvalcm pt}{#1}{
            \pgfmathparse{int(round(100*(#1/(\maxvalcm-\minvalcm))-(\minvalcm*(100/(\maxvalcm-\minvalcm)))))}
            \xdef\tempa{\pgfmathresult}
            \cellcolor{high!\tempa!low!\opacity} #1
    }}
}

\newcommand{\gradientd}[1]{
    \ifdimcomp{#1pt}{>}{\maxvald pt}{#1}{
        \ifdimcomp{#1pt}{<}{\minvald pt}{#1}{
            \pgfmathparse{int(round(100*(#1/(\maxvald-\minvald))-(\minvalpmr*(100/(\maxvald-\minvald)))))}
            \xdef\tempa{\pgfmathresult}
            \cellcolor{high!\tempa!low!\opacity} #1
    }}
}

\newcommand{\gradientdmr}[1]{
    \ifdimcomp{#1pt}{>}{\maxvaldmr pt}{#1}{
        \ifdimcomp{#1pt}{<}{\minvaldmr pt}{#1}{
            \pgfmathparse{int(round(100*(#1/(\maxvaldmr-\minvaldmr))-(\minvaldmr*(100/(\maxvaldmr-\minvaldmr)))))}
            \xdef\tempa{\pgfmathresult}
            \cellcolor{high!\tempa!low!\opacity} #1
    }}
}

\newcommand{\gradientdm}[1]{
    \ifdimcomp{#1pt}{>}{\maxvaldm pt}{#1}{
        \ifdimcomp{#1pt}{<}{\minvaldm pt}{#1}{
            \pgfmathparse{int(round(100*(#1/(\maxvaldm-\minvaldm))-(\minvaldm*(100/(\maxvaldm-\minvaldm)))))}
            \xdef\tempa{\pgfmathresult}
            \cellcolor{high!\tempa!low!\opacity} #1
    }}
}

\newcommand{\gradientg}[1]{
    \ifdimcomp{#1pt}{>}{\maxvalg pt}{#1}{
        \ifdimcomp{#1pt}{<}{\minvalg pt}{#1}{
            \pgfmathparse{int(round(100*(#1/(\maxvalg-\minvalg))-(\minvalg*(100/(\maxvalg-\minvalg)))))}
            \xdef\tempa{\pgfmathresult}
            \cellcolor{high!\tempa!low!\opacity} #1
    }}
}

\newcommand{\gradientpmrg}[1]{
    \ifdimcomp{#1pt}{>}{\maxvalpmrg pt}{#1}{
        \ifdimcomp{#1pt}{<}{\minvalpmrg pt}{#1}{
            \pgfmathparse{int(round(100*(#1/(\maxvalpmrg-\minvalpmrg))-(\minvalpmrg*(100/(\maxvalpmrg-\minvalpmrg)))))}
            \xdef\tempa{\pgfmathresult}
            \cellcolor{high!\tempa!low!\opacity} #1
    }}
}

\newcommand{\gradientpmg}[1]{
    \ifdimcomp{#1pt}{>}{\maxvalpmg pt}{#1}{
        \ifdimcomp{#1pt}{<}{\minvalpmg pt}{#1}{
            \pgfmathparse{int(round(100*(#1/(\maxvalpmg-\minvalpmg))-(\minvalpmg*(100/(\maxvalpmg-\minvalpmg)))))}
            \xdef\tempa{\pgfmathresult}
            \cellcolor{high!\tempa!low!\opacity} #1
    }}
}

\newcommand{\gradientcg}[1]{
    \ifdimcomp{#1pt}{>}{\maxvalcg pt}{#1}{
        \ifdimcomp{#1pt}{<}{\minvalcg pt}{#1}{
            \pgfmathparse{int(round(100*(#1/(\maxvalcg-\minvalcg))-(\minvalcg*(100/(\maxvalcg-\minvalcg)))))}
            \xdef\tempa{\pgfmathresult}
            \cellcolor{high!\tempa!low!\opacity} #1
    }}
}

\newcommand{\gradientcmrg}[1]{
    \ifdimcomp{#1pt}{>}{\maxvalcmrg pt}{#1}{
        \ifdimcomp{#1pt}{<}{\minvalcmrg pt}{#1}{
            \pgfmathparse{int(round(100*(#1/(\maxvalcmrg-\minvalcmrg))-(\minvalcmrg*(100/(\maxvalcmrg-\minvalcmrg)))))}
            \xdef\tempa{\pgfmathresult}
            \cellcolor{high!\tempa!low!\opacity} #1
    }}
}

\newcommand{\gradientcmg}[1]{
    \ifdimcomp{#1pt}{>}{\maxvalcmg pt}{#1}{
        \ifdimcomp{#1pt}{<}{\minvalcmg pt}{#1}{
            \pgfmathparse{int(round(100*(#1/(\maxvalcmg-\minvalcmg))-(\minvalcmg*(100/(\maxvalcmg-\minvalcmg)))))}
            \xdef\tempa{\pgfmathresult}
            \cellcolor{high!\tempa!low!\opacity} #1
    }}
}

\newcommand{\gradientdg}[1]{
    \ifdimcomp{#1pt}{>}{\maxvaldg pt}{#1}{
        \ifdimcomp{#1pt}{<}{\minvaldg pt}{#1}{
            \pgfmathparse{int(round(100*(#1/(\maxvaldg-\minvaldg))-(\minvaldg*(100/(\maxvaldg-\minvaldg)))))}
            \xdef\tempa{\pgfmathresult}
            \cellcolor{high!\tempa!low!\opacity} #1
    }}
}

\newcommand{\gradientdmrg}[1]{
    \ifdimcomp{#1pt}{>}{\maxvaldmrg pt}{#1}{
        \ifdimcomp{#1pt}{<}{\minvaldmrg pt}{#1}{
            \pgfmathparse{int(round(100*(#1/(\maxvaldmrg-\minvaldmrg))-(\minvaldmrg*(100/(\maxvaldmrg-\minvaldmrg)))))}
            \xdef\tempa{\pgfmathresult}
            \cellcolor{high!\tempa!low!\opacity} #1
    }}
}

\newcommand{\gradientdmg}[1]{
    \ifdimcomp{#1pt}{>}{\maxvaldmg pt}{#1}{
        \ifdimcomp{#1pt}{<}{\minvaldmg pt}{#1}{
            \pgfmathparse{int(round(100*(#1/(\maxvaldmg-\minvaldmg))-(\minvaldmg*(100/(\maxvaldmg-\minvaldmg)))))}
            \xdef\tempa{\pgfmathresult}
            \cellcolor{high!\tempa!low!\opacity} #1 
    }}
}

\newcommand{\gradienta}[1]{
    \ifdimcomp{#1pt}{>}{\maxvala pt}{#1}{
        \ifdimcomp{#1pt}{<}{\minvala pt}{#1}{
            \pgfmathparse{int(round(100*(#1/(\maxvala-\minvala))-(\minvala*(100/(\maxvala-\minvala)))))}
            \xdef\tempa{\pgfmathresult}
            \cellcolor{high!\tempa!low!\opacity} #1
    }}
}

\newcommand{\gradientamr}[1]{
    \ifdimcomp{#1pt}{>}{\maxvalamr pt}{#1}{
        \ifdimcomp{#1pt}{<}{\minvalamr pt}{#1}{
            \pgfmathparse{int(round(100*(#1/(\maxvalamr-\minvalamr))-(\minvalamr*(100/(\maxvalamr-\minvalamr)))))}
            \xdef\tempa{\pgfmathresult}
            \cellcolor{high!\tempa!low!\opacity} #1
    }}
}

\newcommand{\gradientam}[1]{
    \ifdimcomp{#1pt}{>}{\maxvalam pt}{#1}{
        \ifdimcomp{#1pt}{<}{\minvalam pt}{#1}{
            \pgfmathparse{int(round(100*(#1/(\maxvalam-\minvalam))-(\minvalam*(100/(\maxvalam-\minvalam)))))}
            \xdef\tempa{\pgfmathresult}
            \cellcolor{high!\tempa!low!\opacity} #1
    }}
}

\usepackage[breaklinks=true,bookmarks=false]{hyperref}

\iccvfinalcopy 

\def\iccvPaperID{9029} 

\ificcvfinal\fi

\begin{document}

\title{Vision Relation Transformer for Unbiased Scene Graph Generation}

\author{Gopika Sudhakaran\textsuperscript{1,3}\\
\and
Devendra Singh Dhami\textsuperscript{1,3}\\
\and
Kristian Kersting\textsuperscript{1,2,3}\\
\and
Stefan Roth\textsuperscript{1,2,3}\\
\and
\newline
\textsuperscript{1}Department of Computer Science, Technical University of Darmstadt, Germany\\
\textsuperscript{2}Centre for Cognitive Science, TU Darmstadt\qquad 
\textsuperscript{3}Hessian Center for AI (hessian.AI)\\
}

\maketitle
\ificcvfinal\thispagestyle{empty}\fi
\newcommand{\dd}[1]{\textcolor{red}{[#1 \textsc{--Dev}]}}

\begin{abstract}
   Recent years have seen a growing interest in Scene Graph Generation (SGG), a
comprehensive visual scene understanding task that aims to predict entity relationships
using a relation encoder-decoder pipeline stacked on top of an object encoder-decoder backbone. Unfortunately, current SGG methods suffer from an information loss regarding the entities' local-level cues during the relation encoding process. To mitigate this, we introduce the \textbf{V}ision r\textbf{E}lation \textbf{T}ransf\textbf{O}rmer (VETO), consisting of a novel local-level entity relation encoder. We further observe that many existing SGG methods claim to be unbiased, but are still biased towards either head or tail classes. To overcome this bias, we introduce a \textbf{M}utually \textbf{E}xclusive \textbf{E}xper\textbf{T} (MEET) learning strategy that captures important relation features without bias towards head or tail classes. Experimental results on the VG and GQA datasets demonstrate that VETO + MEET boosts the predictive performance by up to 47\% over the state of the art 
while being $\sim 10 \times$ smaller.\footnote{Code is available at \url{https://github.com/visinf/veto}}
\end{abstract}

\section{Introduction}

Visual scene understanding has made great strides in recent years, extending beyond standard object detection and recognition tasks to tackle more complex problems such as visual question answering~\cite{antol2015vqa} and image captioning~\cite{hossain2019comprehensive}. One powerful tool for scene understanding is Scene Graph Generation (SGG), which aims to identify the relationships between entities in a scene \cite{lu2016visual}.
However, despite recent advancements, SGG models still have significant limitations when it comes to real-world applications.
\begin{figure}[t]
  \centering
   \includegraphics[width=\linewidth]{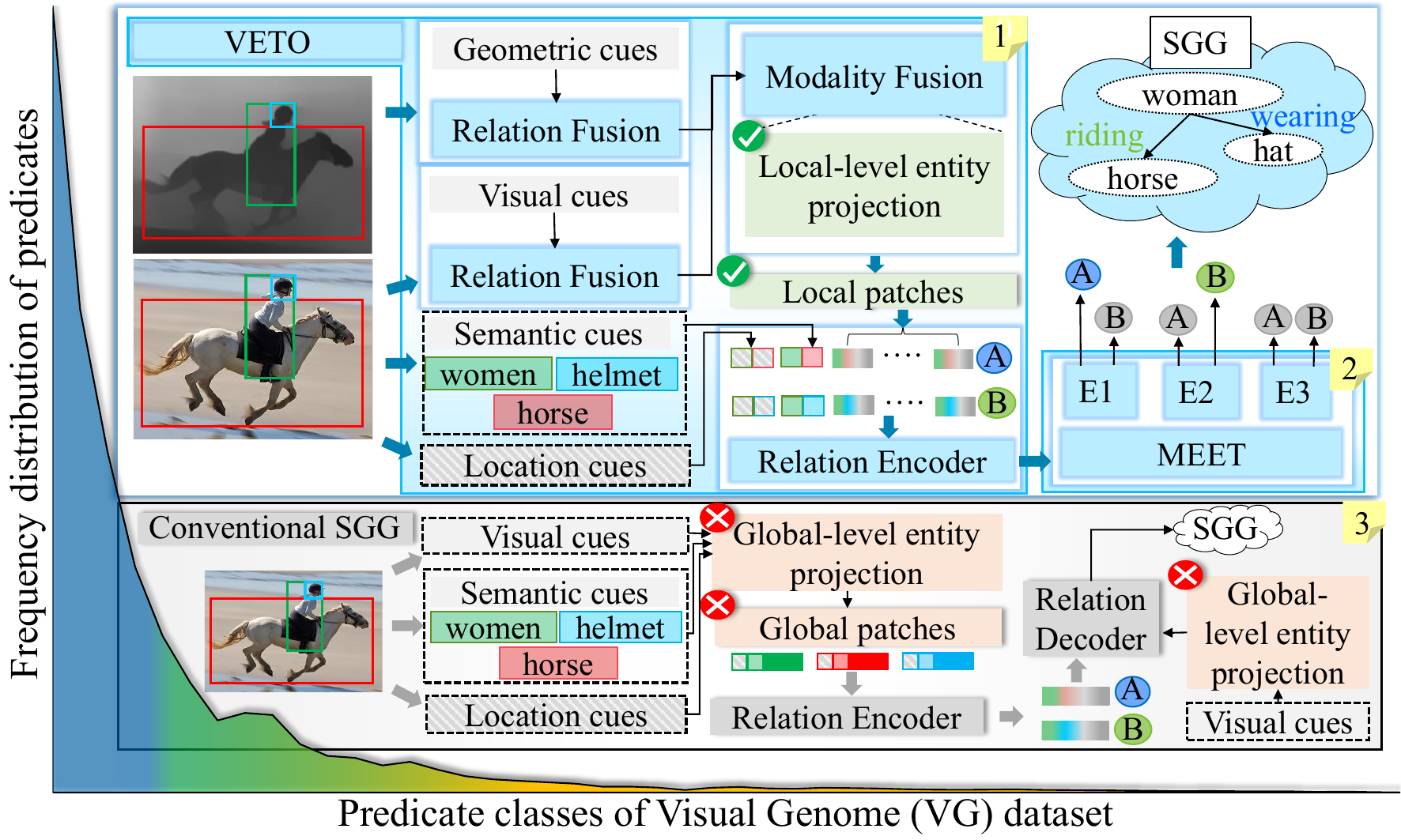}
   \caption{\textbf{VETO-MEET \vs Conventional SGG.} \emph{(1)} VETO: Enhancing the information flow from entity features to relationship prediction by using a local-level entity relation encoder that conducts relation and modality fusion of local-level entity patches. The local-level components (green ticks) keep the model light-weight while reducing information loss. A (blue) and B (green) denote example relation classes taken from the corresponding colored region in the predicate frequency histogram of the VG dataset \cite{krishna2017visual}. 
   \emph{(2)} MEET: Debiased relation decoder that employs out-of-distribution aware mutually exclusive experts (E1--E3). Grey A and B denote an out-of-distribution prediction discarded by the model. 
   \emph{(3)} Conventional SGG: The projection components (red crosses) yield a computationally expensive model and the global-level entity patches result in a local-level information loss.}
\label{fig:intro}
\vspace{-0.5em}
\end{figure}

\begin{figure}
    \centering
    \begin{subfigure}[b]{0.5\textwidth}
        \centering
        \includegraphics[height=2cm, width=0.8\linewidth]{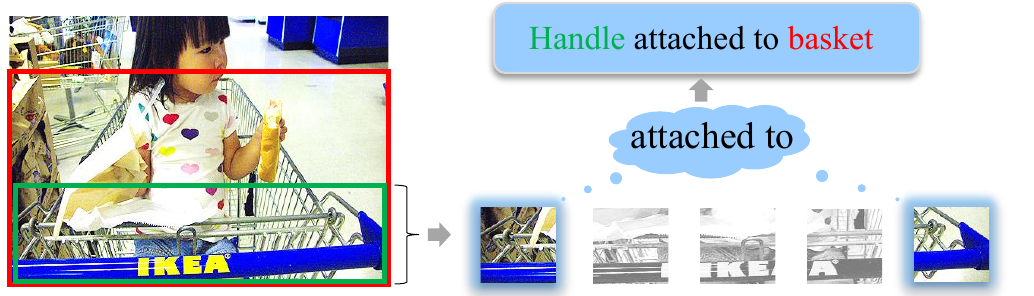}
        \caption{Significance of local-level cues}
    \label{fig.part}
    \end{subfigure}
    \hfill
    \begin{subfigure}[b]{0.4\textwidth}
        \centering
        \includegraphics[width=\linewidth]{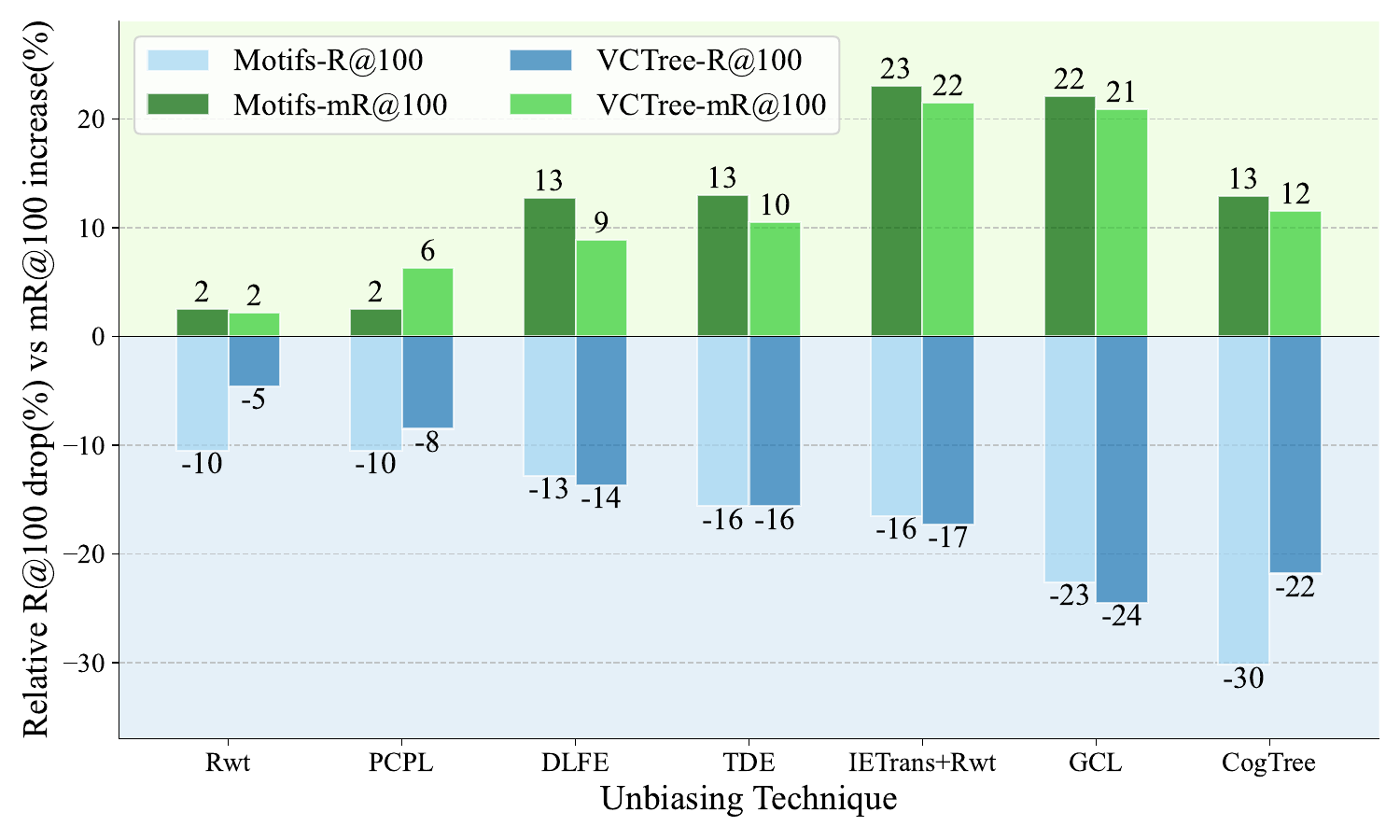}
        \caption{Recall drop of debiased Motifs~\cite{zellers2018neural}, VCTree~\cite{tang2019learning}}
        \label{fig.bias}
    \end{subfigure}
    \caption{\textbf{Challenges in SGG.} \emph{(a)} For establishing the \emph{attached to} relation between \emph{Handle} and \emph{Basket}, the attention should be on the corner regions of the object. 
    \emph{(b)} R@100 drop (\%) and mR@100  increase (\%) of unbiased SGG methods Motifs and VCTree relative to their vanilla versions. The R@100 metric measures the average recall of all predictions, which is higher for models that overfit to the head classes, while mR@100 denotes the per-predicate class mean and is higher for models that overfit to the tail classes. }
    \vspace{-0.5em}
\end{figure}

Conventional SGG approaches, as shown in Fig.~\ref{fig:intro} (panel 3), generate global-level entity patches for relation encoding. Yet, during the relation encoding process, they lose \emph{local-level} entity information. As illustrated in Fig.~\ref{fig.part}, we humans have a tendency to focus on the critical local-level information necessary to construct relations between things in a scene, which is overlooked by current SGG approaches.
Moreover, the major parameter count of current SGG models stems from projections (red-crossed components in Fig.~\ref{fig:intro}) involved in global-level entity patch generation. 
Another challenge with  existing SGG approaches, despite efforts to enhance scene graphs using additional cues like depth maps and knowledge graphs \cite{sharifzadeh2021improving, zareian2020bridging}, is that they are either resource intensive or limited in exploiting cross-modal information. 

Finally but crucially, SGG training setups are challenged by the strong
bias of the visual world around us towards a few frequently occurring relationships, leaving a long tail of under-represented relations.
This is also the case with benchmark SGG datasets, \eg, Visual Genome (VG)~\cite{krishna2017visual}, as depicted by the predicate\footnote{We use the terms predicate/relation interchangeably in this paper.} class frequency distribution in Fig.~\ref{fig:intro}. 
Due to the dominance of few head predicates, conventional SGG models \cite{tang2019learning, zellers2018neural} are biased towards the head classes. Though several unbiased SGG methods have been proposed \cite{dong2022stacked, tang2020unbiased, yu2020cogtree} to overcome this issue, they are prone to over-fitting to the tail classes at the expense of head classes (\cf Fig.~\ref{fig.bias}). Despite recent efforts \cite{dong2022stacked} to fix this bias issue using multi-expert learning strategies, we find that they still over-fit to the tail classes (\enquote{GCL} in Fig.~\ref{fig.bias}). 
Overall, there are two main problems with present unbiased SGG methods: \emph{(1)} Conventional methods, including debiased models, can only learn a limited range of predicates. \emph{(2)} Existing multi-expert SGG models lack adequate exclusivity to enhance both head and tail classes at the same time.

Consequently, we propose the \emph{Vision rElation TransfOrmer (VETO)}. Inspired by Vision Transformers \cite{dosovitskiy2020image} that use image-level patches for classification,
VETO generates \emph{local-level} entity patches for the relation prediction
task. This improves the information flow from entity features to relationship prediction by channeling the attention towards fused \emph{local} feature cues of subject and object entities (Relation Fusion in Fig.~\ref{fig:intro}) and using a local-level entity relation encoder, which processes entity features at the sub-region level. To strengthen the encoder further, we infuse geometric cues into VETO using a Modality Fusion component (\cf Fig.~\ref{fig:intro}), which unites visual and geometric features to yield local-level entity patches. 
Finally, to successfully debias VETO, we propose a multi-expert learning strategy termed \emph{Mutually Exclusive ExperTs (MEET)}.  After splitting the predicate classes into subgroups, we perform in-distribution and out-of-distribution (OOD) sampling for each subgroup. Then we train each expert on every predicate class but each expert will be responsible for only a subset of predicates with out-of-distribution prediction handling predicates outside its subgroup. In contrast to existing multi-expert methods \cite{dong2022stacked}, where expert classifiers are co-dependent to distill knowledge, OOD sampling enables experts to independently interpret every inference sample.


\textbf{Contributions.} Let us summarize:
\emph{(1)} We propose a novel SGG method with a local-level entity relation encoding, which is light-weight and reduces the local-level information loss of entities. \emph{(2)} To strengthen the encoder further, we propose an effective strategy to infuse additional geometric cues. \emph{(3)} We devise a mutually exclusive multi-expert learning strategy that effectively exploits our relation network design by learning subgroup-specific diverse feature representations and discriminating from samples outside its subgroups. \emph{(4)} Our extensive experimentation shows the significance of both VETO and MEET.

\section{Related Work}
\textbf{Scene Graph Generation} (SGG) is a tool for understanding scenes by simplifying the visual relationships into a summary graph. SGG has been receiving increased attention from the research community due to its potential usability in assisting downstream visual reasoning tasks \cite{krishna2018referring, shi2019explainable, wang2019exploring}. While SGG aims to provide a comprehensive view of relationships between objects in a visual scene, there is another set of research that represents interactions as relationships between humans and objects called Human-object Interaction (HOI)~\cite{gao2018ican, ulutan2020vsgnet, iftekhar2023gtnet, xu2019learning}.  In this work, the focus is on SGG and its associated literature, emphasizing the study of object relationships within visual scenes.

The SGG task was first introduced by Lu \etal~\cite{lu2016visual}.  Early approaches mainly focused on including additional features from various sources other than the visual context, resulting in sub-optimal performance \cite{ dai2017detecting, liao2019natural, lu2016visual}. Later work proposed more powerful relation encoders with rich contextual information by employing message passing \cite{xu2017scene}, sequential LSTMs \cite{tang2019learning, zellers2018neural}, and fully-connected graphs \cite{chen2019knowledge, dai2017detecting,  li2017scene,  wang2019exploring, woo2018linknet, xu2017scene, yin2018zoom, zareian2020bridging}.
Recent advancements in attention techniques have also resulted in attention-based SGG methods. Earlier work~\cite{yang2018graph} in this direction used graph attention networks (GAT)~\cite{velickovic2018graph} to capture object-level visual similarity. Recently, Transformers~\cite{vaswani2017attention} have also been used for SGG~\cite{dong2022stacked, koner2020relation, lin2020gps, lu2021context} after their successful adoption across computer vision~\cite{carion2020end, dosovitskiy2020image, ramachandran2019stand}. Current transformer-based SGG methods use attention to capture global context and improve the visual and semantic modality fusion. Lu \etal~\cite{lu2021context} used sequential decoding to capture context, while Dong \etal~\cite{dong2022stacked} employed self- and cross-attention to fuse visual and semantic cues. Deviating from this, we use transformers to capture local-level relation cues as well as joint visual and geometric cues.

\textbf{Scene Graphs with additional knowledge.}
Due to the long-tail distribution of the relationships, it is difficult to obtain enough training data for every relation. To overcome this, using additional knowledge in the form of knowledge graphs~\cite{gu2019scene, yu2017visual,  zareian2020bridging}, depth maps~\cite{sharifzadeh2021improving, yang2018visual}, and data transfer~\cite{zhang2022fine} was proposed. Knowledge graph-based works refine features for relation prediction by reasoning using knowledge from large-scale databases. Yang \etal~\cite{yang2018visual} and Sharifzahed \etal~\cite{sharifzadeh2021improving} use a monocular depth estimator to infer additional depth cues for relation prediction by fusing with visual features. Zhang \etal~\cite{zhang2022fine} expanded the dataset by increasing the SGG annotations through internal and external data transfer.
Our approach can also use depth maps to provide additional geometric knowledge. However, introducing additional knowledge can increase the parameter count and computation time of the model. To tackle this problem, we strategically prune the parameters, resulting in a light-weight yet powerful SGG model.

\textbf{Unbiased Scene Graph Generation.} 
The SGG research community started paying attention to the problem of class imbalance only after the introduction of the less biased mean recall metric by Chen \etal~\cite{chen2019knowledge} and Tang \etal~\cite{tang2019learning}. Subsequently, various unbiasing strategies~\cite{chiou2021recovering, dong2022stacked,  li2021bipartite, suhail2021energy, tang2020unbiased, wang2020tackling, yan2020pcpl, yu2020cogtree,  zhang2022fine} were proposed, many of which can be used in a model-agnostic fashion. Tang \etal~\cite{tang2019learning} used counterfactuals from causal inference to disentangle unbiased representations from the biased ones. Yu \etal~\cite{yu2020cogtree} utilized tree structures to filter irrelevant predicates. Zareian \etal~\cite{zareian2020bridging} and Yan \etal\cite{yan2020pcpl} used re-weighting strategies while Li \etal~\cite{li2021bipartite} employed a re-sampling strategy. Dong \etal~\cite{dong2022stacked} used a multi-expert learning setup that leverages knowledge distillation.
However, while these methods attain high performance on unbiased metrics, they reduce the head class performance significantly as seen in Fig.~\ref{fig.bias}. Hence, to attain an effective balance between the head and tail classes, we propose a mutually exclusive expert learning setup. Our model not only achieves better head and tail class balance but also sets a new state of the art.

\section{Vision rElation TransfOrmer (VETO)}
Our goal is to improve the Scene Graph Generation task that parses an input image to generate a structured graphical representation of entities and their relationships. In particular, we focus on enhancing the overall performance of SGG by improving the prediction on both the head and tail relations. To this end, we introduce a relation network that learns richer entity/predicate representations by focusing on \emph{local-level} entity features and devise a \emph{multi-expert} learning strategy to achieve a better relation prediction trade-off. 

\subsection{Problem setting}
For a given image $\mathbf{I}$, the goal of SGG is to create a summary graph $\mathcal{G}$ that adequately summarizes the information present in the image. At first, we detect all the entities within image $\mathbf{I}$, denoted as $\mathcal{E}=\left\{e_i\right\}_{i=1}^N$. Then we predict the predicates $p_{i \rightarrow j}$ for each subject-object entity pair $\left(e_i, e_j\right)$. Finally, we construct the scene graph $\mathcal{G}$ using the triplet form of the predictions  $\left(e_i, p_{i \rightarrow j}, e_j\right)$ as
\begin{equation}
  \mathcal{G} = \big\{\left(e_i, p_{i \rightarrow j}, e_j\right) \mid e_i, e_j \in \mathcal{E}, \; p_{i \rightarrow j} \in \mathcal{P}\big\}.
\end{equation}

\subsection{The VETO backbone}
\begin{figure*}[t]%
\centering
\includegraphics[width=\linewidth]{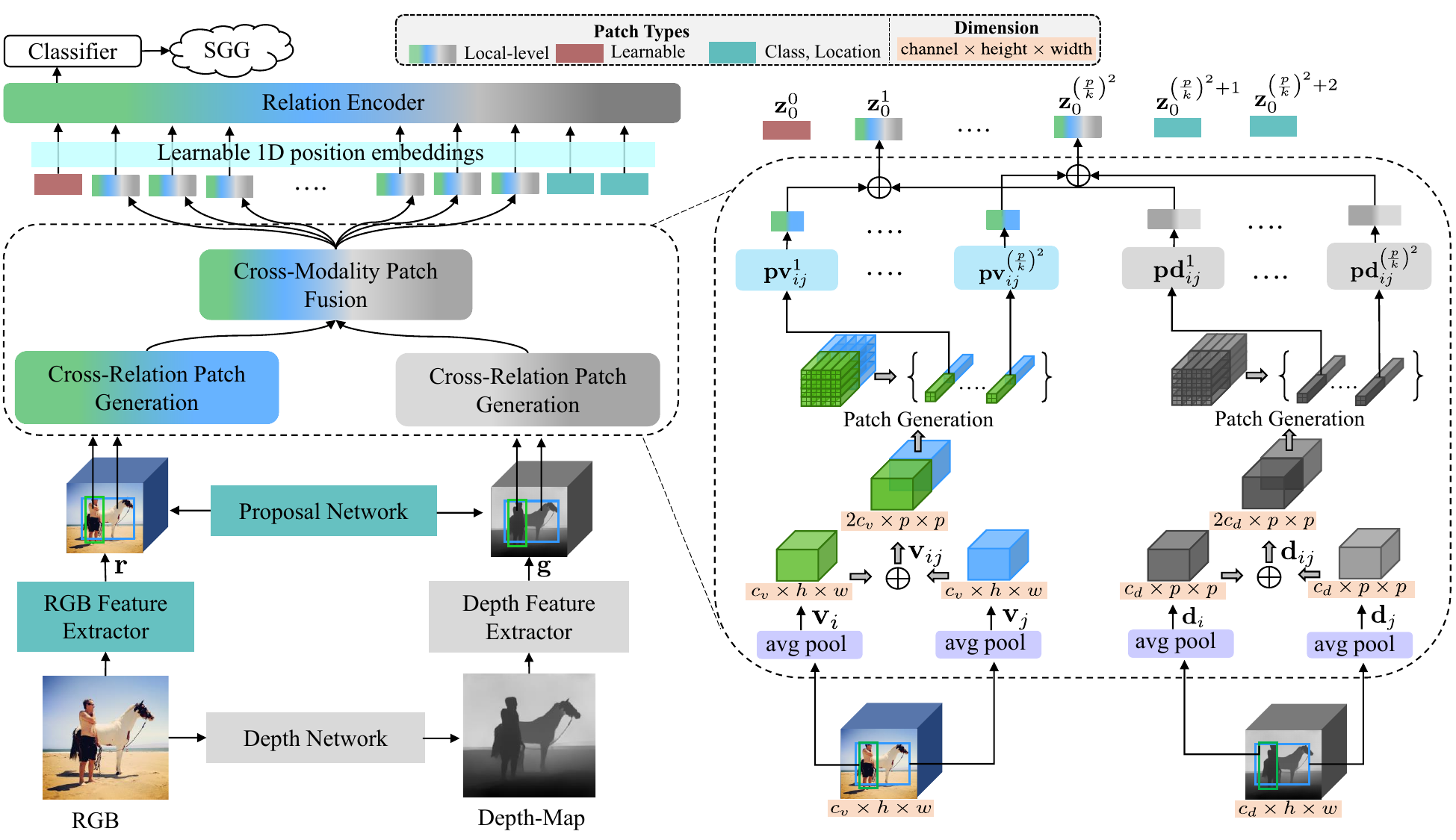}
\caption{\textbf{VETO architecture}. 
An object detector yields entity proposals and entity features $\mathbf{r}$. 
Moreover, a depth map is estimated from the RGB input, which is also passed through the feature extractor to obtain geometric features $\mathbf{g}$. 
Then, for each entity pair, a sequence of local-level patches are generated, which are passed through the transformer-based relation encoder to yield a relation prediction.}
\label{fig:veto}
\vspace{-0.5em}
\end{figure*}

Roughly speaking, as shown in Fig.~\ref{fig:veto}, the VETO model 
comprises a feature extraction and a proposal network as the backbone, which are fed to the relation network. 

\textbf{Feature extraction.} Following previous work, our feature extraction backbone comprises an RGB feature extractor, which is pre-trained and kept frozen \cite{tang2020unbiased, li2021bipartite}, and a depth feature extractor, which is trained from scratch during SGG training \cite{sharifzadeh2021improving}. 

\textbf{Proposal network.}
We use Faster R-CNN \cite{ren2015faster} as our object detector. Entity proposals are obtained directly from the output of object detection, which includes their categories and classification scores. We use the entity proposals to extract scaled RGB features $\mathbf{r}_i$ and their corresponding geometric features $\mathbf{g}_i$ from the depth map. We denote the proposal bounding box as $\mathbf{b}_i$ and its detected class as $\mathbf{c}_i$.

Before explaining the proposed VETO local-level entity generator, let us briefly revisit a conventional SGG \cite{li2021bipartite} pipeline that uses global-level entity projection.

\textbf{Entity global-level patch generator.}
Given the extracted RGB features $\mathbf{r}_i$, a global-level patch generator  in conventional SGG \cite{tang2019learning, zellers2018neural, li2021bipartite} would first densely project the $\mathbf{r}_i$ to a lower-dimensional visual representation $\mathbf{h}_i$ as 
\begin{equation}
\label{eq: dense_proj}
\mathbf{h}_i=f_{h2}\left(f_{h1}\left(\mathbf{r}_i\right)\right),
\end{equation} 
where $f_{h1}$ and $f_{h2}$ are two fully-connected layers. This global-level projection (Fig.~\ref{fig:intro}, panel 3) of visual features is not only parameter heavy but can also result in a local-level information loss of entities 
(Fig.~\ref{fig.part}). 

Given the entity details $\left(\mathbf{h}_i, \mathbf{b}_i, \mathbf{c}_i\right)$, conventional SGG then computes a combined entity representation $\mathbf{q}_i$ using another fully-connected network $f_q$ as
\begin{equation}
\label{eq: entity_rep}
\mathbf{q}_i=f_q\left(\mathbf{h}_i \oplus \mathbf{l}_i \oplus \mathbf{w}_i\right),
\end{equation}
where $\mathbf{l}_i$ is a location feature based on the bounding box
$\mathbf{b}_i$, $\mathbf{w}_i$ is a semantic feature based on a word embedding of
its class $\mathbf{c}_i$, and $\oplus$ is the concatenation operation.

To yield a relationship proposal from entity $i$ to $j$, conventional SGG \cite{li2021bipartite} has an additional entity-level projection (Fig.~\ref{fig:intro}, panel 3), comprising convolutional features of the union region of entity bounding boxes $\mathbf{b}_i$ and $\mathbf{b}_j$, denoted as $\mathbf{u}_{ij}$. The predicate representation $ \mathbf{p}_{i \rightarrow j}$ is then computed as
$
    \mathbf{p}_{i \rightarrow j} = f_u\left(\mathbf{u}_{ij}\right) + f_p\left(\mathbf{q}_i \oplus \mathbf{q}_j \right)$,
where $\mathbf{q}_i \oplus \mathbf{q}_j$ denotes the joint representation of entities $e_i$ and $e_j$, and $f_u$, $f_p$ are two fully-connected networks.

\subsection{The VETO entity local-level patch generator}

In contrast to the entity-level patch generator of conventional SGG, the \emph{local-level entity patch generator} of VETO can inculcate \emph{local-level} and geometric cues of entities for relation prediction to learn richer visual representations and reduce crucial information loss during relation prediction. It consists of a two-stage local-level entity patch generator followed by a transformer-based relation encoder and a fully-connected relation decoder. In particular, our network replaces the parameter heavy and computationally expensive fully-connected layers in the conventional relation network resulting from a global entity-level projection (Eq.~\ref{eq: dense_proj}) with less expensive local-level entity projections. 

For a given image $\mathbf{I}$ and its depth-map, we extract the RGB features $\mathbf{r}$ with $c_v$ channels of size $w \times h$ and the geometric features $\mathbf{g}$ with $c_d$ channels and the same size.
 Our relation network  starts with the patch generation and fusion modules, which we call Cross-Relation Patch Generation (CRPG) and Cross-Modality Patch Fusion (CMPF).


\textbf{Cross-Relation Patch Generation module.}
In order to strengthen our transformer-based relation encoder with \emph{local-level} entity feature dependencies, we introduce a Cross-Relation Patch Generation module (Fig.~\ref{fig:veto}). It generates combined subject-object local-level patches.

 We preserve the local-level entity information by  dividing the RGB features  $\mathbf{r}\in \mathbb{R}^{{c}_v\times h\times w}$ and geometric features $\mathbf{g}\in \mathbb{R}^{{c}_d\times h\times w}$ into $p \times p$ blocks and then average pooling $\mathbf{r}$ and $\mathbf{g}$ features within each block to summarize the average presence of the features that are crucial for relation prediction.  Now for a given subject-object entity pair $\left(e_i, e_j\right)$ the resultant pooled RGB features $\mathbf{v}\in \mathbb{R}^{{c}_v\times p\times p}$ 
 from both entities are fused channel-wise (Relation Fusion) as
\begin{equation}
\label{eq. pool_fuse}
    \mathbf{v}_{ij} = C\left(\mathbf{v}_i, \mathbf{v}_j\right) \in \mathbb{R}^{2{c}_v\times p\times p}, \quad \mathbf{v} = \mathrm{pool}\left(\mathbf{r}\right),
\end{equation}
where $C\left(\cdot\right)$ denotes concatenation along the channel dimension and $\mathrm{pool}$ refers to average pooling. We then split $\mathbf{v}_{ij}$ spatially into sequential patches as
\begin{equation}
\label{eq. patches}
\mathbf{v}_{ij}^{\text {pa }}=\left\{\mathbf{v}_{i j}^{\ell} \in \mathbb{R}^{2{c}_v\times k\times k} \;\Big|\; \ell=1,\ldots,\left(\nicefrac{p}{k}\right)^2  \right\},
\end{equation}
where $p$ denotes the pooled width and height of $\mathbf{v}$, $\mathbf{v}^{\ell}$ refers to the ${\ell}$\textsuperscript{th} patch, and $k$ is the patch size (in terms of blocks). Thus, the CRPG module produces  $(\nicefrac{p}{k})^2$ patches for $\mathbf{v}_{ij}$. Similar to $\mathbf{v}_{ij}$, we obtain combined depth features $\mathbf{d}_{ij}\in \mathbb{R}^{2{c}_d\times p\times p}$ and depth patches $\mathbf{d}_{ij}^{\text{pa}} \in \mathbb{R}^{2{c}_d\times k\times k}$ with depth patch size $2{c}_d\times k\times k$
by repeating Eqs.~\eqref{eq. pool_fuse} and \eqref{eq. patches} for the pooled geometric features $\mathbf{d}\in \mathbb{R}^{{c}_d\times p \times p}$.

\textbf{Cross-Modality Patch Fusion module.}
In order to reduce the model parameters and computational expense further and to strengthen the encoder with additional modality cues, we introduce a Cross-Modality Patch Fusion module. It first projects the $\mathbf{v}_{ij}^{\text{pa}}$ and $\mathbf{d}_{ij}^{\text{pa}}$ from the CRPG module to a lower dimensionality: 
\begin{equation}
\label{eq: local_proj}
    \mathbf{px}_{ij}^{\text{pa}} = f_{x}\left(\mathbf{x}_{ij}^{\text{pa}}, p^\mathbf{x}\right), \quad \mathbf{x} \in \{\mathbf{v}, \mathbf{d}\}.
\end{equation}
The resulting $\mathbf{pv}_{ij}^{\text{pa}}$ and $\mathbf{pd}_{ij}^{\text{pa}}$ denote the locally projected entity patches for RGB and depth features, $f_x$ in Eq.~\eqref{eq: local_proj} is a fully-connected network, which should be understood as $f_x\left(\mathbf{y},  M\right)=\mathbf{W}\mathbf{y}$, where $\mathbf{y}\in\mathbb{R}^N$ and $\mathbf{W}\in \mathbb{R}^{M\times N}$.  

Then we fuse the corresponding patches $\mathbf{v}$ and $\mathbf{d}$ of both modalities to capture their dependencies while also further reducing the parameters and computational complexity by reducing the length of the token sequence from $2\left({\nicefrac{p}{k}}\right)^2$ to $\left({\nicefrac{p}{k}}\right)^2$. 
This ensures that dependent modality information is closely knit to be efficiently exploited by the subsequent relation encoder: 
\begin{equation}
\mathbf{z}_0^{\text{pa}} = \left\{\left(\mathbf{pv}_{ij}^\ell \oplus  \mathbf{pd}_{ij}^\ell\right) \;\Big|\; \ell=1,\dots, \left(\nicefrac{p}{k}\right)^2  \right\},
\end{equation}
where $\mathbf{z}_0^{\text{pa}}$ represents the patch-based input embedding tokens to the first layer of our transformer-based relation encoder and  $\oplus$ denotes the concatenation operation. 

Overall, the local-level entity projections in VETO enable capturing the crucial local-level cues that global-level entity projection may overlook while simultaneously reducing the overall number of parameters.

\textbf{Additional cues.} Unlike conventional SGG in which location features $\mathbf{l}$ and semantic features $\mathbf{w}$ are fused with RGB features to form the entity representation, \cf Eq.~\eqref{eq: entity_rep}, we fuse them separately for each subject-object pair and use them as additional input tokens to our relation encoder:
\begin{eqnarray}
    {\mathbf{z}_0^{\left(\nicefrac{p}{k}\right)^2+1}}&=&h\left(f_{l}\left(\mathbf{l}_i \oplus \mathbf{l}_j, {p^{v}+p^{d}}\right)\right)\\
    {\mathbf{z}_0^{\left(\nicefrac{p}{k}\right)^2+2}}&=&h\left(f_{w}\left(\mathbf{w}_i \oplus \mathbf{w}_j, {p^{v}+p^{d}}\right)\right),
\end{eqnarray}
where $f_{l}(\cdot)$ and $f_{w}(\cdot)$ are fully-connected networks and $h(\cdot)$ is a non-linear function.

\textbf{Relation encoder.}
We use a transformer-based relation encoder architecture. The key to successfully capturing the relationship between subject-object pairs in SGG is to adequately express the subject-object joint features that account for their intricate relationship.  The Multi-head Self Attention (MSA) component of the transformer can jointly attend to diverse features by enabling each head to focus on a different subspace with distinct semantic or syntactic meanings. Hence, we propose to capture the subject-object inductive bias in SGG with a MSA component by feeding it embedding tokens $\mathbf{z}_0^{\text{pa}}$ enriched with local-level subject-object information. MSA can be formalized as
$\operatorname{MSA}(Q, K, V)=\operatorname{Concat}\left(\mathrm{SA}_{1}, \mathrm{SA}_{2}, \ldots, \mathrm{SA}_{h}\right) W^{O}$
where $\mathrm{SA}_{i}=\operatorname{softmax}\left(\frac{Q_{i} K_{i}^{T}}{\sqrt{d}}\right) V_{i}$ denotes self-attention, $Q$, $K$, and $V$ refer to query, key, and value, and  $W^{O}$ are trainable parameters.
The input token to the transformer is split into multiple heads to be attended to in parallel.

We prepend the entity local-level information enriched input patches $\mathbf{z}_0^\text{pa}$ from the CMPF module with a learnable class token $\mathbf{z}_{0}^{\text{cl}}$.  Additionally, a learnable 1D positional embedding $\mathbf{z}^{\text{pos}}$ is added to each token to preserve the subject-object positional information. 
The resultant sequence is passed as input to the encoder. 
The transformer consists of $L$ encoder layers, each with MSA and MLP blocks. Each encoder layer also contains a LayerNorm (LN) before every block and residual connections after every block:
\begin{eqnarray}
\label{token}
\mathbf{z}_{0}&=&\left\{\mathbf{z}_{0}^{\text{cl}} ; \mathbf{z}_0^{\text{pa}} ; \mathbf{z}_0^{\left(\nicefrac{p}{k}\right)^2+1} ; \mathbf{z}_0^{\left(\nicefrac{p}{k}\right)^2+2} \right\}+\mathbf{z}^{\text{pos}}\\
\mathbf{z}_{\ell}^{\prime}&=&\operatorname{MSA}\left(\operatorname{LN}\left(\mathbf{z}_{\ell-1}\right)\right)+\mathbf{z}_{\ell-1}, \quad \ell=1 \ldots L\\
\mathbf{z}_{\ell}&=&\operatorname{MLP}\left(\mathrm{LN}\left(\mathbf{z}_{\ell}^{\prime}\right)\right)+\mathbf{z}_{\ell}^{\prime} , \quad \ell=1 \ldots L\\
\label{prediction}
\mathbf{y}&=&\mathrm{LN}\left(\mathbf{z}_{L}^{0}\right),
\end{eqnarray}
where $\mathbf{z}^{\text{pos}} \in \mathbb{R}^{(\left(\nicefrac{p}{k}\right)^2+3) \times \left({p^{v}+p^{d}}\right)}$, and $L$ denotes the number of total encoder layers. Each encoder layer contains MSA, MLP, and LN blocks as well as residual connections.
Finally, $\mathbf{y}$ is linearly projected to the total number of predicate relations, forming the relation classification head.
\section{MEET: Mutually Exclusive ExperT learning}

As pointed out by Dong \etal~\cite{dong2022stacked}, a single classifier cannot achieve satisfactory performance on all the predicate classes due to the extreme unbalance of SGG datasets. To tackle this issue, they propose to deploy multiple classifiers. However, their classifiers are not mutually exclusive and result in a significant deterioration of head class performance (\enquote{GCL} in Fig.~\ref{fig.bias}), as they distill knowledge from head to tail classifiers.
We hypothesise that learning mutually exclusive experts, each responsible for a predicate subgroup, can reduce model bias towards a specific set of predicates.
 

Therefore, we propose Mutually Exclusive-ExperT learning (MEET). MEET splits the predicate classes into balanced groups based on the predicate frequency in the training set.
First, we sort the predicate classes according to their frequency in descending order into $\mathcal{P}_\text{sort} = \{p_i\}_{i=1}^{M}$. The sorted set is split into $G$ predicate groups $\{\mathcal{P}_\text{sort}^g\}_{g=1}^{G}$. We use the same class split as \cite{dong2022stacked}, yet in contrast, MEET deploys mutually exclusive classifier experts $\{\mathbf{E}^{g}\}_{g=1}^{G}$ responsible for classification within each predicate group $\mathcal{P}_\text{sort}^g$. 

\begin{algorithm}[t]
\caption{MEET: Mutually Exclusive ExperT learning}
\label{alg:1}
\begin{algorithmic}[1]
\State Input: predicate classes $\mathcal{P}=\left\{{p_i}\right\}_{i=1}^{M}$, experts $\mathbf{E} =\left\{{\mathbf{E}}\right\}_{g=1}^{G}$, where $G$ denotes the total number of experts
\State$\mathcal{P}_\text{sort} \gets$ sorted predicate set $\mathcal{P}$
\State$\mathcal{P}^{g}_\text{sort} \gets$ $g$\textsuperscript{th} sorted predicate group of $\mathcal{P}_\text{sort}$
\State $\mathcal{S}^g \gets$ relation representation sample $\mathbf{z}_{L}^{0}$ of $g$\textsuperscript{th} group
\State $f_s\left(x, y \right) = \max\big(\min\big(\nicefrac{x}{y}, 1.0 \big), 0.01\big)$

\For {$g=1$ to ${G}$}
    \State $\mathit{Index} \gets$ index of the central predicate in $\mathcal{P}_\text{sort}^{g}$
    \State $\mathit{centre} \gets \text{Freq}(\mathcal{P}_\text{sort}^{g}(\mathit{Index}))$
    \State $I_\text{dist}^{g} = \left\{f_s\left( \mathit{centre}, \text{Freq}\left(p\right)\right) \mid \forall p \in \mathcal{P}_\text{sort}^{g}\right\}$
    \State $O_\text{dist}^{g} = \left\{f_s\left( \mathit{centre}, \text{Freq}\left(p\right)\right)  \mid \forall p \notin \mathcal{P}_\text{sort}^g\right\}$
    \State $\mathbf{y}^g(p) = \left\{\begin{array}{cl}
\text{Index}\left(p\right) \text { in } \mathcal{P}_\text{sort}^{g}, & \text { for } p \in \mathcal{P}_\text{sort}^{g} \\
\left|\mathcal{P}_\text{sort}^{g}\right| + 1, & \text { for } p \notin \mathcal{P}_\text{sort}^{g}
\end{array}\right.$
    \State Sample from distribution: $\mathcal{S}_{\text{in}}^{g} \sim I_\text{dist}^{g}, \mathcal{S}_{\text{out}}^{g} \sim O_\text{dist}^{g}$
    \State $\mathbf{w}^{g} = \mathbf{E}^g\big(\left\{\mathcal{S}_{\text{in}}^{g}; \mathcal{S}_{\text{out}}^{g}\right\}\big)$ \Comment{expert output}
\EndFor
\State $\mathcal{L} = \sum_{g=1}^{G}\mathcal{L}_{CE}\left(\mathbf{w}^g, \mathbf{y}^g\right) $ \Comment{multi-expert loss}\vspace{2mm}

\State \textbf{Evaluation stage:}
\State $\hat{\mathbf{w}}^g = \big[\mathbf{w}^{g}_i\big]_{i=1}^{|\mathcal{P}_\text{sort}^{g}|}$ \Comment{discard OOD predictions}
\State $\mathcal{R}_\text{conf} = \{\max_i \hat{\mathbf{w}}^g_i\}_{g=1}^G$ 
\State $\mathcal{R}_\text{label} = \Big\{\mathcal{P}_\text{sort}\left(\arg\max_i \hat{\mathbf{w}}^{g}_i + \sum_{j=0}^{g-1}\big|\mathcal{P}_\text{sort}^{j}\big|\right)\Big\}_{g=1}^G $ 

\end{algorithmic}

\end{algorithm}

Unfortunately, training each expert exclusively on a subgroup of predicates $\mathcal{P}_\text{sort}^g$ can be challenged during the evaluation stage with samples out of its classification space, resulting in uncertain predictions. To overcome this issue, we train out-of-distribution aware experts. We summarize MEET in Algorithm \ref{alg:1}. \textbf{(Lines 9--10)}: During training, we adjust the in-distribution  $I_\text{dist}^{g}$ and out-of-distribution $O_\text{dist}^{g}$ sampling frequency within each expert to prevent the experts from being overwhelmed with OOD samples.  $\text{Freq}\left(\cdot\right)$ denotes the frequency count of a predicate class. \textbf{(Line 11)}: We re-map the relation labels to accommodate an out-of-distribution pseudo-label for every expert group. \textbf{(Lines 12, 13, 15)}: For a given image $\mathbf{I}$, each expert $\mathbf{E}^{g}$ is trained on the in-distribution and out-of-distribution samples $\mathcal{S}_\text{in}^{g}$ and $\mathcal{S}_\text{out}^{g}$, respectively. \textbf{(Lines 17--19)}: During the evaluation stage, we discard the OOD predictions from each group and map the prediction labels back to the original labels.


\section{Experimental Evaluation}
We aim to answer the following questions: 
\textbf{(Q1)}~Does VETO + MEET improve the SOTA in unbiased SGG?
\textbf{(Q2)}~What is the impact of MEET on other SGG methods?
\textbf{(Q3)}~Does local-level projection reduce the model size?
\textbf{(Q4)}~Does SGG benefit from local-level patch generation? 
\textbf{(Q5)}~Does a depth map improve SGG performance?

\begin{table*}
    \centering
    \caption{\textbf{Recall (R), mean Recall (mR), and their average (A) on VG} (the higher, the better). Colors in the table vary from blue to green to depict the performance improvement. `+' denotes the combination with a model-agnostic unbiasing strategy. Double citations refer to the original model and its reproduced variant on a ResNeXt-101-FPN backbone. The superscript `†' denotes the method uses Faster-RCNN with VGG-16 as the object detector.}
    \vspace{-1mm}
    \begin{adjustbox}{max width=\linewidth, center}
    \begin{tabular}{l|ccc|ccc|ccc}
    
    \toprule
    \multicolumn{1}{c|}{\multirow{2}{*}{\textbf{Mode}l}} & \multicolumn{3}{c|}{ \textbf{PredCls} } & \multicolumn{3}{c|}{ \textbf{SGCls} } & \multicolumn{3}{c}{ \textbf{SGDet} }\\ 
    
    \cmidrule(lr){2-4}\cmidrule(lr){5-7}\cmidrule(lr){8-10}
    
    \multicolumn{1}{c|}{} & \multicolumn{1}{c}{ \textbf{R@k:}\text{ } \textbf{50 / 100} } & \multicolumn{1}{c}{ \textbf{mR@k:}\text{ } \textbf{50 / 100}} & \multicolumn{1}{c|}{ \textbf{A@k:} \text{ } \textbf{50 / 100} } & \multicolumn{1}{c}{ \textbf{R@k:} \text{ } \textbf{50 / 100} } & \multicolumn{1}{c}{ \textbf{mR@k:} \text{ } \textbf{50 / 100}} & \multicolumn{1}{c|}{ \textbf{A@k:} \text{ } \textbf{50 / 100} } & \multicolumn{1}{c}{ \textbf{R@k:} \text{ } \textbf{50 / 100} } & \multicolumn{1}{c}{ \textbf{mR@k:} \text{ } \textbf{50 / 100}} & \multicolumn{1}{c}{ \textbf{A@k:} \text{ } \textbf{50 / 100}} \\ 
    
    \midrule 
    
     IMP~\cite{suhail2021energy, dong2022stacked} & \gradient{61.1} /\gradient{63.1}  & \gradientpmr{11.0} / \gradientpmr{11.8} & \gradientpm{36.1} / \gradientpm{37.4} & \gradientc{37.4} / \gradientc{38.3} & \gradientcmr{6.4}\text{ } / \gradientcmr{6.7}\text{ } & \gradientcm{21.9} / \gradientcm{22.5} & \gradientd{23.6} / \gradientd{28.7} & \gradientdmr{3.3} \text{ } / \gradientdmr{4.1} \text{ } & \gradientdm{13.5} / \gradientdm{16.4} \\
    
    KERN\textsuperscript{†}~\cite{chen2019knowledge} & \gradient{65.8} /\gradient{67.6}  & \gradientpmr{17.7} / \gradientpmr{19.2} & \gradientpm{41.8} / \gradientpm{43.4} & \gradientc{36.7} / \gradientc{37.4} & \gradientcmr{9.4} / \gradientcmr{10.0} & \gradientcm{23.1} / \gradientcm{23.7} & \gradientd{27.1} / \gradientd{29.8} & \gradientdmr{6.4}\text{ } / \gradientdmr{7.3} \text{ } & \gradientdm{16.8} / \gradientdm{18.6} \\
    
    GB-Net + Rwt\textsuperscript{†}~\cite{zareian2020bridging} & \gradient{66.6} /\gradient{68.2}  & \gradientpmr{22.1} / \gradientpmr{24.0} & \gradientpm{44.4} / \gradientpm{46.1} & \gradientc{37.3} / \gradientc{38.0} & \gradientcmr{12.7} / \gradientcmr{13.4} & \gradientcm{25.0} / \gradientcm{25.7} & \gradientd{26.3} / \gradientd{29.9} & \gradientdmr{7.1} \text{ } / \gradientdmr{8.5} \text{ } & \gradientdm{16.7} / \gradientdm{18.5} \\
    
    DT2-ACBS~\cite{desai2021learning} & \gradient{23.3} /\gradient{25.6}  & \gradientpmr{35.9} / \gradientpmr{39.7} & \gradientpm{29.6} / \gradientpm{32.7} & \gradientc{16.2} / \gradientc{17.6} & \gradientcmr{24.8} / \gradientcmr{27.5} & \gradientcm{20.5} / \gradientcm{22.6} & \gradientd{15.0} / \gradientd{16.3} & \gradientdmr{22.0} / \gradientdmr{24.0} & \gradientdm{18.5} / \gradientdm{20.2} \\
    
    PCPL\textsuperscript{†}~\cite{yan2020pcpl} & \gradient{50.8} /\gradient{52.6}  & \gradientpmr{35.2} / \gradientpmr{37.8} & \gradientpm{43.0} / \gradientpm{45.2} & \gradientc{27.6} / \gradientc{28.4} & \gradientcmr{18.6} / \gradientcmr{19.6} & \gradientcm{23.1} / \gradientcm{24.0} & \gradientd{14.6} / \gradientd{18.6} & \gradientdmr{9.5} / \gradientdmr{11.7} & \gradientdm{12.1} / \gradientdm{15.2} \\
    
    GPS-Net \cite{dong2022stacked, lin2020gps} & \gradient{65.2} /\gradient{67.1}  & \gradientpmr{15.2} / \gradientpmr{16.6} & \gradientpm{40.2} / \gradientpm{41.9} & \gradientc{37.8} / \gradientc{39.2} & \gradientcmr{8.5} \text{ } / \gradientcmr{9.1} \text{ }& \gradientcm{23.2} / \gradientcm{24.2} & \gradientd{31.1} / \gradientd{35.9} & \gradientdmr{6.7} \text{ } / \gradientdmr{8.6} \text{ } & \gradientdm{18.9} / \gradientdm{22.2} \\
    
    SG-CogTree \cite{yu2020cogtree} & \gradient{38.4} / \gradient{39.7} & \gradientpmr{28.4} / \gradientpmr{31.0} & \gradientpm{33.4} / \gradientpm{35.3} & \gradientc{22.9} / \gradientc{23.4} & \gradientcmr{15.7} / \gradientcmr{16.7} & \gradientcm{19.3} / \gradientcm{20.1} & \gradientd{19.5} / \gradientd{21.7} & \gradientdmr{11.1} / \gradientdmr{12.7} & \gradientdm{16.8} / \gradientdm{17.2}  \\

    BGNN \cite{li2021bipartite} &  \gradient{59.2} /  \gradient{61.3} & \gradientpmr{30.4} / \gradientpmr{32.9} & \gradientpm{44.8} / \gradientpm{47.1} & \gradientc{37.4} / \gradientc{38.5} & \gradientcmr{14.3} / \gradientcmr{16.5} & \gradientcm{25.9} / \gradientcm{27.5} & \gradientd{31.0} / \gradientd{35.8} & \gradientdmr{10.7} / \gradientdmr{12.6} & \gradientdm{20.9} / \gradientdm{24.2}  \\
    \midrule
    VTransE \cite{tang2020unbiased, zhang2017visual} & \gradient{65.7} / \gradient{67.6} & \gradientpmr{14.7} / \gradientpmr{15.8} & \gradientpm{40.2} / \gradientpm{41.7} & \gradientc{38.6} / \gradientc{39.4} & \gradientcmr{8.2} / \gradientcmr{8.7} & \gradientcm{23.4} / \gradientcm{24.1} & \gradientd{29.7} / \gradientd{34.3} & \gradientdmr{5.0}\text{ } / \gradientdmr{6.0}\text{ } & \gradientdm{17.3} / \gradientdm{20.2}  \\
    
    VTransE + TDE \cite{tang2020unbiased} & \gradient{43.1} / \gradient{48.5} & \gradientpmr{24.6} / \gradientpmr{28.0} & \gradientpm{33.9} / \gradientpm{38.3} & \gradientc{25.7} / \gradientc{28.5} & \gradientcmr{12.9} / \gradientcmr{14.8} & \gradientcm{19.3} / \gradientcm{21.7} & \gradientd{18.7} / \gradientd{22.6} & \gradientdmr{8.6}\text{ } / \gradientdmr{10.5} & \gradientdm{13.7} / \gradientdm{16.7} \\
    
    VTransE + GCL \cite{dong2022stacked} & \gradient{35.4} / \gradient{37.3} & \gradientpmr{34.2} / \gradientpmr{36.3} & \gradientpm{34.8} / \gradientpm{36.8} & \gradientc{25.8} / \gradientc{26.9} & \gradientcmr{20.5} / \gradientcmr{21.2} & \gradientcm{22.8} / \gradientcm{23.7} & \gradientd{14.6} / \gradientd{17.1} & \gradientdmr{13.6} / \gradientdmr{15.5} & \gradientdm{14.1} / \gradientdm{16.3}\\
    
     VTransE  + \textbf{MEET} (ours) & \gradient{58.3} / \gradient{64.9} & \gradientpmr{18.3} / \gradientpmr{24.9} & \gradientpm{38.3} / \gradientpm{44.9} & \gradientc{35.8} / \gradientc{39.1} & \gradientcmr{12.8} / \gradientcmr{16.7} & \gradientcm{24.3} / \gradientcm{27.9} & \gradientd{22.0} / \gradientd{27.6} & \gradientdmr{5.8}\text{ } / \gradientdmr{7.6}\text{ } & \gradientdm{13.9} / \gradientdm{17.6}\\
     
     \midrule
    
    Motifs \cite{tang2020unbiased, zellers2018neural} & \gradient{65.2} / \gradient{67.0} & \gradientpmr{14.8} / \gradientpmr{16.1} & \gradientpm{40.0} / \gradientpm{41.6} & \gradientc{38.9} / \gradientc{39.8} & \gradientcmr{8.3} / \gradientcmr{8.8} & \gradientcm{23.6} / \gradientcm{24.3} & \gradientd{32.8} / \gradientd{37.2} & \gradientdmr{6.8}\text{ } / \gradientdmr{7.9}\text{ } &  \gradientdm{19.8} / \gradientdm{22.6} \\
    
    Motifs + Rwt \cite{chiou2021recovering} &\gradient{54.7} / \gradient{56.5} & \gradientpmr{17.3} / \gradientpmr{18.6} & \gradientpm{36.0} / \gradientpm{37.6} & \gradientc{29.5} / \gradientc{31.5} & \gradientcmr{11.2} / \gradientcmr{11.7} & \gradientcm{20.4} / \gradientcm{21.6} & \gradientd{24.4} / \gradientd{29.3} & \gradientdmr{9.2}\text{ } / \gradientdmr{10.9} & \gradientdm{16.8} / \gradientdm{20.1} \\
    
    Motifs + TDE \cite{tang2020unbiased} & \gradient{46.2} / \gradient{51.4} & \gradientpmr{25.5} / \gradientpmr{29.1} & \gradientpm{35.9} / \gradientpm{40.3} & \gradientc{27.7} / \gradientc{29.9} & \gradientcmr{13.1} / \gradientcmr{14.9} & \gradientcm{20.4} / \gradientcm{22.4} & \gradientd{16.9} / \gradientd{20.3} & \gradientdmr{8.2}\text{ } / \gradientdmr{9.8}\text{ } & \gradientdm{12.5} / \gradientdm{15.1}\\
    
    Motifs + PCPL \cite{dong2022stacked, yan2020pcpl} & \gradient{54.7} / \gradient{56.5} & \gradientpmr{17.3} / \gradientpmr{18.6} & \gradientpm{36.0} / \gradientpm{37.6} & \gradientc{29.5} / \gradientc{31.5} & \gradientcmr{11.2} / \gradientcmr{11.7} & \gradientcm{20.4} / \gradientcm{21.6} & \gradientd{24.4} / \gradientd{29.3} & \gradientdmr{9.2} \text{ }/ \gradientdmr{10.9} & \gradientdm{16.8} / \gradientdm{20.1} \\
    
    Motifs + CogTree \cite{yu2020cogtree} & \gradient{35.6} / \gradient{36.8} & \gradientpmr{26.4} / \gradientpmr{29.0} & \gradientpm{31.0} / \gradientpm{32.9} & \gradientc{21.6} / \gradientc{22.2} & \gradientcmr{14.9} / \gradientcmr{16.1} & \gradientcm{18.3} / \gradientcm{19.2} & \gradientd{20.0} / \gradientd{22.1} & \gradientdmr{10.4} / \gradientdmr{11.8} & \gradientdm{15.2} / \gradientdm{17.0} \\
    
    Motifs + DLFE \cite{chiou2021recovering} & \gradient{52.5} / \gradient{54.2} & \gradientpmr{26.9} / \gradientpmr{28.8} & \gradientpm{39.7} / \gradientpm{41.5} & \gradientc{32.3} / \gradientc{33.1} & \gradientcmr{15.2} / \gradientcmr{15.9} & \gradientcm{23.8} / \gradientcm{24.5} & \gradientd{25.4} / \gradientd{29.4} & \gradientdmr{11.7} / \gradientdmr{13.8} & \gradientdm{18.6} / \gradientdm{21.6} \\
    
    Motifs + EMB \cite{suhail2021energy} & \gradient{65.2} / \gradient{67.3} & \gradientpmr{18.0} / \gradientpmr{19.5} & \gradientpm{41.6} / \gradientpm{43.4} & \gradientc{39.2} / \gradientc{40.0} & \gradientcmr{10.2} / \gradientcmr{11.0} & \gradientcm{24.7} / \gradientcm{25.5} & \gradientd{31.7} / \gradientd{36.3} & \gradientdmr{7.7}\text{ } / \gradientdmr{9.3}\text{ } & \gradientdm{19.7} / \gradientdm{22.8} \\
    
    Motifs + GCL \cite{dong2022stacked} & \gradient{42.7} / \gradient{44.4} & \gradientpmr{36.1} / \gradientpmr{38.2} & \gradientpm{39.4} / \gradientpm{41.3} & \gradientc{26.1} / \gradientc{27.1} & \gradientcmr{20.8} / \gradientcmr{21.8} & \gradientcm{23.5} / \gradientcm{24.5} & \gradientd{18.4} / \gradientd{22.0} & \gradientdmr{16.8} / \gradientdmr{19.3} & \gradientdm{17.6} / \gradientdm{20.7} \\
    
    Motifs + IETrans + Rwt \cite{zhang2022fine} & \gradient{48.6} / \gradient{50.5} & \gradientpmr{35.8} / \gradientpmr{39.1} & \gradientpm{42.2} / \gradientpm{44.8} & \gradientc{29.4} / \gradientc{30.2} & \gradientcmr{21.5} / \gradientcmr{22.8} & \gradientcm{25.5} / \gradientcm{26.5} & \gradientd{23.5} / \gradientd{27.2} & \gradientdmr{15.5} / \gradientdmr{18.0} & \gradientdm{19.5} / \gradientdm{22.6} \\
    
    Motifs + \textbf{MEET} (ours) & \gradient{67.4} / \gradient{72.7} & \gradientpmr{25.3} / \gradientpmr{33.5} & \gradientpm{46.4} / \gradientpm{53.1} & \gradientc{40.5} / \gradientc{43.2} & \gradientcmr{19.0} / \gradientcmr{23.7} & \gradientcm{29.8} / \gradientcm{33.5} & \gradientd{27.9} / \gradientd{33.3} & \gradientdmr{8.5}\text{ } / \gradientdmr{11.8} & \gradientdm{18.2} / \gradientdm{22.6} \\
    
    \midrule
    
    VCTree \cite{tang2020unbiased, tang2019learning} & \gradient{65.4} / \gradient{67.2} & \gradientpmr{16.7} / \gradientpmr{18.2} & \gradientpm{41.1} / \gradientpm{42.7} & \gradientc{46.7} / \gradientc{47.6} & \gradientcmr{11.8} / \gradientcmr{12.5} & \gradientcm{29.3} / \gradientcm{30.1} & \gradientd{31.9} / \gradientd{36.2} & \gradientdmr{7.4}\text{ } / \gradientdmr{8.7}\text{ } & \gradientdm{19.7} / \gradientdm{22.5} \\
    
    VCTree + Rwt \cite{chiou2021recovering} & \gradient{60.7} / \gradient{62.6} & \gradientpmr{19.4} / \gradientpmr{20.4} & \gradientpm{40.1} / \gradientpm{41.5} & \gradientc{42.3} / \gradientc{43.5} & \gradientcmr{12.5} / \gradientcmr{13.1} & \gradientcm{27.4} / \gradientcm{28.3} & \gradientd{27.8} / \gradientd{32.0} & \gradientdmr{8.7}\text{ } / \gradientdmr{10.1} & \gradientdm{18.3} /\gradientdm{21.1} \\
    
    VCTree + TDE \cite{tang2020unbiased} & \gradient{47.2} / \gradient{51.6} & \gradientpmr{25.4} / \gradientpmr{28.7} & \gradientpm{36.3} / \gradientpm{40.2} & \gradientc{25.4} / \gradientc{27.9} & \gradientcmr{12.2} / \gradientcmr{14.0} & \gradientcm{18.8} / \gradientcm{21.0} & \gradientd{19.4} / \gradientd{23.2} & \gradientdmr{9.3}\text{ } / \gradientdmr{11.1} & \gradientdm{14.5} / \gradientdm{17.2} \\
    
    VCTree + PCPL \cite{dong2022stacked, yan2020pcpl} & \gradient{56.9} / \gradient{58.7} & \gradientpmr{22.8} / \gradientpmr{24.5} & \gradientpm{39.9} / \gradientpm{41.6} & \gradientc{40.6} / \gradientc{41.7} & \gradientcmr{15.2} / \gradientcmr{16.1} & \gradientcm{27.9} / \gradientcm{28.9} & \gradientd{26.6} / \gradientd{30.3} & \gradientdmr{10.8} / \gradientdmr{12.6} & \gradientdm{18.4} / \gradientdm{21.5} \\
    
    VCTree + CogTree \cite{yu2020cogtree} & \gradient{44.0} / \gradient{45.4} & \gradientpmr{27.6} / \gradientpmr{29.7} & \gradientpm{35.8} / \gradientpm{37.6} & \gradientc{30.9} / \gradientc{31.7} & \gradientcmr{18.8} / \gradientcmr{19.9} & \gradientcm{24.9} / \gradientcm{25.8} & \gradientd{18.2} / \gradientd{20.4} & \gradientdmr{10.4} / \gradientdmr{12.1} & \gradientdm{14.3} / \gradientdm{16.3} \\
    
     VCTree + DLFE \cite{chiou2021recovering} & \gradient{51.8} / \gradient{53.5} & \gradientpmr{25.3} / \gradientpmr{27.1} & \gradientpm{38.6} / \gradientpm{40.3} & \gradientc{33.5} / \gradientc{34.6} & \gradientcmr{18.9} / \gradientcmr{20.0} & \gradientcm{26.2} / \gradientcm{27.3} & \gradientd{22.7} / \gradientd{26.3} & \gradientdmr{11.8} / \gradientdmr{13.8} & \gradientdm{17.5} / \gradientdm{20.1} \\
     
     VCTree + EMB \cite{suhail2021energy} & \gradient{64.0} / \gradient{65.8} & \gradientpmr{18.2} / \gradientpmr{19.7} & \gradientpm{41.1} / \gradientpm{42.8} & \gradientc{44.7}/ \gradientc{45.8} & \gradientcmr{12.5} / \gradientcmr{13.5} & \gradientcm{28.6} / \gradientcm{30.0} & \gradientd{31.4} / \gradientd{35.9} & \gradientdmr{7.7}\text{ } / \gradientdmr{9.1}\text{ } & \gradientdm{19.5} / \gradientdm{22.5} \\
     
     VCTree + GCL \cite{dong2022stacked} & \gradient{40.7} / \gradient{42.7} & \gradientpmr{37.1} / \gradientpmr{39.1} & \gradientpm{38.9} / \gradientpm{40.1} & \gradientc{27.7} / \gradientc{28.7} & \gradientcmr{22.5} / \gradientcmr{23.5} & \gradientcm{25.1} / \gradientcm{26.1} & \gradientd{17.4} / \gradientd{20.7} & \gradientdmr{15.2} / \gradientdmr{17.5} & \gradientdm{16.3} / \gradientdm{19.1} \\
     
     VCTree + IETrans + Rwt \cite{chiou2021recovering} & \gradient{48.0} / \gradient{49.9} & \gradientpmr{37.0} / \gradientpmr{39.7} & \gradientpm{42.5} / \gradientpm{43.5} & \gradientc{30.0} / \gradientc{30.9} & \gradientcmr{19.9} / \gradientcmr{21.8} & \gradientcm{25.0} / \gradientcm{26.4} & \gradientd{23.6} / \gradientd{27.8} & \gradientdmr{12.0} / \gradientdmr{14.9} & \gradientdm{17.8} / \gradientdm{21.4}  \\
     
     VCTree + \textbf{MEET} (ours) & \gradient{62.0} / \gradient{69.8} & \gradientpmr{25.5} / \gradientpmr{34.5} & \gradientpm{43.8} / \gradientpm{52.2} & \gradientc{35.4} / \gradientc{39.2} & \gradientcmr{14.5} / \gradientcmr{18.6} & \gradientcm{25.0} / \gradientcm{28.9} & \gradientd{26.4} / \gradientd{31.2} & \gradientdmr{8.2}\text{ } / \gradientdmr{11.5} & \gradientdm{17.3} / \gradientdm{21.4} \\
     
     \midrule
     
     
     SHA \cite{dong2022stacked} & \gradient{64.3} / \gradient{66.4} & \gradientpmr{18.8} / \gradientpmr{20.5} & \gradientpm{41.5} / \gradientpm{43.5} & \gradientc{38.0} / \gradientc{39.0} & \gradientcmr{10.9} / \gradientcmr{11.6} & \gradientcm{24.5} / \gradientcm{25.3} & \gradientd{30.6} / \gradientd{34.9} & \gradientdmr{7.8}\text{ } / \gradientdmr{9.1}\text{ } & \gradientdm{19.2} / \gradientdm{22.0} \\
     
     SHA + GCL \cite{dong2022stacked} & \gradient{35.1} / \gradient{37.2} & \gradientpmr{41.6} / \gradientpmr{44.1} & \gradientpm{38.4} / \gradientpm{40.7} & \gradientc{22.8} / \gradientc{23.9} & \gradientcmr{23.0} / \gradientcmr{24.3} & \gradientcm{22.9} / \gradientcm{24.1} & \gradientd{14.9} / \gradientd{18.2} & \gradientdmr{17.9} / \gradientdmr{20.9} & \gradientdm{16.4} / \gradientdm{19.6} \\
     
     SHA + \textbf{MEET} (ours) & \gradient{66.3} / \gradient{72.4} & \gradientpmr{28.0} / \gradientpmr{36.2} & \gradientpm{47.2} / \gradientpm{54.3} & \gradientc{37.9} / \gradientc{41.2} & \gradientcmr{16.1} / \gradientcmr{20.7} & \gradientcm{27.0} / \gradientcm{31.0} & \gradientd{24.2} / \gradientd{29.7} & \gradientdmr{7.7}\text{ } / \gradientdmr{9.8}\text{ } & \gradientdm{16.0} / \gradientdm{19.8} \\
    
    \midrule
    
    \textbf{VETO} (ours) & \gradient{64.2} / \gradient{66.3} & \gradientpmr{22.8} / \gradientpmr{24.7} & \gradientpm{43.5} / \gradientpm{45.5} & \gradientc{35.7} / \gradientc{36.9} & \gradientcmr{11.1} / \gradientcmr{11.9} & \gradientcm{23.4} / \gradientcm{24.4} & \gradientd{27.5} / \gradientd{31.5} & \gradientdmr{8.1} / \gradientdmr{9.5} & \gradientdm{17.8} / \gradientdm{20.5} \\
    
    \textbf{VETO} (ours) + Rwt & \gradient{61.9} / \gradient{63.9} & \gradientpmr{33.1} / \gradientpmr{35.1} & \gradientpm{47.5} / \gradientpm{49.5} & \gradientc{35.1} / \gradientc{36.3} & \gradientcmr{16.1} / \gradientcmr{17.1} & \gradientcm{25.6} / \gradientcm{26.7} & \gradientd{26.2} / \gradientd{30.4} & \gradientdmr{10.0} / \gradientdmr{11.7} & \gradientdm{18.1} / \gradientdm{21.1} \\
    
    
    \textbf{VETO + MEET} (ours) & \gradient{74.0} / \gradient{78.9} & \gradientpmr{42.0} / \gradientpmr{52.4} & \gradientpm{58.0} / \gradientpm{65.7} & \gradientc{41.1} / \gradientc{44.0} & \gradientcmr{22.3} / \gradientcmr{27.4} & \gradientcm{31.7} / \gradientcm{35.7} & \gradientd{28.6} / \gradientd{34.0} & \gradientdmr{10.6} / \gradientdmr{13.8} & \gradientdm{19.6} / \gradientdm{23.9} \\

    \bottomrule
    \end{tabular}
    \end{adjustbox}
    \label{tab:vg}
    \vspace{-0.5em}
\end{table*}

\begin{table*}
    \centering
    \caption{\textbf{Recall (R), mean Recall (mR), and their average (A) on GQA} (the higher, the better). 
    Conventions as described in Tab.~\ref{tab:vg}.} 
    \vspace{-1mm}
    \begin{adjustbox}{max width=\linewidth, center}
    \begin{tabular}{l|ccc|ccc|ccc}
    
    \toprule
    \multicolumn{1}{c|}{\multirow{2}{*}{\textbf{Mode}l}} & \multicolumn{3}{c|}{ \textbf{PredCls} } & \multicolumn{3}{c|}{ \textbf{SGCls} } & \multicolumn{3}{c}{ \textbf{SGDet} }\\ 
    
    \cmidrule(lr){2-4}\cmidrule(lr){5-7}\cmidrule(lr){8-10}
    
    \multicolumn{1}{c|}{} & \multicolumn{1}{c}{ \textbf{R@k:}\text{ } \textbf{50/100} } & \multicolumn{1}{c}{ \textbf{mR@k:}\text{ } \textbf{50 / 100}} & \multicolumn{1}{c|}{ \textbf{A@k:}\text{ } \textbf{50/100} } &  \multicolumn{1}{c}{ \textbf{R@k:}\text{ } \textbf{50/100} } & \multicolumn{1}{c|}{ \textbf{mR@k:}\text{ } \textbf{50 / 100}} & \multicolumn{1}{c|}{ \textbf{A@k:}\text{ } \textbf{50/100} } & \multicolumn{1}{c}{ \textbf{R@k:}\text{ } \textbf{50/100} } & \multicolumn{1}{c|}{ \textbf{mR@k:}\text{ } \textbf{50 / 100}} & \multicolumn{1}{c}{ \textbf{A@k:}\text{ } \textbf{50/100} } \\ 
    
    \midrule
    VTransE \cite{dong2022stacked, zhang2017visual} &  \gradientg{55.7} / \gradientg{57.9} & \gradientpmrg{14.0} / \gradientpmrg{15.0} & \gradientpmg{34.9} / \gradientpmg{36.5} & \gradientcg{33.4} / \gradientcg{34.2} & \gradientcmrg{8.1}\text{ } / \gradientcmrg{8.7}\text{ } & \gradientcmg{20.9} / \gradientcmg{21.5} & \gradientdg{27.2}/\gradientdg{30.7} & \gradientdmrg{5.8}\text{ } / \gradientdmrg{6.6}\text{ } & \gradientdmg{16.5} / \gradientdmg{18.7} \\
    VTransE + GCL \cite{dong2022stacked} & \gradientg{35.5} / \gradientg{37.4} & \gradientpmrg{30.4} / \gradientpmrg{32.3} & \gradientpmg{33.0} / \gradientpmg{34.9} & \gradientcg{22.9} / \gradientcg{23.6} & \gradientcmrg{16.6} / \gradientcmrg{17.4} & \gradientcmg{19.8} / \gradientcmg{20.5} & \gradientdg{15.3} / \gradientdg{18.0} & \gradientdmrg{14.7} / \gradientdmrg{16.4} & \gradientdmg{15.0} / \gradientdmg{17.2}\\
    VTransE + \textbf{MEET} (ours) & \gradientg{55.4} / \gradientg{60.9} & \gradientpmrg{15.2} / \gradientpmrg{21.7} & \gradientpmg{35.3} / \gradientpmg{41.3} & \gradientcg{28.1} / \gradientcg{30.7} & \gradientcmrg{9.2} / \gradientcmrg{12.0} & \gradientcmg{18.7} / \gradientcmg{21.4} & \gradientdg{24.0} / \gradientdg{27.9} & \gradientdmrg{5.6} / \gradientdmrg{7.8} & \gradientdmg{14.8} / \gradientdmg{17.9}\\
     \midrule
     
    Motifs\cite{dong2022stacked, zellers2018neural} & \gradientg{65.3} / \gradientg{66.8} &  \gradientpmrg{16.4} / \gradientpmrg{17.1} & \gradientpmg{40.9} / \gradientpmg{42.0} & \gradientcg{34.2} / \gradientcg{34.9} & \gradientcmrg{8.2} / \gradientcmrg{8.6} & \gradientcmg{21.2} / \gradientcmg{21.9} & \gradientdg{28.9} / \gradientdg{33.1} & \gradientdmrg{6.4} / \gradientdmrg{7.7} & \gradientdmg{17.7} / \gradientdmg{20.4} \\
    Motifs + GCL\cite{dong2022stacked} & \gradientg{44.5} / \gradientg{46.2} & \gradientpmrg{36.7} / \gradientpmrg{38.1} & \gradientpmg{40.6} / \gradientpmg{42.2} & \gradientcg{23.2} / \gradientcg{24.0} & \gradientcmrg{17.3} / \gradientcmrg{18.1} & \gradientcmg{20.3} / \gradientcmg{21.1} & \gradientdg{18.5} / \gradientdg{21.8} & \gradientdmrg{16.8} / \gradientdmrg{18.8} &  \gradientdmg{17.7} / \gradientdmg{20.3}\\
    Motifs +\textbf{MEET} (ours) & \gradientg{63.6} / \gradientg{68.4} & \gradientpmrg{25.0} / \gradientpmrg{30.4} & \gradientpmg{44.3} / \gradientpmg{49.4} & \gradientcg{33.7} / \gradientcg{36.1} & \gradientcmrg{17.3} / \gradientcmrg{19.9} & \gradientcmg{25.5} / \gradientcmg{28.0} & \gradientdg{26.8} / \gradientdg{30.8} & \gradientdmrg{9.3}\text{ } / \gradientdmrg{12.5} &  \gradientdmg{18.1} / \gradientdmg{21.7}\\
    \midrule
    
    VCTree\cite{dong2022stacked, tang2019learning} & \gradientg{63.8} / \gradientg{65.7} & \gradientpmrg{16.6} / \gradientpmrg{17.4} & \gradientpmg{40.2} / \gradientpmg{41.6} & \gradientcg{34.1} / \gradientcg{34.8} & \gradientcmrg{7.9} / \gradientcmrg{8.3} & \gradientcmg{21.0} / \gradientcmg{21.6} & \gradientdg{28.3} / \gradientdg{31.9} & \gradientdmrg{6.5}\text{ } / \gradientdmrg{7.4}\text{ } & \gradientdmg{17.4} / \gradientdmg{19.7} \\
    VCTree + GCL\cite{dong2022stacked} & \gradientg{44.8} / \gradientg{46.6} & \gradientpmrg{35.4} / \gradientpmrg{36.7} & \gradientpmg{40.1} / \gradientpmg{41.7} & \gradientcg{23.7} / \gradientcg{24.5} & \gradientcmrg{17.3} / \gradientcmrg{18.0} & \gradientcmg{20.5} / \gradientcmg{21.3} & \gradientdg{17.6} / \gradientdg{20.7} & \gradientdmrg{15.6} / \gradientdmrg{17.8} & \gradientdmg{16.6} / \gradientdmg{19.3} \\
    VCTree + \textbf{MEET} (ours) & \gradientg{57.3} / \gradientg{63.7} & \gradientpmrg{26.1} / \gradientpmrg{31.3} & \gradientpmg{41.7} / \gradientpmg{47.5} & \gradientcg{28.3}/\gradientcg{31.5} & \gradientcmrg{12.3}/\gradientcmrg{14.0} & \gradientcmg{20.3} / \gradientcmg{22.8} & \gradientdg{25.1}/\gradientdg{28.7} & \gradientdmrg{7.6}\text{ }/\gradientdmrg{10.1} & \gradientdmg{16.4} / \gradientdmg{19.4} \\
    \midrule
    
    SHA\cite{dong2022stacked} & \gradientg{63.3} / \gradientg{65.2} & \gradientpmrg{19.5} / \gradientpmrg{21.1} & \gradientpmg{41.4} / \gradientpmg{43.2} & \gradientcg{32.7} / \gradientcg{33.6} & \gradientcmrg{8.5} / \gradientcmrg{9.0} & \gradientcmg{20.6} / \gradientcmg{21.3} & \gradientdg{25.5} / \gradientdg{29.1} & \gradientdmrg{6.6}\text{ } / \gradientdmrg{7.8}\text{ } &  \gradientdmg{16.1} / \gradientdmg{18.5}\\
    SHA + GCL\cite{dong2022stacked} & \gradientg{42.7} / \gradientg{44.5} & \gradientpmrg{41.0} / \gradientpmrg{42.7} & \gradientpmg{41.9} / \gradientpmg{43.6} & \gradientcg{21.4} / \gradientcg{22.2} & \gradientcmrg{20.6} / \gradientcmrg{21.3} & \gradientcmg{18.1} / \gradientcmg{21.9} & \gradientdg{14.8} / \gradientdg{17.9} & \gradientdmrg{17.8} / \gradientdmrg{20.1} & \gradientdmg{16.3} / \gradientdmg{19.0}\\
    
    SHA + \textbf{MEET} (ours) & \gradientg{69.7} / \gradientg{74.4} & \gradientpmrg{34.2} / \gradientpmrg{42.3} & \gradientpmg{52.0} / \gradientpmg{58.4} & \gradientcg{31.1} / \gradientcg{33.7} & \gradientcmrg{12.9} / \gradientcmrg{15.6} & \gradientcmg{22.0} / \gradientcmg{24.7} & \gradientdg{25.3} / \gradientdg{28.9} & \gradientdmrg{7.2}\text{ } / \gradientdmrg{10.1} & \gradientdmg{16.3} / \gradientdmg{19.5}\\
    
    \midrule
    \textbf{VETO} (ours) & \gradientg{64.5} / \gradientg{66.0} & \gradientpmrg{21.2} / \gradientpmrg{22.1} & \gradientpmg{42.9} / \gradientpmg{44.0} & \gradientcg{30.4} / \gradientcg{31.5} & \gradientcmrg{8.6} / \gradientcmrg{9.1} & \gradientcmg{19.5} / \gradientcmg{20.3} & \gradientdg{26.1} / \gradientdg{29.0} & \gradientdmrg{7.0}\text{ } / \gradientdmrg{8.1}\text{ } & \gradientdmg{16.6} / \gradientdmg{18.6} \\
    
    \textbf{VETO + MEET} (ours) & \gradientg{73.9} / \gradientg{78.3} & \gradientpmrg{43.3} / \gradientpmrg{50.5} & \gradientpmg{58.6} / \gradientpmg{64.4} & \gradientcg{34.6} / \gradientcg{37.2} & \gradientcmrg{19.7} / \gradientcmrg{22.5} & \gradientcmg{27.2} / \gradientcmg{29.9}  & \gradientdg{26.7} / \gradientdg{31.0} & \gradientdmrg{12.1} / \gradientdmrg{16.0} & \gradientdmg{19.4} / \gradientdmg{23.5} \\
    
    \bottomrule
    
\end{tabular}
    \end{adjustbox}
    \label{tab:gqa}
    \vspace{-0.5em}
\end{table*}

\subsection{Experimental setup}
\textbf{Dataset.} We evaluate our approach on two common SGG datasets: Visual Genome (VG) \cite{krishna2017visual} and GQA \cite{hudson2019gqa}. For VG we adopt the popular VG150 split, which consists of 150 object classes and 50 predicate classes in line with previous SGG work \cite{xu2017scene, zellers2018neural, chen2019knowledge,yu2020cogtree,lin2020gps, tang2019learning,tang2020unbiased, li2021bipartite, suhail2021energy}. For GQA we adopt the GQA200 split used by Dong \etal~\cite{dong2022stacked}. For both datasets, depth maps are generated using the monocular depth estimator of Yin \etal \cite{yin2021learning}.

\textbf{Evaluation protocol.} We evaluate our model on the most common SGG tasks \cite{xu2017scene, zellers2018neural}: \emph{(1)} Predicate Classification (PredCls) predicts the relationships for all object pairs by using both the ground-truth bounding boxes and classes; \emph{(2)} Scene Graph Classification (SGCls) predicts both the object classes and their pairwise relationships by using ground-truth bounding boxes; \emph{(3)} Scene Graph Detection (SGDet) detects, classifies, and predicts the pairwise relationships for all the objects in an image.

\textbf{Evaluation metrics.} Following previous work \cite{dong2022stacked, li2021bipartite}, we use Recall@k (R@k) and mean Recall@k (mR@k) as our evaluation metrics. We also report the Average of recall and mean recall (A@k) to show the combined performance improvement of R@k and mR@k. The A@k metric is relevant because previous models with improved mR@k have lower R@k and vice-versa (\cf Fig.~\ref{fig.bias}).

\textbf{Implementation details.} We implement VETO and MEET in PyTorch on Nvidia A100 GPUs. Following prior work \cite{tang2020unbiased, li2019visualbert}, we adopt a ResNeXt-101-FPN \cite{xie2017aggregated} backbone and a Faster R-CNN \cite{ren2015faster} object detector. The parameters of backbone and detector are kept frozen. 
VETO contains 6 relation encoder layers with 6 attention heads for each MSA~\cite{dosovitskiy2020image} component, uses embeddings of size 576, a patch size of 2, and a pooled entity resolution of 8 for entity patch generation. For VETO + Rwt, we use the importance weighting of 
\cite{zareian2020bridging}.
We train our model using the Adam optimizer \cite{kingma2014adam}, batch size $12$, and an initial learning rate of $\num{1.2e-3}$. We apply a linear learning rate warmup over the first $3$K iterations and train for $125$K iterations using a learning rate decay with maximum decay step $3$ and patience $2$.

\subsection{Experimental results}
Using this protocol we are now able to address {\bf Qs. 1--5}.

\textbf{(Q1) Comparison with state of the art.} As shown in Tabs.~\ref{tab:vg} and \ref{tab:gqa}, our VETO + MEET model fulfills the fundamental requirement of unbiased SGG, \ie it improves on both R@k and mR@k metrics, yielding state-of-the-art-performance in terms of the balanced A@k metric across \emph{all} tasks for \emph{both} datasets (except for SGDet in VG where we are comparable). The heat-map pattern reveals that previous models with high mR@k, \eg, SHA + GCL \cite{dong2022stacked}, gain performance improvements on the under-represented predicates while losing significantly on the more frequent ones as revealed by the lower R@k. Our VETO model with a simple re-weighting technique (VETO + Rwt, Tab.~\ref{tab:vg}) already outperforms leading baselines without notably reducing the R@k metric. This is exemplified by the A@k metric and the heat-map hue having less within-row variance than the baselines. 
 In  addition,  our final model VETO + MEET outperforms the previously best Motifs + IETrans + Rwt~\cite{zhang2022fine} by a remarkable 47\% and 48\% relative improvement on A@100 for PredCls for VG and GQA, respectively. To the best of our knowledge, our VETO + MEET model is the first to attain state-of-the-art results on \emph{both} R@k and mR@k metrics for PredCls. It also yields state-of-the-art results on mR@100 for SGCls. 

\textbf{(Q2) MEET with other SGG approaches.}  Among the models trained with MEET in Tabs.~\ref{tab:vg} and~\ref{tab:gqa}, Motifs + MEET and SHA + MEET show notable improvements on the A@k metric for the PredCls and SGCls tasks. However, there is a significant performance gap in comparison to VETO + MEET on all the metrics and tasks. This shows the unbiasing capabilities of MEET as well as the significance of VETO in reducing information loss by using local-level information, resulting in improved SGG performance.

\begin{figure}[t]
   \centering
   \includegraphics[width=0.9\linewidth]{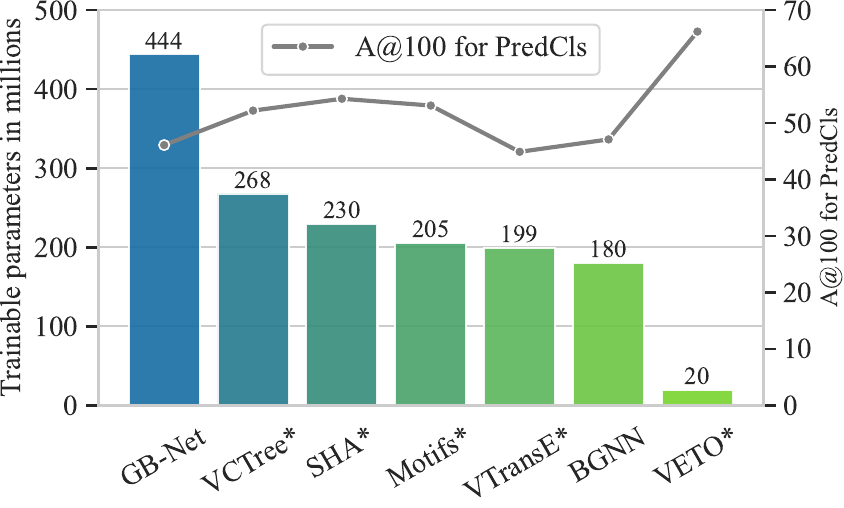}
   \vspace{-2mm}
   \caption{\textbf{No.~of trainable parameters} (Mio.) \textbf{\vs performance} (A@100) of leading SGG models (* denotes debiasing with MEET).
   VETO outperforms prior work and is $\sim 10\times$ leaner.}
\label{fig:param}
\vspace{-0.2cm}
\end{figure}

\textbf{(Q3) Light-weight VETO}. The comparison of SGG models in terms of the number of trainable parameters in Fig.~\ref{fig:param} shows how our local-level entity projections significantly reduce the number of parameters compared to global entity-level projections (\cf Fig.~\ref{fig:intro}). VETO with 20 million parameters is 20 times lighter than GB-Net~\cite{zareian2020bridging}, which uses knowledge graphs as additional modality, and $\sim$10 times lighter than other leading SGG models. Despite this, VETO clearly outperforms previous models in A@k.

\begin{table}
    \centering
    
    \caption{\textbf{Ablation study of VETO on VG}. \textbf{L}: Local-level Entity Patch Generation; \textbf{CR}: Cross-Relation Patch Generation; \textbf{CM}: Cross-Modality Patch Fusion; \textbf{D}: Depth.}
    \vspace{-1mm}
    \begin{adjustbox}{max width=\linewidth, center}
    \begin{tabular}{cccc|ccc}
    
    \toprule
    \multicolumn{4}{c|}{\multirow{1}{*}{\textbf{VETO Components}}} & \multicolumn{3}{c}{\multirow{1}{*}{\textbf{SGCls}}}\\ 
    
    \midrule
    
    \multicolumn{1}{c}{\textbf{L}} & \multicolumn{1}{c}{ \textbf{CR} } & \multicolumn{1}{c}{ \textbf{CM}} & \multicolumn{1}{c|}{ \textbf{D} } & \multicolumn{1}{c}{ \textbf{R@k:}\text{ } \textbf{50 / 100} } & \multicolumn{1}{c}{ \textbf{mR@k:}\text{ } \textbf{50 / 100}} & \multicolumn{1}{c}{ \textbf{A@k:}\text{ } \textbf{50 / 100} } \\ 
    
    \midrule
    \xmark& \xmark & \xmark & \xmark&  \text{ }\gradienta{32.0} \text{ } /  \gradienta{33.5} & \gradientamr{7.6} \text{ } / \gradientamr{8.3} \text{ } & \gradientam{19.8} / \gradientam{20.9} \\
    \xmark& \xmark & \xmark & \cmark &  \text{ }\gradienta{33.2}\text{ }  / \gradienta{34.5} & \gradientamr{7.0} \text{ } / \gradientamr{7.6} \text{ } & \gradientam{20.1}  / \gradientam{21.1} \\
    \cmark & \xmark  & \xmark & \xmark & \gradienta{34.8} / \gradienta{36.2} & \gradientamr{14.1} / \gradientamr{15.1} & \gradientam{24.5} / \gradientam{25.7} \\
    \cmark & \xmark & \xmark & \cmark & \gradienta{35.1} / \gradienta{36.4} & \gradientamr{13.0} / \gradientamr{14.1} & \gradientam{24.1} / \gradientam{25.3} \\
    \cmark & \xmark & \cmark & \cmark & \gradienta{35.1} / \gradienta{36.3} & \gradientamr{14.4} / \gradientamr{15.4} & \gradientam{24.8} / \gradientam{25.9} \\
    \cmark & \cmark & \xmark & \xmark & \gradienta{35.4} / \gradienta{36.6} & \gradientamr{15.2} / \gradientamr{16.1} & \gradientam{25.3} / \gradientam{26.4} \\
    
    \cmark & \cmark & \cmark & \cmark & \gradienta{35.1} / \gradienta{36.3} & \gradientamr{16.1} / \gradientamr{17.1} & \gradientam{25.6} / \gradientam{26.7} \\
    \hline

\end{tabular}
    \end{adjustbox}
    \label{tab:ablation}
    \vspace{-0.5em}
\end{table}

\textbf{(Q4) Benefit of local-level patch generation.} Tab.~\ref{tab:ablation} provides an ablation study of the VETO components. The first two rows denote a transformer-based SGG model without local-level patches. We observe that introducing the local-level patch generation (rows 3 \& 4) notably improves every metric with a $\sim$23\% improvement of A@k, highlighting the significance of local-level patches. We also observe an overall improvement when incrementally adding the VETO components in the subsequent rows. In general, the use of local-level information and the Cross-Relation Patch Generation (2\textsuperscript{nd} to last row) significantly improves performance, with a relative improvement of approximately 28\% in A@k compared to the first ablation that does not use local-level patches. We also observe that adding the depth map components to the final model yields an additional small improvement of the mR@k metric.




\begin{table}[tb]
    \centering
    \caption{\textbf{Impact of depth map quality} (the higher, the better). For fairer comparison, the Depth-VRD model \cite{sharifzadeh2021improving} is reproduced on the ResNeXt-101-FPN backbone. Both models are also debiased using the reweighting strategy of~\cite{zareian2020bridging}. } 
    \vspace{-1mm}
    \begin{adjustbox}{max width=\linewidth, center}
    \begin{tabular}{@{}lccccc@{}}
    
    \toprule  \multicolumn{1}{@{}c}{\textbf{Model}} & \multicolumn{1}{c}{\textbf{mR@k}}  & \multicolumn{1}{c}{\textbf{VG-Depth.v1}} & \multicolumn{1}{@{}c}{\textbf{VG-Depth.v2}} & \multicolumn{1}{c@{}}{\textbf{Improvement}} \\ 
    \midrule
    \multicolumn{1}{@{}c}{\multirow{3}{*}{VETO}} & 20 & 25.3 & 27.5 & \multicolumn{1}{c@{}}{\textbf{9\%}} \\
    \multicolumn{1}{@{}c}{} & 50 & 31.2 & 33.1 & \multicolumn{1}{c@{}}{\textbf{6\%}} \\
    \multicolumn{1}{@{}c}{} & 100 & 33.5 & 35.1 & \multicolumn{1}{c@{}}{\textbf{5\%}} \\
    \midrule
    \multicolumn{1}{@{}c}{\multirow{3}{*}{Depth-VRD \cite{sharifzadeh2021improving}}} & 20 & 17.8 & 18.2 & \multicolumn{1}{c@{}}{2\%} \\
    \multicolumn{1}{@{}c}{} & 50 &  21.7 & 21.9 & \multicolumn{1}{c@{}}{1\%} \\
    \multicolumn{1}{@{}c}{} & 100 & 23.1 & 23.1 & \multicolumn{1}{c@{}}{0\%} \\
    \bottomrule
    \end{tabular}
    \end{adjustbox}
    \label{tab:depth_compare1}
    \vspace{-0.5em}
\end{table}

\begin{table}[tb]
\centering
\caption{\textbf{Improvement of VETO over Depth-VRD for PredCls.} Conventions as described in Tab.~\ref{tab:depth_compare1}.}
\label{tab:depth_compare2}
\small
\vspace{-1mm}
\begin{tabularx}{\linewidth}{@{}Xccc@{}}
\toprule
\textbf{Depth Map} & \textbf{mR@20} & \textbf{mR@50} & \textbf{mR@100} \\
\midrule
VG-Depth.v1 & 42\% & 44\% & 45\% \\
VG-Depth.v2 & 51\% & 51\% & 52\% \\
\bottomrule
\end{tabularx}
\vspace{-0.5em}
\end{table}

\begin{figure*}[h]
    
    \hfill
        \centering
        
        \includegraphics[width=0.12\linewidth]{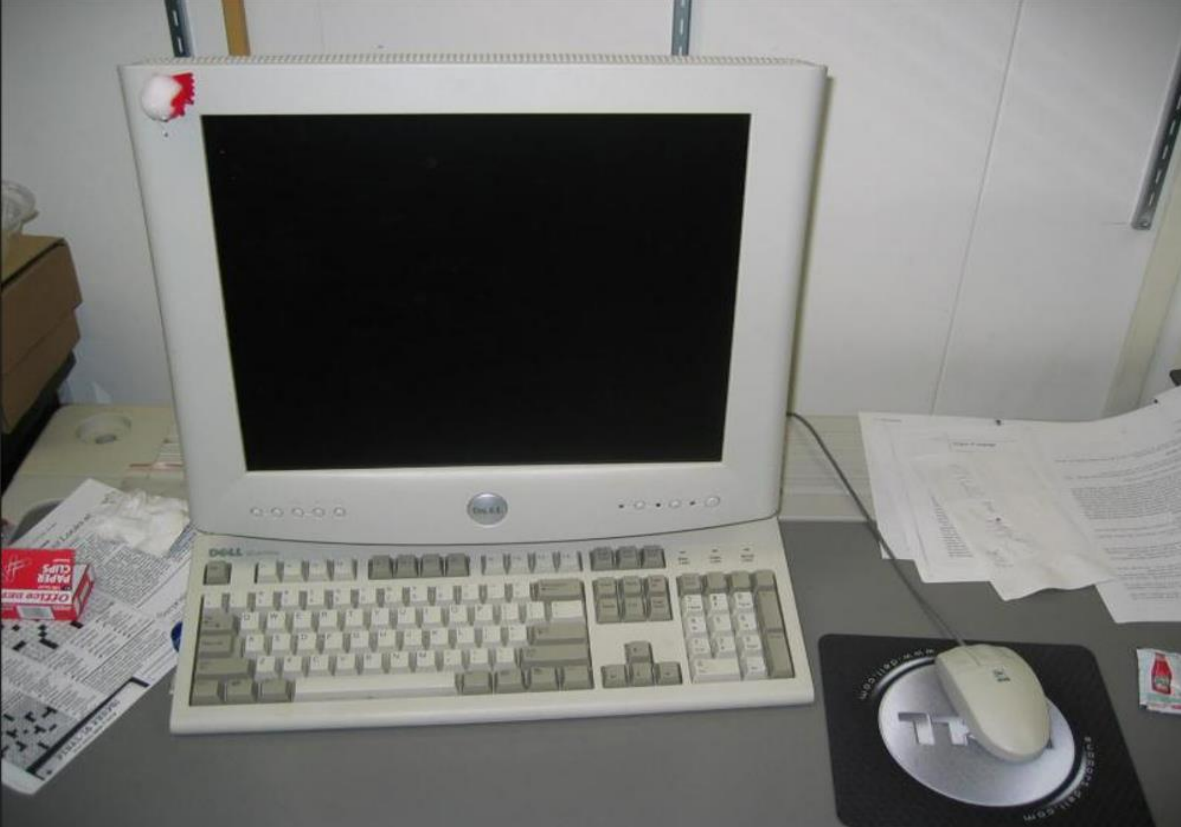}
        \includegraphics[width=0.12\linewidth]{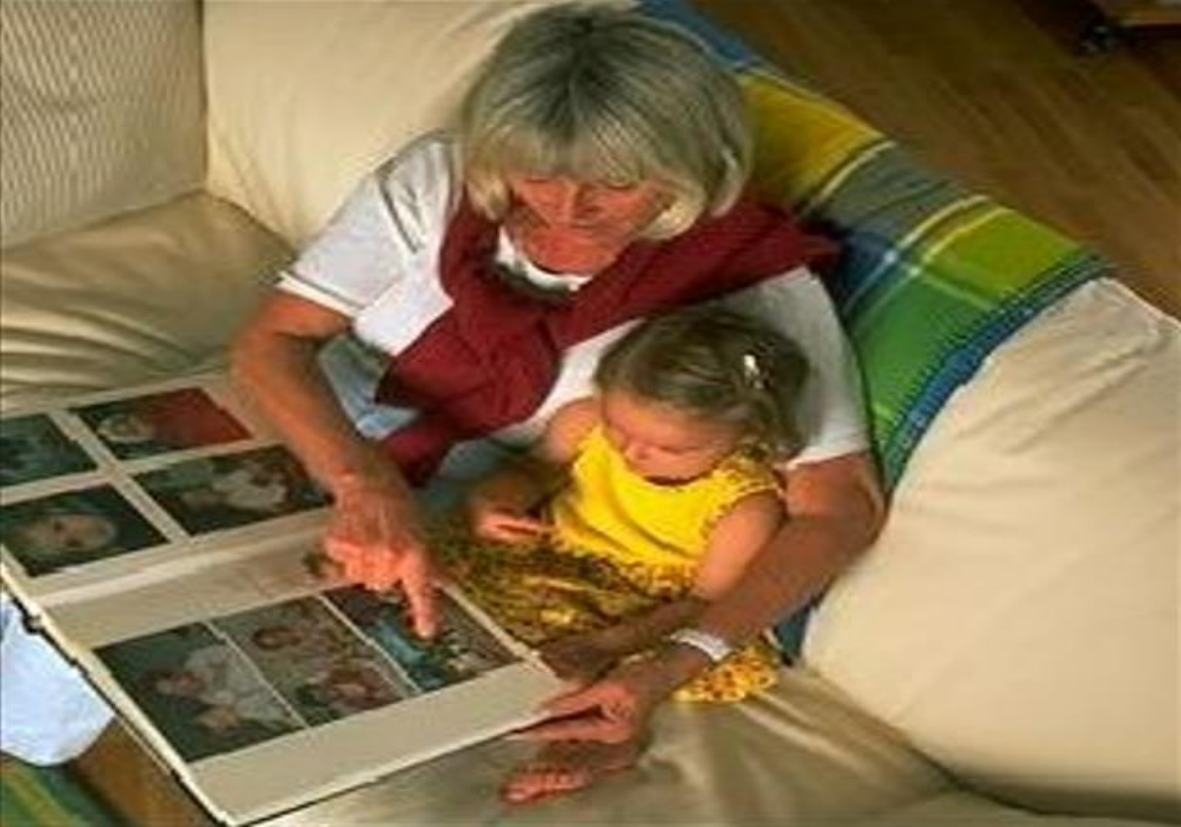}
        \includegraphics[width=0.12\linewidth]{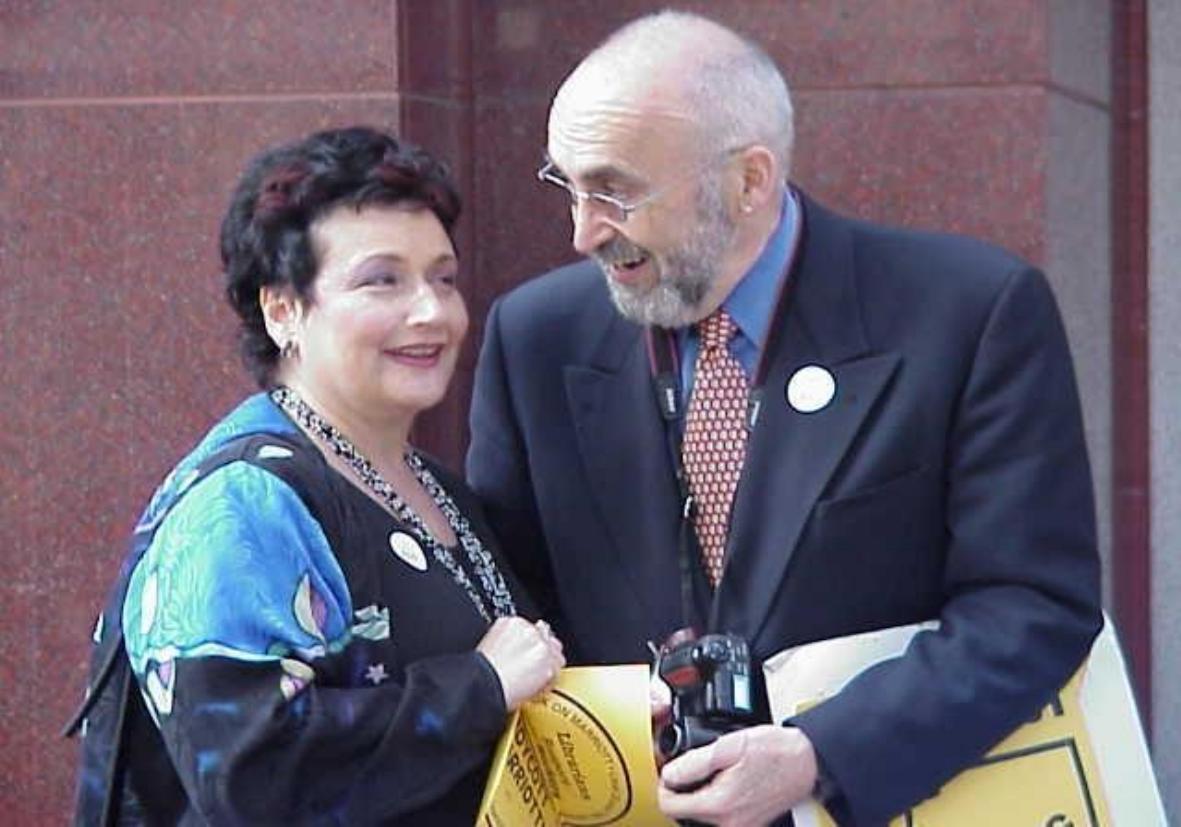}
        \includegraphics[width=0.12\linewidth]{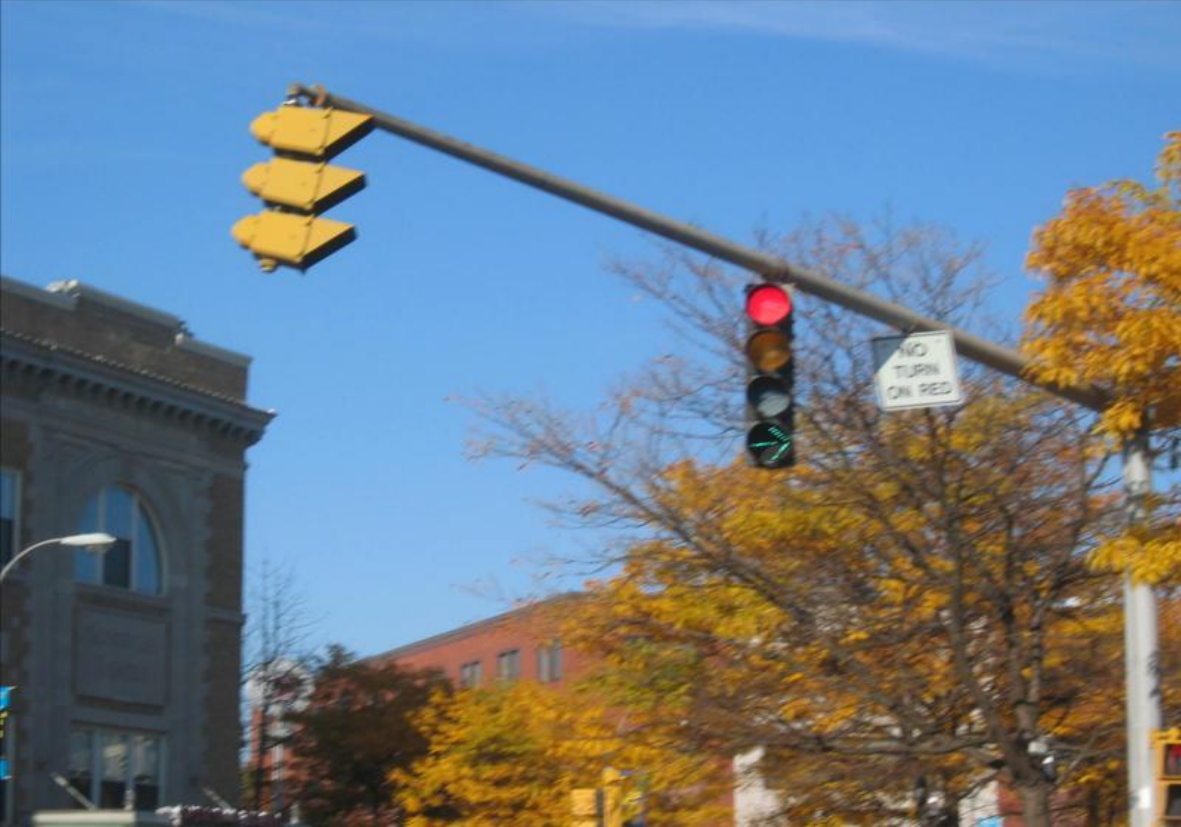}
        \includegraphics[width=0.12\linewidth]{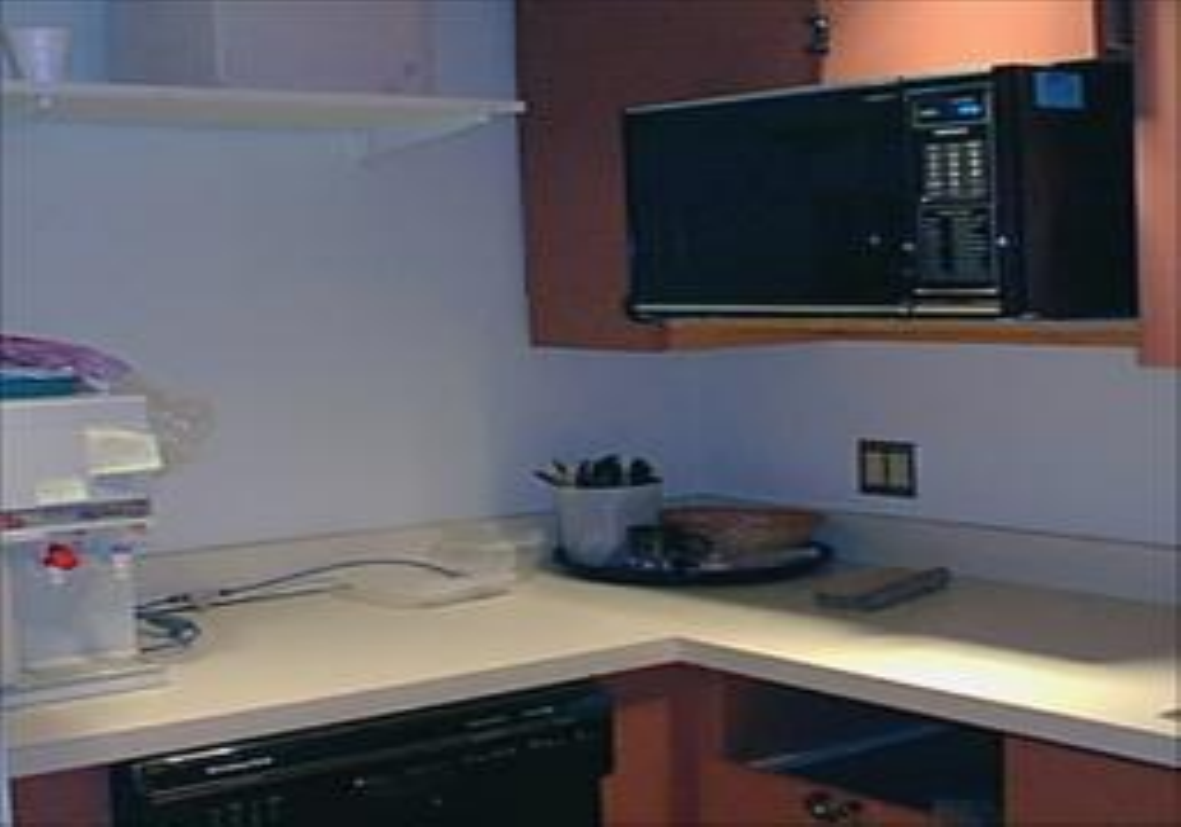}
        \includegraphics[width=0.12\linewidth]{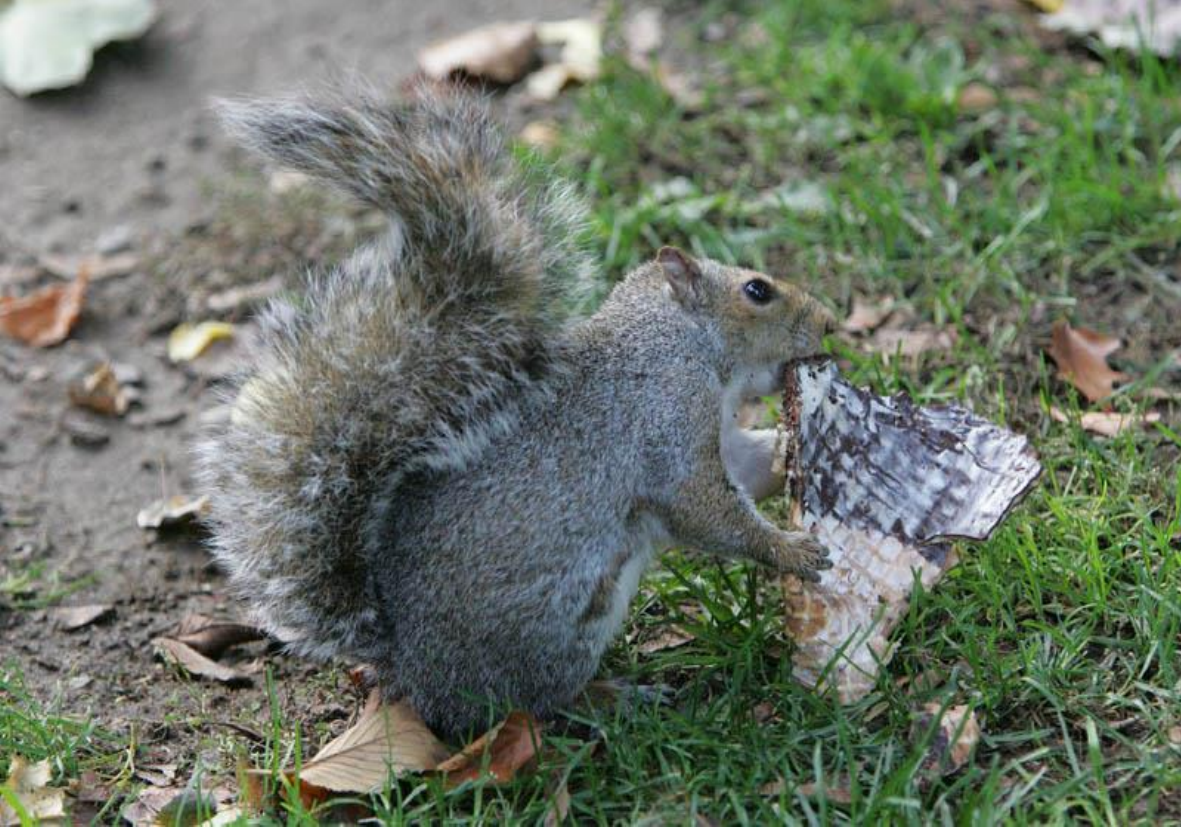}
        \includegraphics[width=0.12\linewidth]{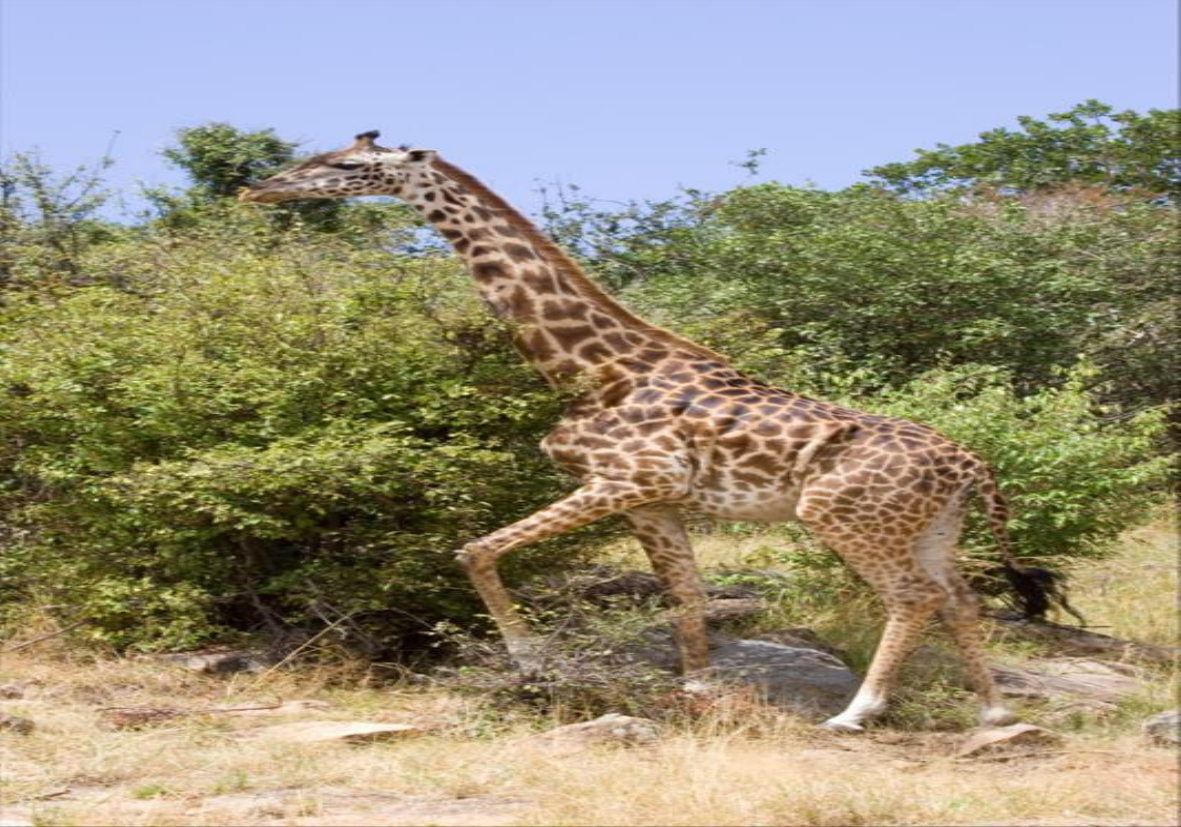}
        \includegraphics[width=0.12\linewidth]{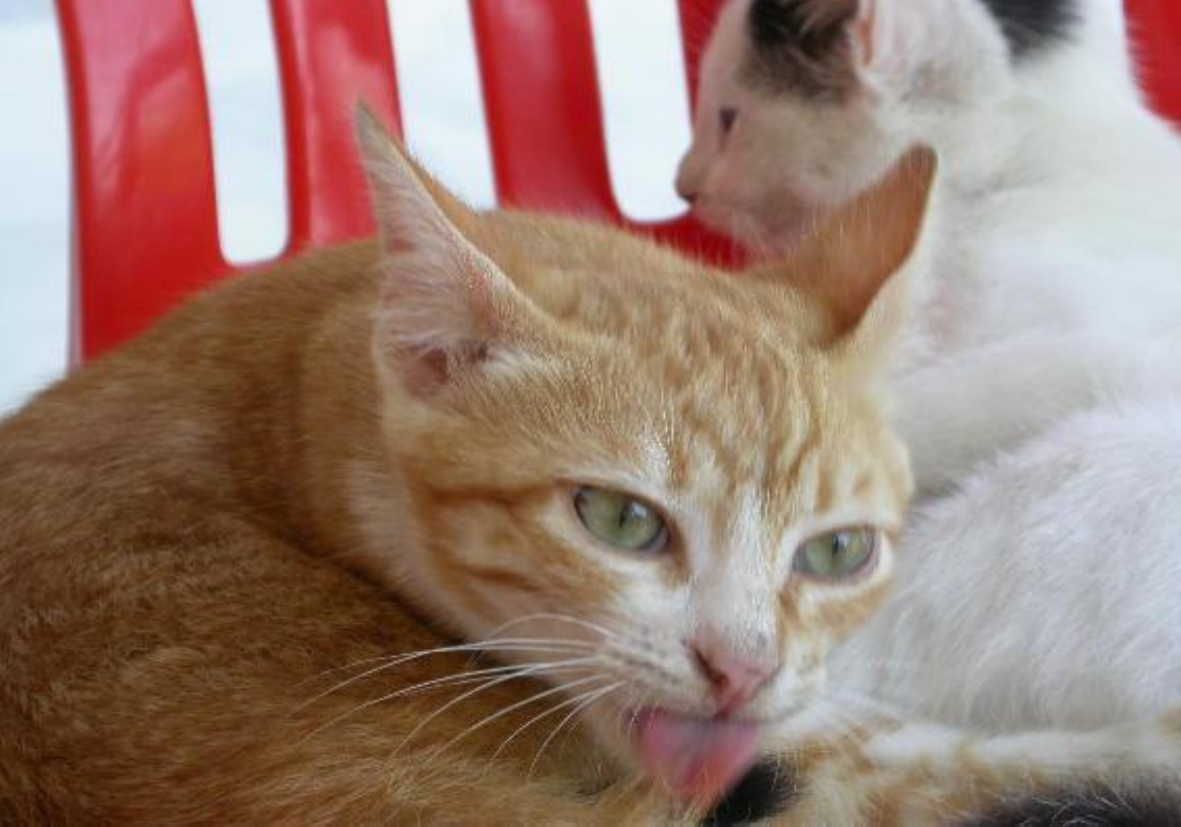}\\
        
        \centering
        \includegraphics[width=0.12\linewidth]{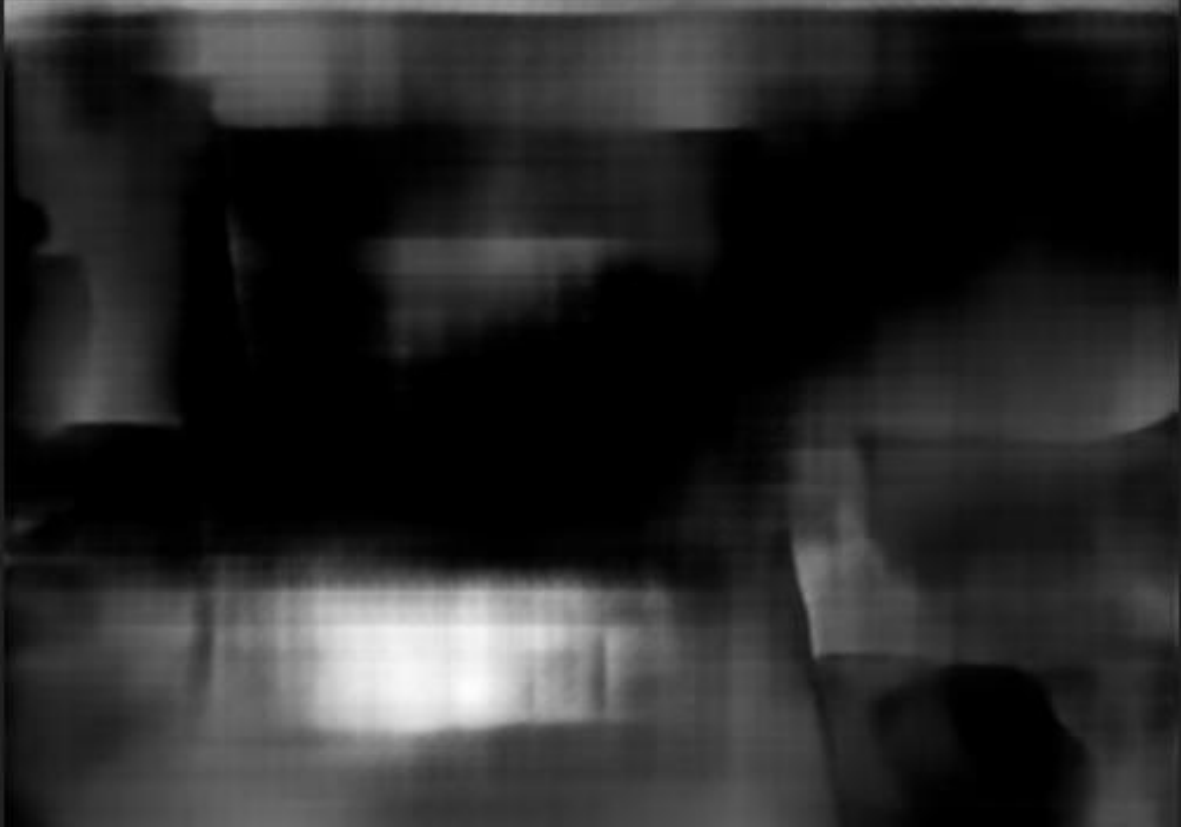}
        \includegraphics[width=0.12\linewidth]{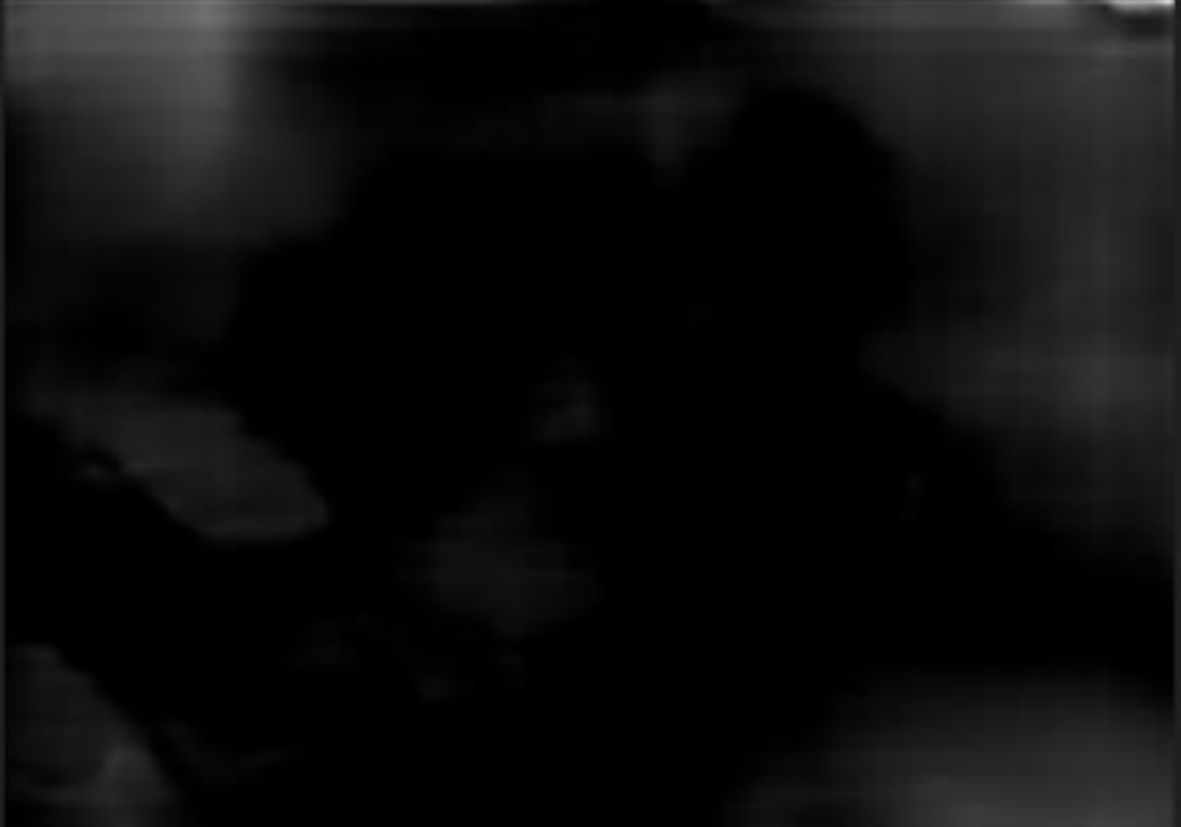}
        \includegraphics[width=0.12\linewidth]{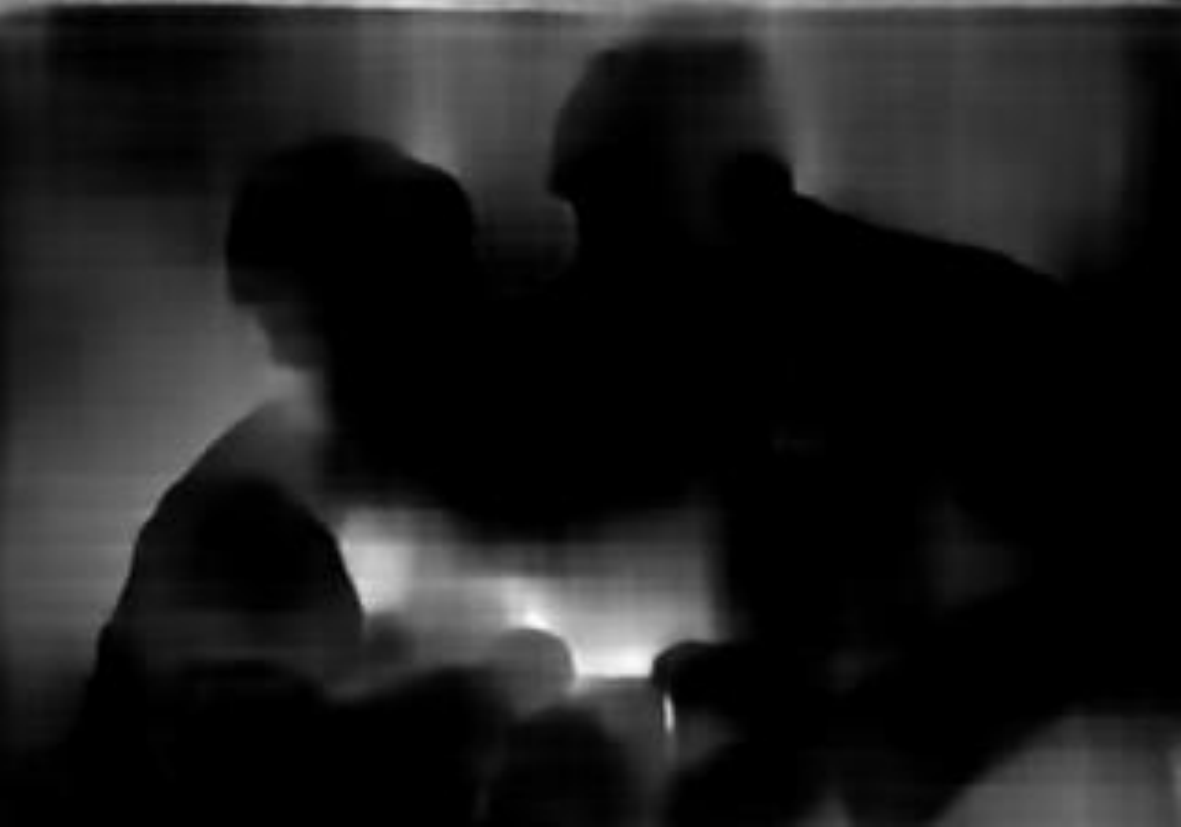}
        \includegraphics[width=0.12\linewidth]{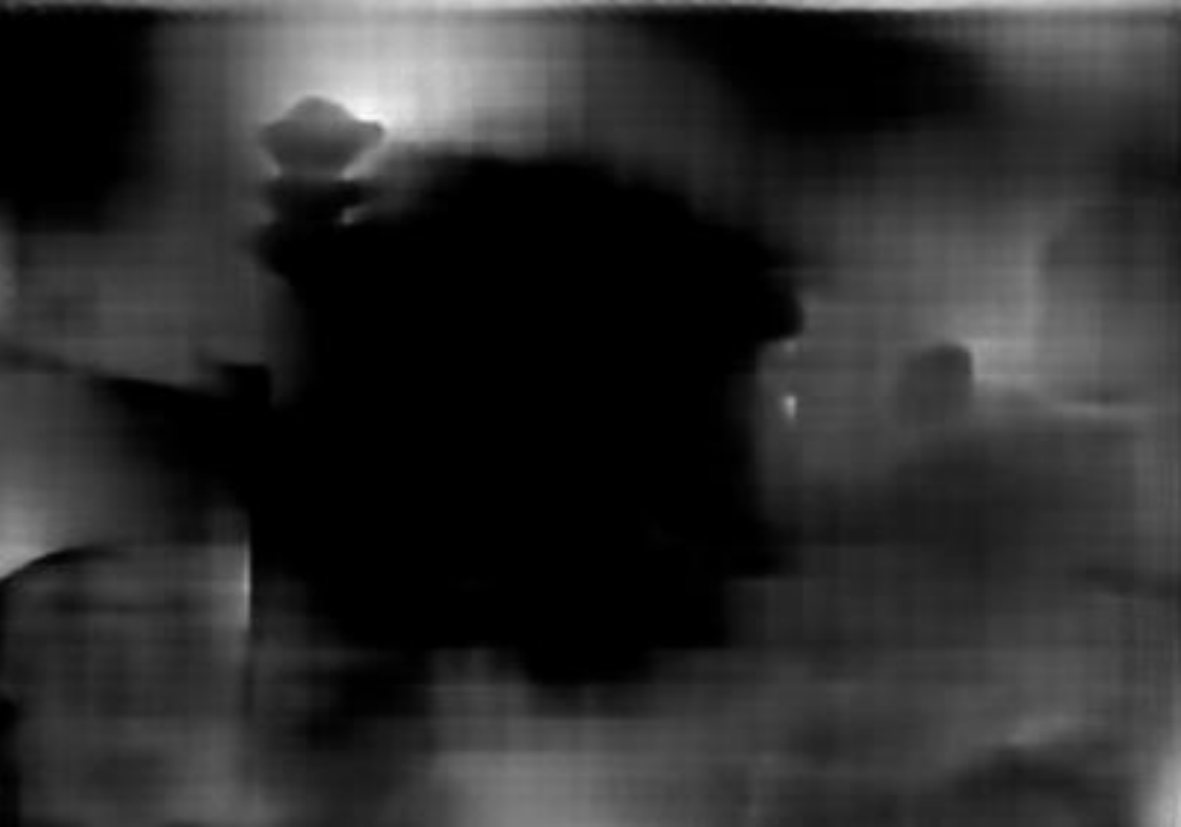}
        \includegraphics[width=0.12\linewidth]{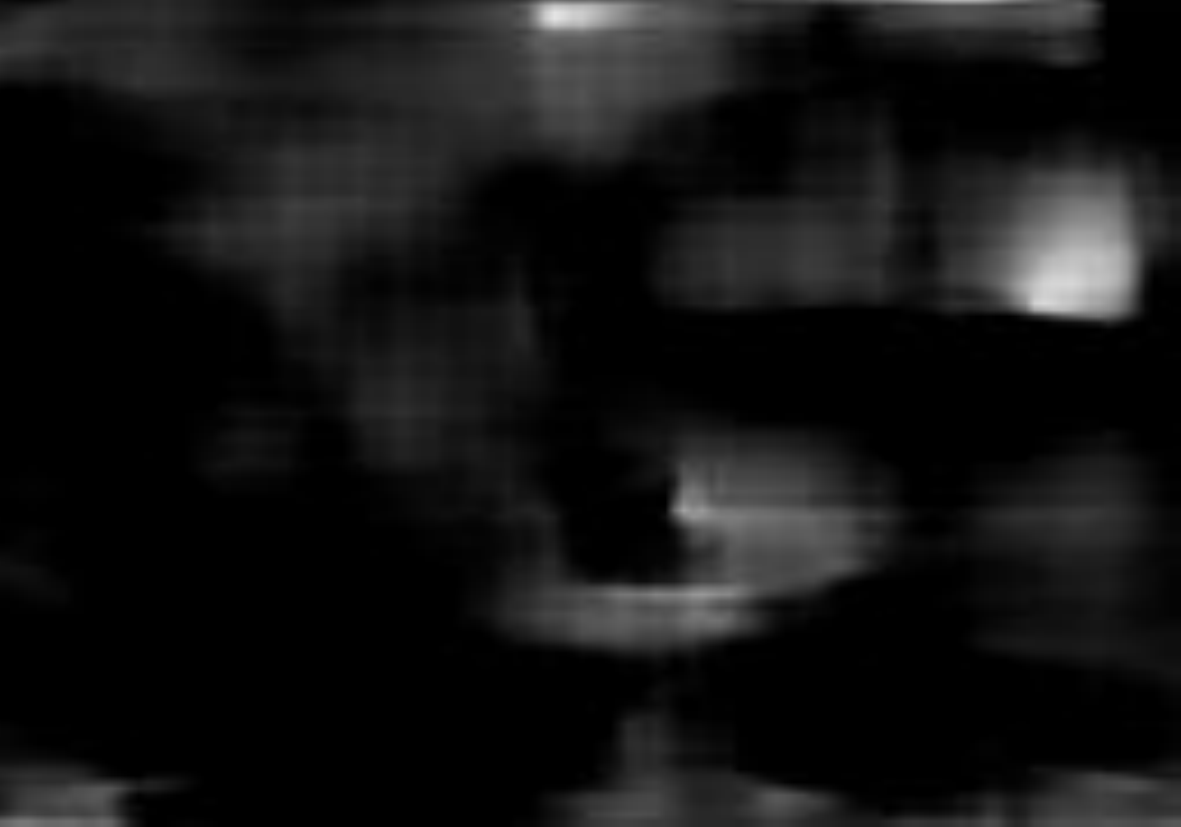}
        \includegraphics[width=0.12\linewidth]{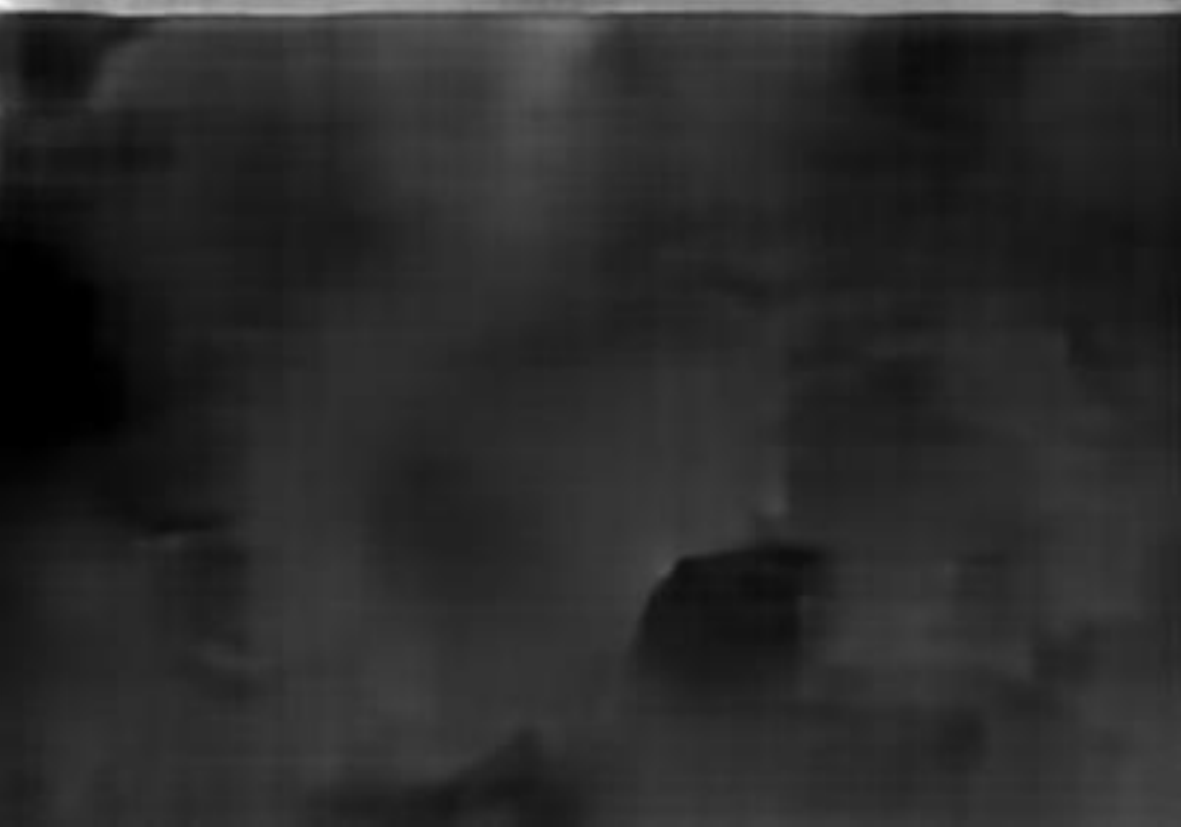}
        \includegraphics[width=0.12\linewidth]{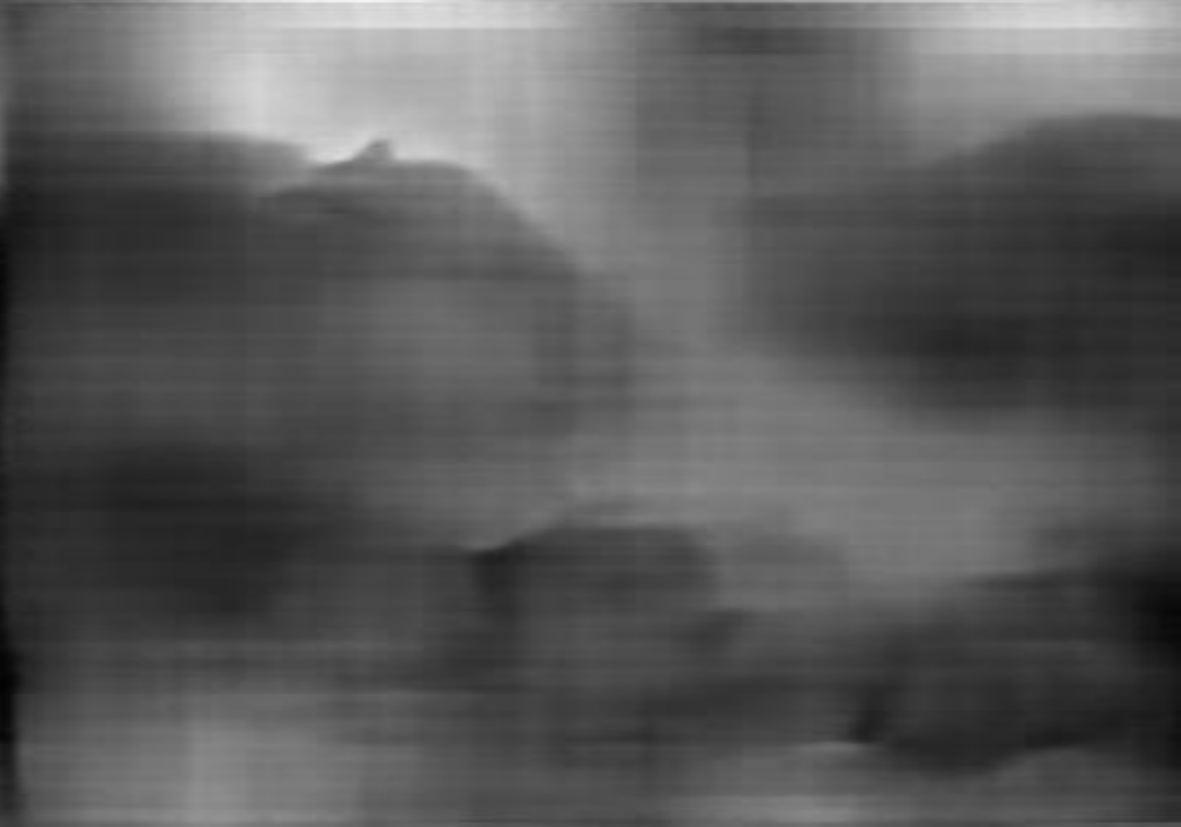}
        \includegraphics[width=0.12\linewidth]{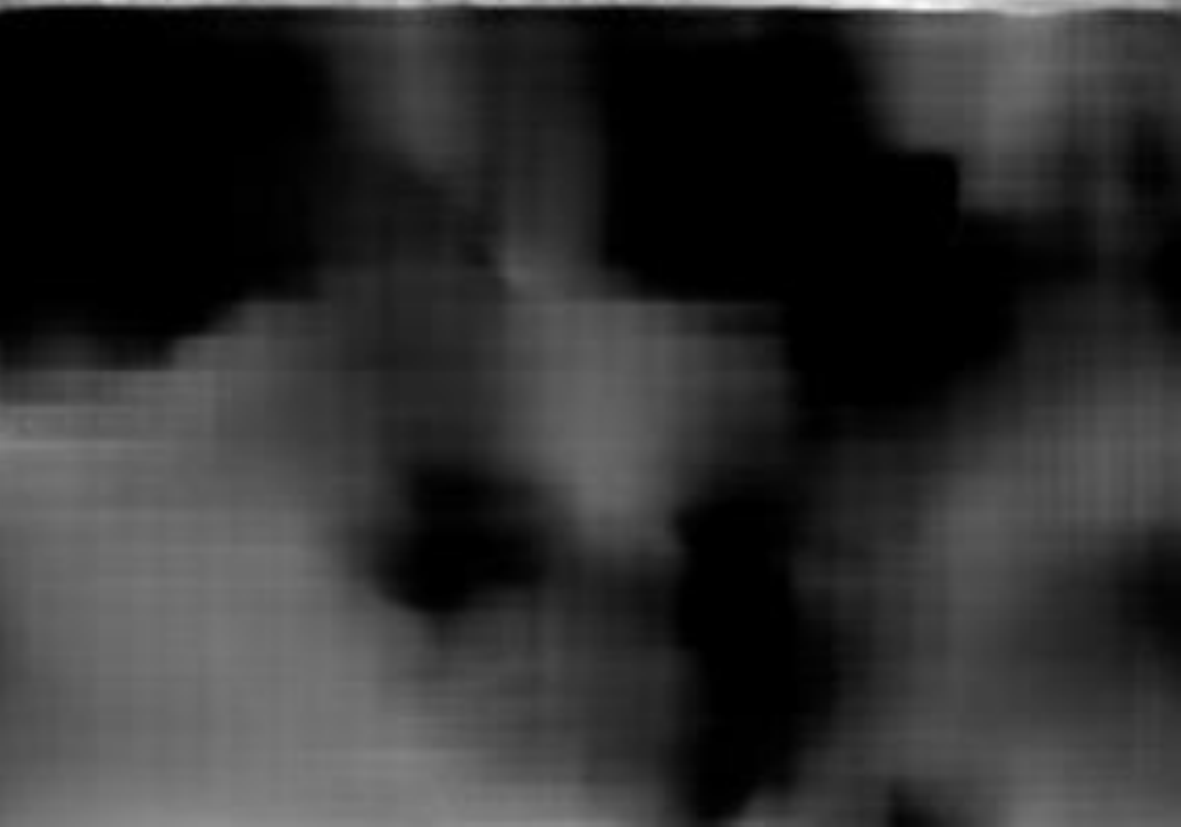}\\

        \centering
        \includegraphics[width=0.12\linewidth]{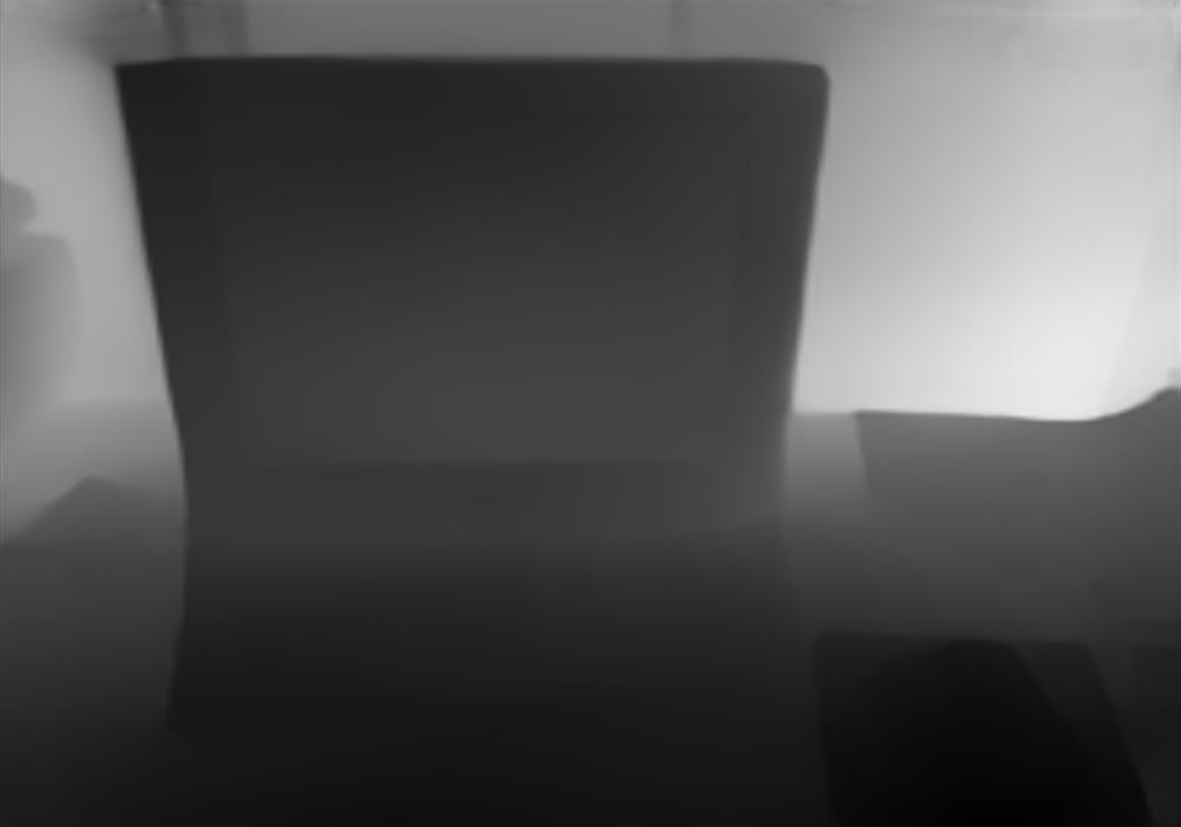}
        \includegraphics[width=0.12\linewidth]{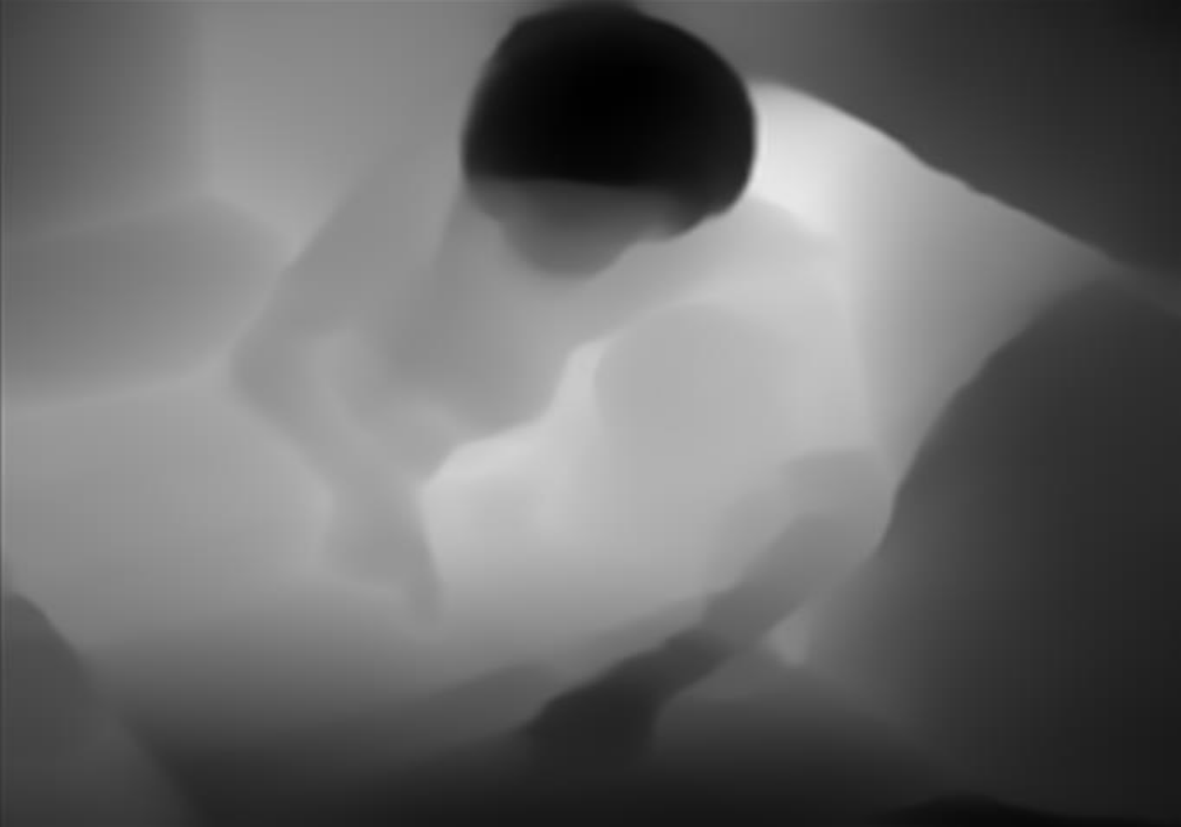}
        \includegraphics[width=0.12\linewidth]{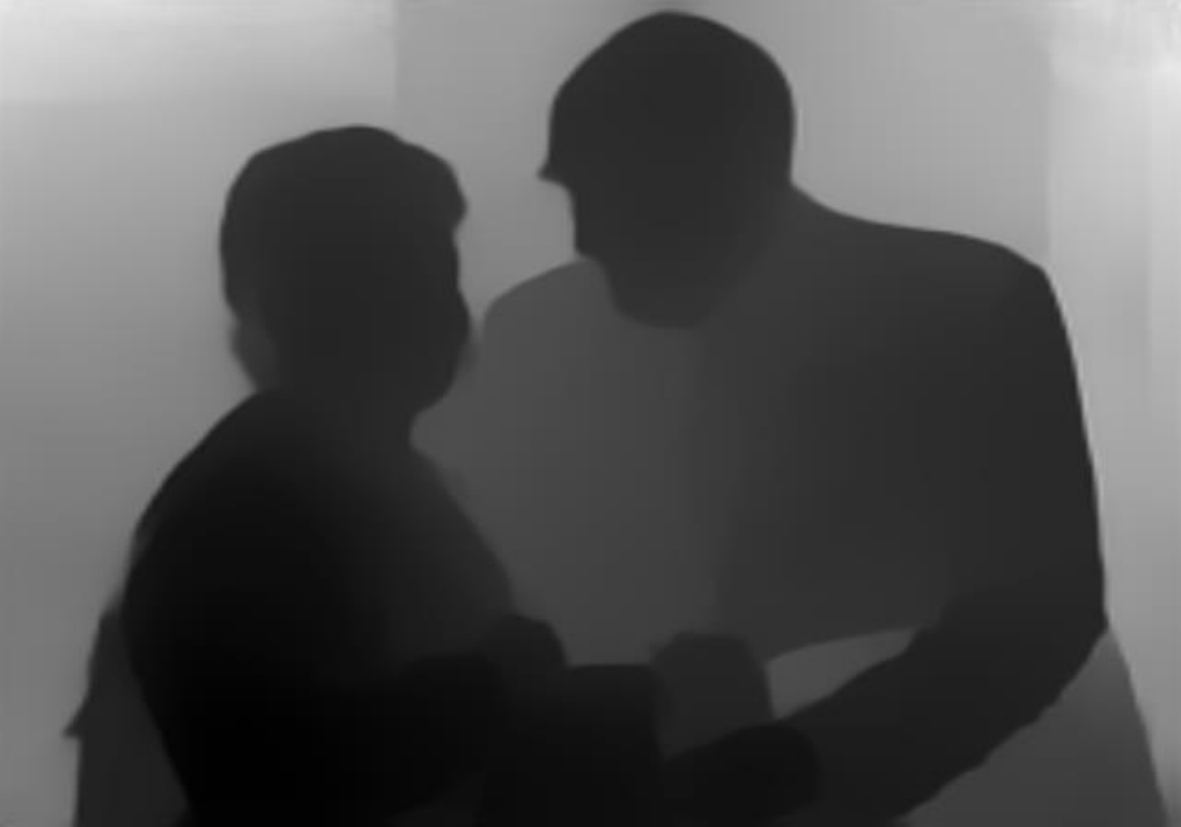}
        \includegraphics[width=0.12\linewidth]{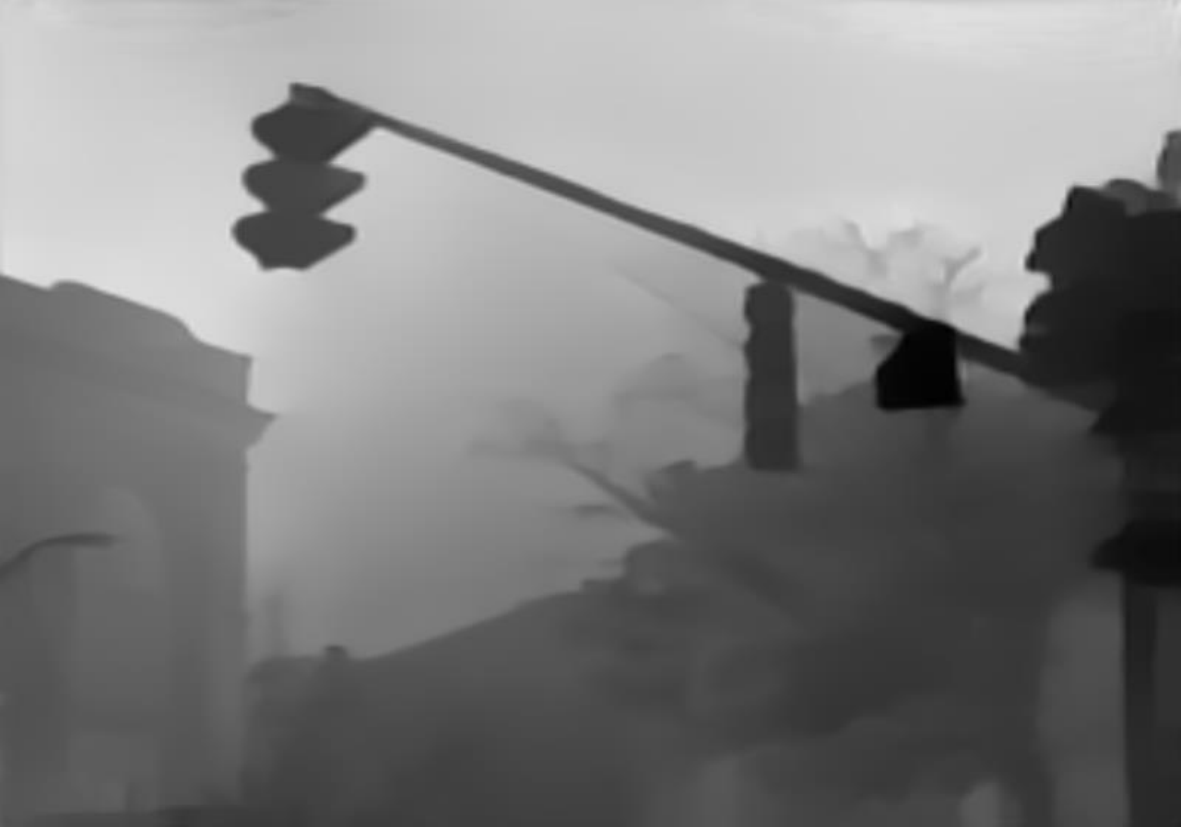}
        \includegraphics[width=0.12\linewidth]{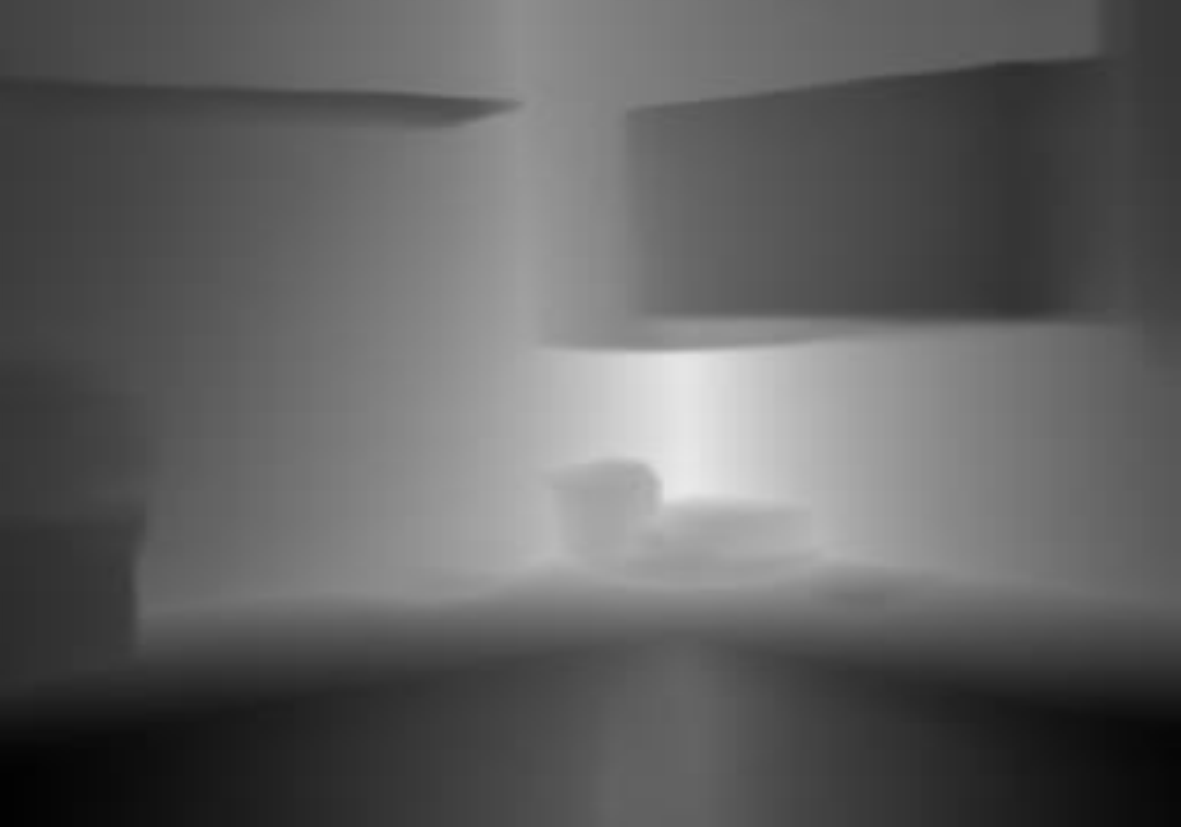}
        \includegraphics[width=0.12\linewidth]{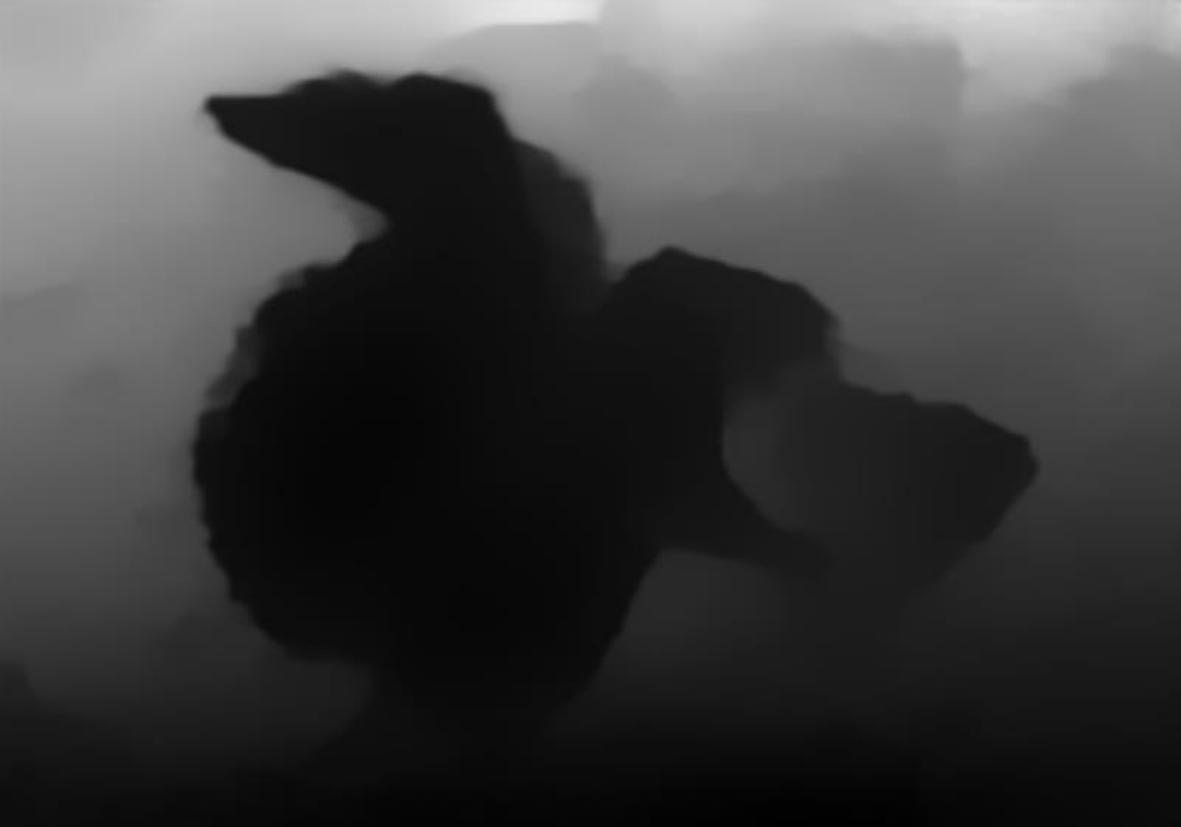}
        \includegraphics[width=0.12\linewidth]{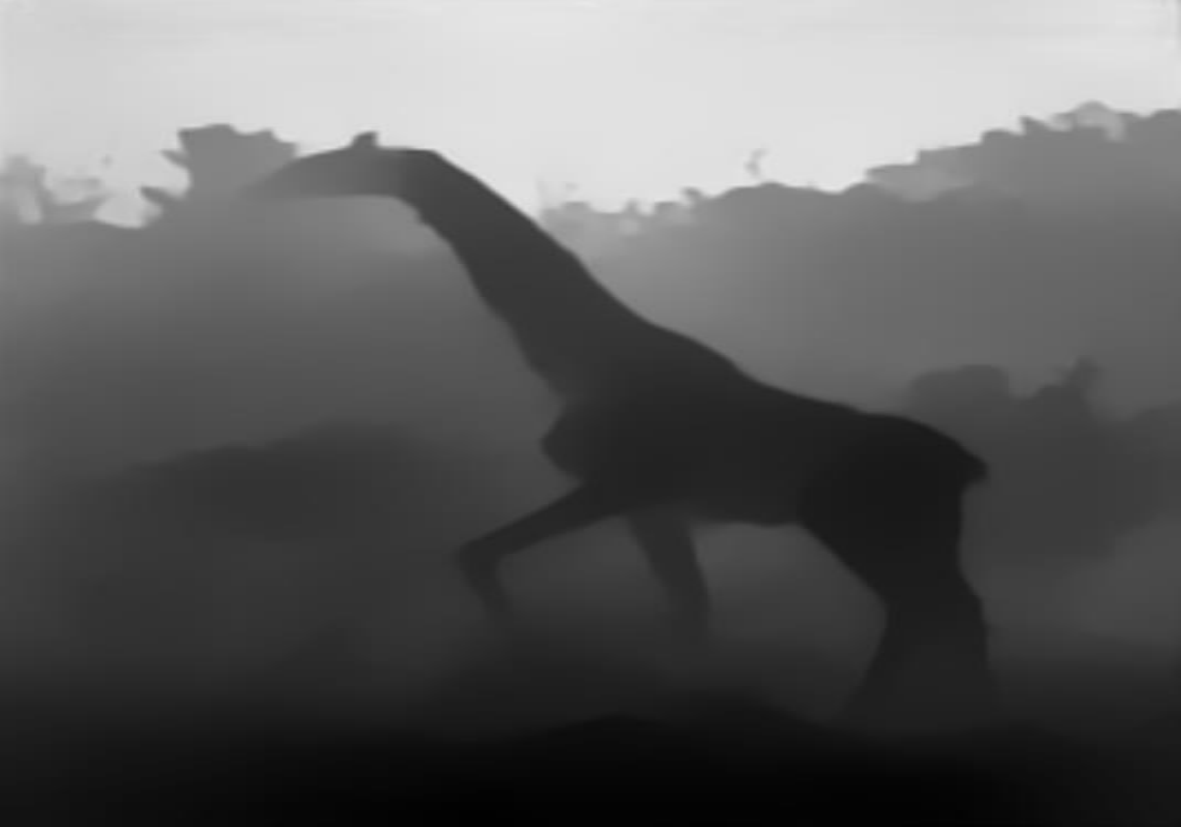}
        \includegraphics[width=0.12\linewidth]{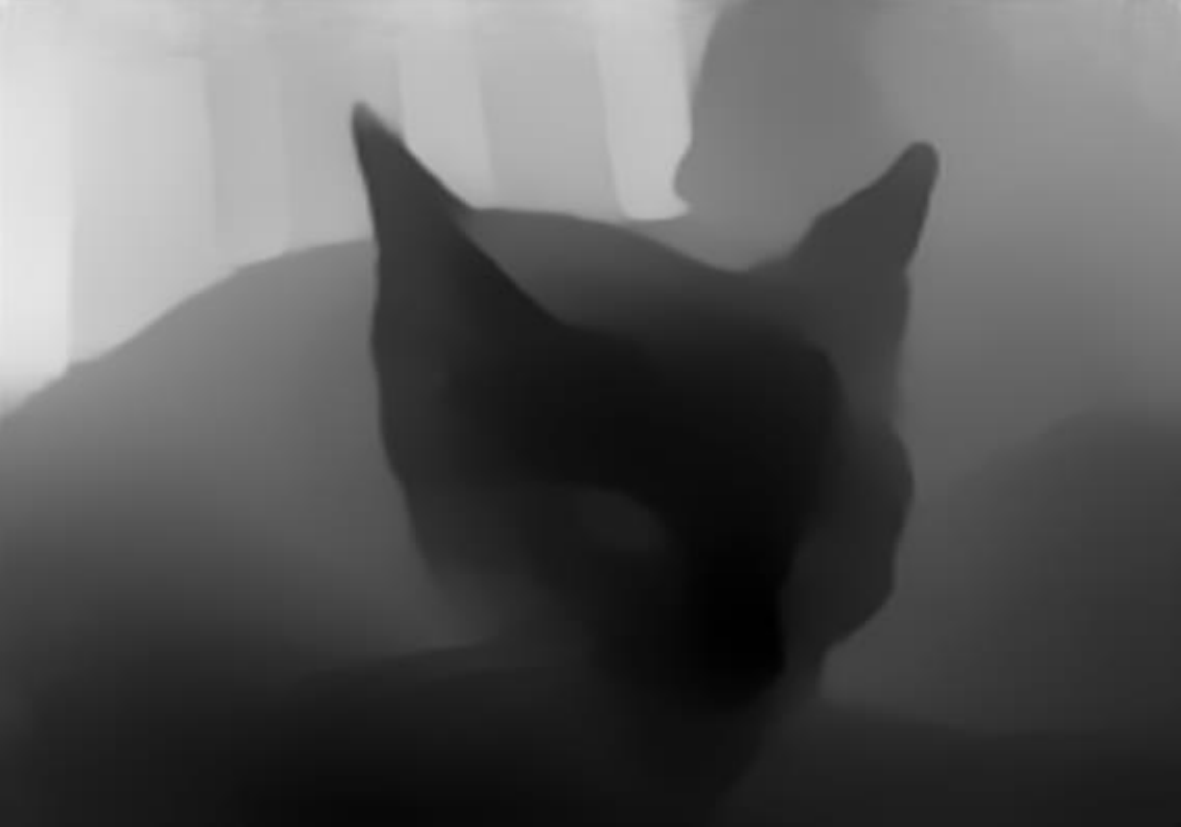}
        
\caption{Failure cases reported by \cite{sharifzadeh2021improving}.
The second row shows ``noisy'' depth maps from \cite{sharifzadeh2021improving} (VG-Depth.v1).
\label{fig:depth_map}
The bottom row represents the improved depth maps used in VETO (VG-Depth.v2), generated using the monocular depth estimator of~\cite{yin2021learning}.}\label{fig:depth}
\vspace{-0.5em}
\end{figure*}

\textbf{(Q5) Benefit of depth map.} Comparing rows 1 and 2 of Tab.~\ref{tab:ablation} shows that introducing the depth modality to the model without local-level patches yields only a slight R@k improvement while mR@k drops. We observe a similar trend when comparing rows 3 and 4. However, after introducing the Cross-Modality Patch Fusion module, mR@k and A@k improve.   We also perform an extensive depth data analysis to investigate the modality fusion potential of VETO.
Fig.~\ref{fig:depth_map} shows “noisy” depth-map samples used by Sharifzadeh \etal\ for Depth-VRD~\cite{sharifzadeh2021improving} (VG-Depth.v1); the bottom row shows the corresponding high-quality depth maps as extracted by the monocular depth estimator of Yin \etal~\cite{yin2021learning} (VG-Depth.v2). 
We use VG-Depth.v1 and VG-Depth.v2 to compare and contrast VETO and Depth-VRD. 
We analyse the significance of the depth-map quality on the SGG performance and the importance of a careful architectural design to make use of the depth map efficiently. As depicted in Tab.~\ref{tab:depth_compare1}, the performance improvement for our model on VG-Depth.v2 generated using
the monocular depth estimator of Yin \etal~\cite{yin2021learning} over VG-Depth.v1~\cite{sharifzadeh2021improving} is around 7\%. 
To the contrary, Depth-VRD shows only a minor improvement of 1\% on the high-quality VG-Depth.v2 dataset. 
Furthermore, the improvement of VETO over Depth-VRD in Tab.~\ref{tab:depth_compare2} shows that, overall, VETO has a significant improvement of more than 40\% respectively 50\% in mR@k over Depth-VRD for the VG-Depth.v1 respectively VG-Depth.v2 depth maps.

\begin{figure}[t]
\centering
   \includegraphics[width=\linewidth]{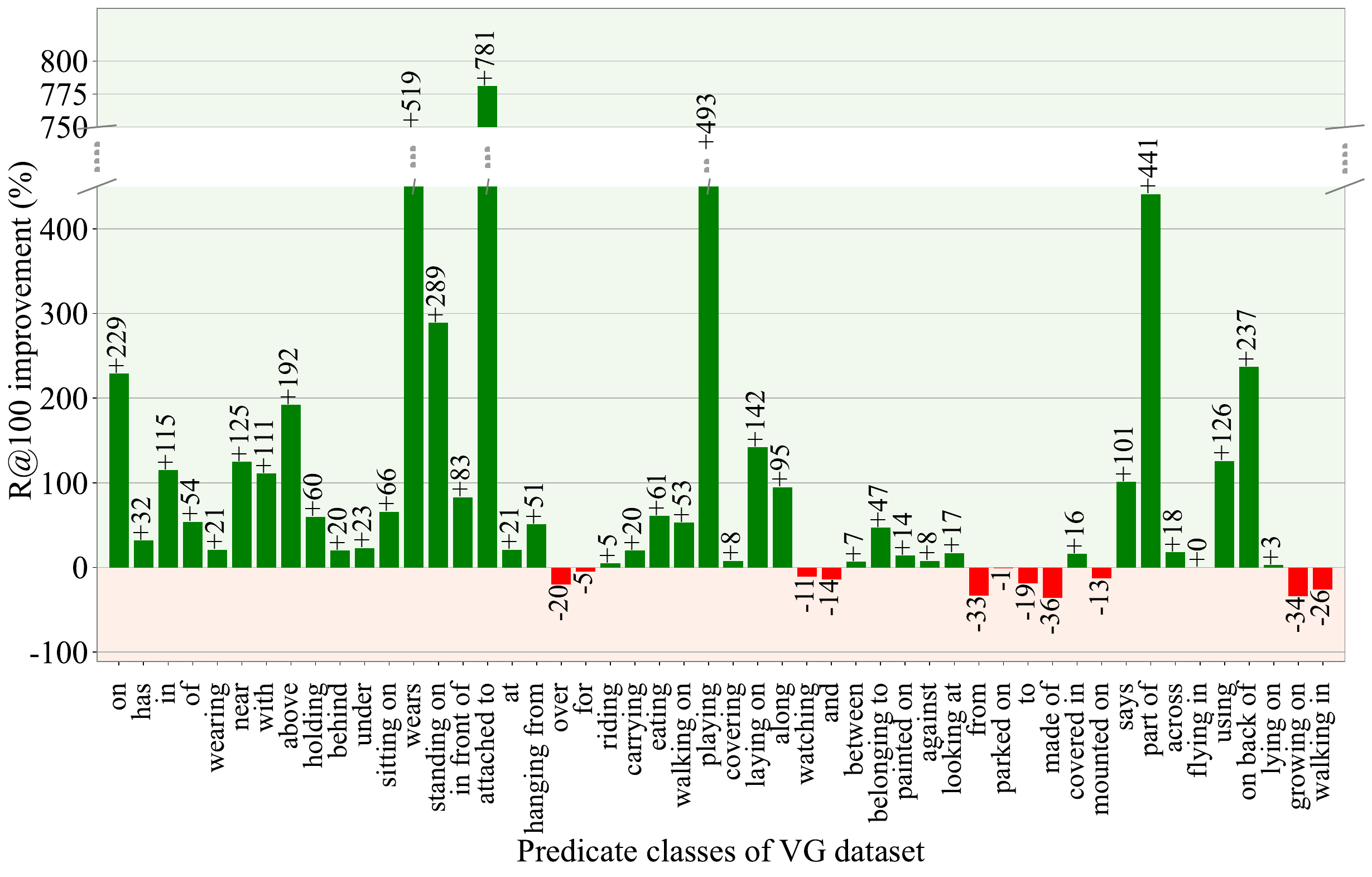}\\
   \caption{\textbf{R@100 improvement  on PredCls for VETO + MEET over SHA + GCL \cite{dong2022stacked}.} 
   The predicates are sorted based on their frequency in descending order.}
\label{fig:r100}
\vspace{-0.5em}
\end{figure}

\begin{figure}
\centering
   \includegraphics[ width=\linewidth]{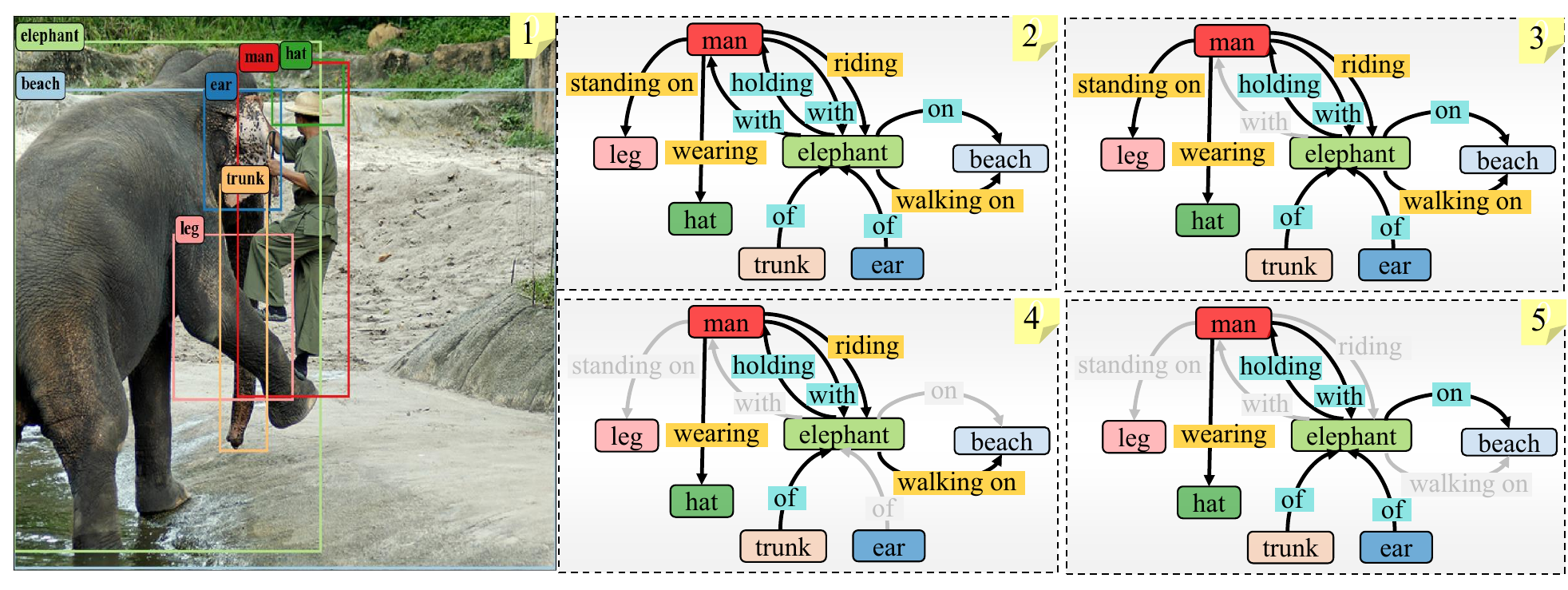}\\
   \caption{\textbf{Qualitative example}. Head and tail relations are highlighted in blue and yellow, respectively. Greyed out relations and arrows denote missed predictions. \emph{(1)} VG sample with ground-truth bounding boxes and labels; \emph{(2)}  SGG ground-truth; \emph{(3)} VETO + MEET predicts both head and tail classes; \emph{(4)} SHA + GCL~\cite{dong2022stacked} misses head classes such as \emph{on}, \emph{of}; \emph{(5)} Motifs~\cite{zellers2018neural} misses many tail classes such as \emph{walking on}, \emph{riding}, or \emph{standing on}. }
\label{fig:sgg}
\vspace{-0.5em}
\end{figure}

\subsubsection{Further results} 
Fig.~\ref{fig:r100} shows the predicate-specific improvement of VETO + MEET over SHA + GCL (sorted from frequent to less frequent).  
Notice that VETO + MEET improves on every part of the distribution (head, body, and tail). As emphasized in Fig.~\ref{fig.part}, we find an enormous performance boost over SHA + GCL~\cite{dong2022stacked} for relations that can be enhanced with local-level information, \eg, \emph{attached to} (781\% improvement) and \emph{part of} (441\% improvement). This once again highlights the efficacy of VETO + MEET. 

Fig.~\ref{fig:sgg} shows an illustrative example for the challenges of current SGG models, \eg, SHA + GCL~\cite{dong2022stacked} overfitting to the tail classes after debiasing (panel 4) or Motifs~\cite{zellers2018neural} overfitting to the head classes (panel 5). The generated SG from VETO + MEET (panel 3) attains a better balance between the head and tail predictions.

\subsubsection{SGDet sensitivity analysis}

\begin{table}
\centering
\caption{\textbf{
SGDet sensitivity analysis:} Motifs \vs VETO.}
\label{tab:sgdet_sensitivity}
\footnotesize
\vspace{-1mm}
\begin{tabularx}{\linewidth}{@{}X|c|cc|cc}
\toprule
& & \multicolumn{2}{c|}{\textbf{A@50 drop (\%)}} & \multicolumn{2}{c}{\textbf{A@100 drop (\%)}} \\
\textbf{Detector} & \textbf{mAP drop (\%)} & \multicolumn{1}{c}{\textbf{Motifs}} & \multicolumn{1}{c|}{\textbf{VETO}} & 
\multicolumn{1}{c}{\textbf{Motifs}} & \multicolumn{1}{c@{}}{\textbf{VETO}}\\
\midrule
OD1 & 13 & 3.3 & 5.7 & 3.5 & 4.2\\
OD2 & 32 & 28.0 & 29.5 & 27.0 & 28.0\\
\bottomrule
\end{tabularx}
\vspace{-0.5em}
\end{table}

The experimental results obtained from the VG dataset in Tab.~\ref{tab:vg} indicate that VETO's SGDet performance is slightly lower compared to the baselines. To explore whether our model's performance is affected by the object detector's accuracy, we conducted a sensitivity analysis. Tab.~\ref{tab:sgdet_sensitivity} displays the results of using weaker Object Detectors (OD1 \& 2) with 13\% and \ 32\%  lower mAP, respectively. The resulting drop in A@k reveals that VETO is indeed slightly more sensitive to the detector accuracy. 
However, it is worth noting that despite this sensitivity, our lightweight VETO outperforms state-of-the-art (SOTA) methods on the A@k metric in 2 out of 3 tasks on VG and 3 out of 3 tasks on GQA. Furthermore, VETO's performance in SGDet (VG) is comparable to the SOTA methods.

\section{Conclusion}
We have identified three primary concerns with current SGG models: a loss of local-level information, excessive parameter usage, and biased relation predictions. To address these issues, we introduce the Vision Relation Transformer (VETO) and the Mutually Exclusive Expert Learning (MEET) methods. In most of the cases, our approach achieves superior performance on both biased and unbiased evaluation metrics. Some interesting avenues for future work include improving the contrasting power of multi-experts and reducing the label dependency of experts.

\section*{Acknowledgement}
This work was funded by the Hessian Ministry of Science and the Arts (HMWK) through the
projects “The Third Wave of Artificial Intelligence -- 3AI” and hessian.AI. This work was also supported by the EU ICT-48 Network of AI Research Excellence Center “TAILOR” (EU Horizon 2020, GA No 952215), and the Collaboration Lab “AI in Construction” (AICO). 

\clearpage

{\small
\bibliographystyle{ieee_fullname}
\bibliography{main}

\begin{thebibliography}{10}\itemsep=-1pt

\bibitem{antol2015vqa}
Stanislaw Antol, Aishwarya Agrawal, Jiasen Lu, Margaret Mitchell, Dhruv Batra,
  C.~Lawrence Zitnick, and Devi Parikh.
\newblock {VQA}: {V}isual question answering.
\newblock In {\em Proceedings of the IEEE {I}nternational {C}onference on
  {C}omputer {V}ision}, pages 2425--2433, 2015.

\bibitem{carion2020end}
Nicolas Carion, Francisco Massa, Gabriel Synnaeve, Nicolas Usunier, Alexander
  Kirillov, and Sergey Zagoruyko.
\newblock End-to-end object detection with transformers.
\newblock In {\em Proceedings of the European Conference on Computer Vision},
  volume~1, pages 213--229. Springer, 2020.

\bibitem{chen2019knowledge}
Tianshui Chen, Weihao Yu, Riquan Chen, and Liang Lin.
\newblock Knowledge-embedded routing network for scene graph generation.
\newblock In {\em Proceedings of the IEEE/CVF Conference on Computer Vision and
  Pattern Recognition}, pages 6163--6171, 2019.

\bibitem{chiou2021recovering}
Meng-Jiun Chiou, Henghui Ding, Hanshu Yan, Changhu Wang, Roger Zimmermann, and
  Jiashi Feng.
\newblock Recovering the unbiased scene graphs from the biased ones.
\newblock In {\em Proceedings of the 29th ACM International Conference on
  Multimedia}, pages 1581--1590, 2021.

\bibitem{dai2017detecting}
Bo Dai, Yuqi Zhang, and Dahua Lin.
\newblock Detecting visual relationships with deep relational networks.
\newblock In {\em Proceedings of the IEEE/CVF Conference on Computer Vision and
  Pattern Recognition}, pages 3076--3086, 2017.

\bibitem{desai2021learning}
Alakh Desai, Tz-Ying Wu, Subarna Tripathi, and Nuno Vasconcelos.
\newblock Learning of visual relations: The devil is in the tails.
\newblock In {\em Proceedings of the IEEE/CVF International Conference on
  Computer Vision}, pages 15404--15413, 2021.

\bibitem{dong2022stacked}
Xingning Dong, Tian Gan, Xuemeng Song, Jianlong Wu, Yuan Cheng, and Liqiang
  Nie.
\newblock Stacked hybrid-attention and group collaborative learning for
  unbiased scene graph generation.
\newblock In {\em Proceedings of the IEEE/CVF Conference on Computer Vision and
  Pattern Recognition}, pages 19427--19436, 2022.

\bibitem{dosovitskiy2020image}
Alexey Dosovitskiy, Lucas Beyer, Alexander Kolesnikov, Dirk Weissenborn,
  Xiaohua Zhai, Thomas Unterthiner, Mostafa Dehghani, Matthias Minderer, Georg
  Heigold, Sylvain Gelly, et~al.
\newblock An image is worth 16x16 words: {T}ransformers for image recognition
  at scale.
\newblock In {\em Proceedings of the International Conference on Learning
  Representations}, 2021.

\bibitem{gao2018ican}
Chen Gao, Yuliang Zou, and Jia-Bin Huang.
\newblock Ican: Instance-centric attention network for human-object interaction
  detection.
\newblock In {\em British Machine Vision Conference}, 2018.

\bibitem{gu2019scene}
Jiuxiang Gu, Handong Zhao, Zhe Lin, Sheng Li, Jianfei Cai, and Mingyang Ling.
\newblock Scene graph generation with external knowledge and image
  reconstruction.
\newblock In {\em Proceedings of the IEEE/CVF Conference on Computer Vision and
  Pattern Recognition}, pages 1969--1978, 2019.

\bibitem{hossain2019comprehensive}
MD.~Zakir Hossain, Ferdous Sohel, Mohd~Fairuz Shiratuddin, and Hamid Laga.
\newblock A comprehensive survey of deep learning for image captioning.
\newblock {\em ACM Computing Surveys}, 51(6):1--36, 2019.

\bibitem{hudson2019gqa}
Drew~A. Hudson and Christopher~D. Manning.
\newblock {GQA}: A new dataset for real-world visual reasoning and
  compositional question answering.
\newblock In {\em Proceedings of the IEEE/CVF Conference on Computer Vision and
  Pattern Recognition}, pages 6700--6709, 2019.

\bibitem{iftekhar2023gtnet}
ASM Iftekhar, Satish Kumar, R~Austin McEver, Suya You, and BS Manjunath.
\newblock Gtnet: Guided transformer network for detecting human-object
  interactions.
\newblock In {\em Pattern Recognition and Tracking XXXIV}, volume 12527, pages
  192--205. SPIE, 2023.

\bibitem{kingma2014adam}
Diederik~P. Kingma and Jimmy Ba.
\newblock Adam: A method for stochastic optimization.
\newblock In {\em Proceedings of the International Conference on Learning
  Representations}, 2015.

\bibitem{koner2020relation}
Rajat Koner, Suprosanna Shit, and Volker Tresp.
\newblock Relation transformer network.
\newblock {\em arXiv preprint arXiv:2004.06193}, 2020.

\bibitem{krishna2018referring}
Ranjay Krishna, Ines Chami, Michael Bernstein, and Li Fei-Fei.
\newblock Referring relationships.
\newblock In {\em Proceedings of the IEEE Conference on Computer Vision and
  Pattern Recognition}, pages 6867--6876, 2018.

\bibitem{krishna2017visual}
Ranjay Krishna, Yuke Zhu, Oliver Groth, Justin Johnson, Kenji Hata, Joshua
  Kravitz, Stephanie Chen, Yannis Kalantidis, Li-Jia Li, David~A. Shamma,
  et~al.
\newblock Visual {G}enome: {C}onnecting language and vision using crowdsourced
  dense image annotations.
\newblock {\em International Journal of Computer Vision}, 123(1):32--73, 2017.

\bibitem{li2019visualbert}
Liunian~Harold Li, Mark Yatskar, Da Yin, Cho-Jui Hsieh, and Kai-Wei Chang.
\newblock Visual{BERT}: A simple and performant baseline for vision and
  language.
\newblock {\em arXiv preprint arXiv:1908.03557}, 2019.

\bibitem{li2021bipartite}
Rongjie Li, Songyang Zhang, Bo Wan, and Xuming He.
\newblock Bipartite graph network with adaptive message passing for unbiased
  scene graph generation.
\newblock In {\em Proceedings of the IEEE/CVF Conference on Computer Vision and
  Pattern Recognition}, 2021.

\bibitem{li2017scene}
Yikang Li, Wanli Ouyang, Bolei Zhou, Kun Wang, and Xiaogang Wang.
\newblock Scene graph generation from objects, phrases and region captions.
\newblock In {\em Proceedings of the IEEE/CVF Conference on Computer Vision and
  Pattern Recognition}, pages 1261--1270, 2017.

\bibitem{liao2019natural}
Wentong Liao, Bodo Rosenhahn, Ling Shuai, and Michael Ying~Yang.
\newblock Natural language guided visual relationship detection.
\newblock In {\em Proceedings of the IEEE/CVF Conference on Computer Vision and
  Pattern Recognition Workshops}, 2019.

\bibitem{lin2020gps}
Xin Lin, Changxing Ding, Jinquan Zeng, and Dacheng Tao.
\newblock {GPS-Net:} {G}raph property sensing network for scene graph
  generation.
\newblock In {\em Proceedings of the IEEE/CVF Conference on Computer Vision and
  Pattern Recognition}, pages 3746--3753, 2020.

\bibitem{lu2016visual}
Cewu Lu, Ranjay Krishna, Michael Bernstein, and Li Fei-Fei.
\newblock Visual relationship detection with language priors.
\newblock In {\em Proceedings of the European Conference on Computer Vision},
  volume~1, pages 852--869, 2016.

\bibitem{lu2021context}
Yichao Lu, Himanshu Rai, Jason Chang, Boris Knyazev, Guangwei Yu, Shashank
  Shekhar, Graham~W. Taylor, and Maksims Volkovs.
\newblock Context-aware scene graph generation with {Seq2Seq} transformers.
\newblock In {\em Proceedings of the IEEE/CVF International Conference on
  Computer Vision}, pages 15931--15941, 2021.

\bibitem{ramachandran2019stand}
Prajit Ramachandran, Niki Parmar, Ashish Vaswani, Irwan Bello, Anselm Levskaya,
  and Jon Shlens.
\newblock Stand-alone self-attention in vision models.
\newblock In {\em Advances in Neural Information Processing Systems},
  volume~32, 2019.

\bibitem{ren2015faster}
Shaoqing Ren, Kaiming He, Ross Girshick, and Jian Sun.
\newblock Faster {R-CNN}: Towards real-time object detection with region
  proposal networks.
\newblock In {\em Advances in Neural Information Processing Systems},
  volume~28, 2015.

\bibitem{sharifzadeh2021improving}
Sahand Sharifzadeh, Sina~Moayed Baharlou, Max Berrendorf, Rajat Koner, and
  Volker Tresp.
\newblock Improving visual relation detection using depth maps.
\newblock In {\em Proceedings of the 25th International Conference on Pattern
  Recognition}, pages 3597--3604, 2021.

\bibitem{shi2019explainable}
Jiaxin Shi, Hanwang Zhang, and Juanzi Li.
\newblock Explainable and explicit visual reasoning over scene graphs.
\newblock In {\em Proceedings of the IEEE/CVF Conference on Computer Vision and
  Pattern Recognition}, pages 8376--8384, 2019.

\bibitem{suhail2021energy}
Mohammed Suhail, Abhay Mittal, Behjat Siddiquie, Chris Broaddus, Jayan Eledath,
  Gerard Medioni, and Leonid Sigal.
\newblock Energy-based learning for scene graph generation.
\newblock In {\em Proceedings of the IEEE/CVF Conference on Computer Vision and
  Pattern Recognition}, pages 13936--13945, 2021.

\bibitem{tang2020unbiased}
Kaihua Tang, Yulei Niu, Jianqiang Huang, Jiaxin Shi, and Hanwang Zhang.
\newblock Unbiased scene graph generation from biased training.
\newblock In {\em Proceedings of the IEEE/CVF Conference on Computer Vision and
  Pattern Recognition}, pages 3716--3725, 2020.

\bibitem{tang2019learning}
Kaihua Tang, Hanwang Zhang, Baoyuan Wu, Wenhan Luo, and Wei Liu.
\newblock Learning to compose dynamic tree structures for visual contexts.
\newblock In {\em Proceedings of the IEEE/CVF Conference on Computer Vision and
  Pattern recognition}, pages 6619--6628, 2019.

\bibitem{ulutan2020vsgnet}
Oytun Ulutan, ASM Iftekhar, and Bangalore~S Manjunath.
\newblock Vsgnet: Spatial attention network for detecting human object
  interactions using graph convolutions.
\newblock In {\em Proceedings of the IEEE/CVF Conference on Computer Vision and
  Pattern Recognition}, pages 13617--13626, 2020.

\bibitem{vaswani2017attention}
Ashish Vaswani, Noam Shazeer, Niki Parmar, Jakob Uszkoreit, Llion Jones,
  Aidan~N Gomez, {\L}ukasz Kaiser, and Illia Polosukhin.
\newblock Attention is all you need.
\newblock In {\em Advances in Neural Information Processing Systems},
  volume~30, 2017.

\bibitem{velickovic2018graph}
Petar Veli{\v{c}}kovi{\'{c}}, Guillem Cucurull, Arantxa Casanova, Adriana
  Romero, Pietro Li{\`{o}}, and Yoshua Bengio.
\newblock {Graph attention networks}.
\newblock {\em International Conference on Learning Representations}, 2018.

\bibitem{wang2020tackling}
Tzu-Jui~Julius Wang, Selen Pehlivan, and Jorma Laaksonen.
\newblock Tackling the unannotated: Scene graph generation with bias-reduced
  models.
\newblock {\em Proceedings of the British Machine Vision Conference}, 2020.

\bibitem{wang2019exploring}
Wenbin Wang, Ruiping Wang, Shiguang Shan, and Xilin Chen.
\newblock Exploring context and visual pattern of relationship for scene graph
  generation.
\newblock In {\em Proceedings of the IEEE/CVF Conference on Computer Vision and
  Pattern Recognition}, pages 8188--8197, 2019.

\bibitem{woo2018linknet}
Sanghyun Woo, Dahun Kim, Donghyeon Cho, and In~So Kweon.
\newblock Link{N}et: Relational embedding for scene graph.
\newblock {\em Advances in Neural Information Processing Systems}, 31, 2018.

\bibitem{xie2017aggregated}
Saining Xie, Ross Girshick, Piotr Doll{\'a}r, Zhuowen Tu, and Kaiming He.
\newblock Aggregated residual transformations for deep neural networks.
\newblock In {\em Proceedings of the IEEE Conference on Computer Vision and
  Pattern Recognition}, pages 1492--1500, 2017.

\bibitem{xu2019learning}
Bingjie Xu, Yongkang Wong, Junnan Li, Qi Zhao, and Mohan~S Kankanhalli.
\newblock Learning to detect human-object interactions with knowledge.
\newblock In {\em Proceedings of the IEEE/CVF Conference on Computer Vision and
  Pattern Recognition}, 2019.

\bibitem{xu2017scene}
Danfei Xu, Yuke Zhu, Christopher~B. Choy, and Li Fei-Fei.
\newblock Scene graph generation by iterative message passing.
\newblock In {\em Proceedings of the IEEE Conference on Computer Vision and
  Pattern Recognition}, pages 5410--5419, 2017.

\bibitem{yan2020pcpl}
Shaotian Yan, Chen Shen, Zhongming Jin, Jianqiang Huang, Rongxin Jiang, Yaowu
  Chen, and Xian-Sheng Hua.
\newblock {PCPL:} {P}redicate-correlation perception learning for unbiased
  scene graph generation.
\newblock In {\em Proceedings of the 28th ACM International Conference on
  Multimedia}, pages 265--273, 2020.

\bibitem{yang2018visual}
Hsuan-Kung Yang, An-Chieh Cheng, Kuan-Wei Ho, Tsu-Jui Fu, and Chun-Yi Lee.
\newblock Visual relationship prediction via label clustering and incorporation
  of depth information.
\newblock In {\em Proceedings of the European Conference on Computer Vision
  Workshops}, volume~2, pages 571--581, 2018.

\bibitem{yang2018graph}
Jianwei Yang, Jiasen Lu, Stefan Lee, Dhruv Batra, and Devi Parikh.
\newblock Graph {R-CNN} for scene graph generation.
\newblock In {\em Proceedings of the European Conference on Computer Vision},
  volume~1, pages 670--685, 2018.

\bibitem{yin2018zoom}
Guojun Yin, Lu Sheng, Bin Liu, Nenghai Yu, Xiaogang Wang, Jing Shao, and
  Chen~Change Loy.
\newblock Zoom-{N}et: Mining deep feature interactions for visual relationship
  recognition.
\newblock In {\em Proceedings of the European Conference on Computer Vision},
  volume~3, pages 322--338, 2018.

\bibitem{yin2021learning}
Wei Yin, Jianming Zhang, Oliver Wang, Simon Niklaus, Long Mai, Simon Chen, and
  Chunhua Shen.
\newblock Learning to recover 3d scene shape from a single image.
\newblock In {\em Proceedings of the IEEE/CVF Conference on Computer Vision and
  Pattern Recognition}, pages 204--213, 2021.

\bibitem{yu2020cogtree}
Jing Yu, Yuan Chai, Yujing Wang, Yue Hu, and Qi Wu.
\newblock Cogtree: Cognition tree loss for unbiased scene graph generation.
\newblock {\em Proceedings of the 30th International Joint Conference on
  Artificial Intelligence}, 2021.

\bibitem{yu2017visual}
Ruichi Yu, Ang Li, Vlad~I. Morariu, and Larry~S. Davis.
\newblock Visual relationship detection with internal and external linguistic
  knowledge distillation.
\newblock In {\em Proceedings of the IEEE/CVF Conference on Computer Vision and
  Pattern Recognition}, pages 1974--1982, 2017.

\bibitem{zareian2020bridging}
Alireza Zareian, Svebor Karaman, and Shih-Fu Chang.
\newblock Bridging knowledge graphs to generate scene graphs.
\newblock In {\em Proceedings of the European Conference on Computer Vision},
  volume~3, pages 606--623, 2020.

\bibitem{zellers2018neural}
Rowan Zellers, Mark Yatskar, Sam Thomson, and Yejin Choi.
\newblock Neural motifs: {S}cene graph parsing with global context.
\newblock In {\em Proceedings of the IEEE Conference on Computer Vision and
  Pattern Recognition}, pages 5831--5840, 2018.

\bibitem{zhang2022fine}
Ao Zhang, Yuan Yao, Qianyu Chen, Wei Ji, Zhiyuan Liu, Maosong Sun, and Tat-Seng
  Chua.
\newblock Fine-grained scene graph generation with data transfer.
\newblock In {\em Proceedings of the European Conference on Computer Vision},
  volume~2, pages 409--424. Springer, 2022.

\bibitem{zhang2017visual}
Hanwang Zhang, Zawlin Kyaw, Shih-Fu Chang, and Tat-Seng Chua.
\newblock Visual translation embedding network for visual relation detection.
\newblock In {\em Proceedings of the IEEE/CVF Conference on Computer Vision and
  Pattern Recognition}, pages 5532--5540, 2017.

\end{thebibliography}
}

\end{document}